\documentclass[11pt]{article}

\usepackage{graphicx}
\usepackage{subcaption}
\usepackage{amsmath}
\usepackage{bm}
\usepackage{amsfonts}
\usepackage{amssymb}
\usepackage{url}
\usepackage{authblk} 
\usepackage{xcolor}
\usepackage{algpseudocode}
\usepackage{algorithm2e}
\RestyleAlgo{ruled}
\usepackage[toc,title,page]{appendix}
\usepackage{hyperref}
\usepackage{cases}
\usepackage{multirow}
\usepackage{caption}
\usepackage{subcaption}
\usepackage[usestackEOL]{stackengine}
\usepackage{booktabs}
\usepackage{changepage}
\usepackage{comment}
\usepackage{geometry}
\usepackage{siunitx}

\usepackage{soul}

\newcommand{\dtoprule}{\specialrule{1pt}{0pt}{0.4pt}%
            \specialrule{0.3pt}{0pt}{\belowrulesep}%
            }
\newcommand{\dbottomrule}{\specialrule{0.3pt}{0pt}{0.4pt}%
            \specialrule{1pt}{0pt}{\belowrulesep}%
            }

\newcommand{\norm}[1]{\left\lVert#1\right\rVert}
\newcommand{\cmnt}[1]{\textcolor{blue}{#1}}
\newcommand{\psilow}{\psi^{\text{low}}}
\newcommand{\psihigh}{\psi^{\text{high}}}
\newcommand{\perf}{\mathcal{P}}
\newcommand{\action}{\bm{a}}

\begin{document}

\title{World Modeling for Autonomous Wheel Loaders}

\author[1,2]{Koji Aoshima}
\author[2]{Arvid Fälldin}
\author[3,2]{Eddie Wadbro}
\author[2,4]{Martin Servin}

\affil[1]{Komatsu Ltd.}
\affil[2]{Umeå University}
\affil[3]{Karlstad University}
\affil[4]{Algoryx Simulation}

\date{\today}
\maketitle

\begin{abstract} 
This paper presents a method for learning world models for wheel loaders performing automatic loading actions on a pile of soil. Data-driven models were learned to output the resulting pile state, loaded mass, time, and work for a single loading cycle given inputs that include a heightmap of the initial pile shape and action parameters for an automatic bucket-filling controller. Long-horizon planning of sequential loading in a dynamically changing environment is thus enabled as repeated model inference. The models, consisting of deep neural networks, were trained on data from 3D multibody dynamics simulation of over 10,000 random loading actions in gravel piles of different shapes. The accuracy and inference time for predicting the loading performance and the resulting pile state were, on average, 95\% in \SI{1.2}{ms} and 97\% in \SI{4.5}{ms}, respectively. Long-horizon predictions were found feasible over 40 sequential loading actions.
\end{abstract}


\section{Introduction}
\label{sec:introduction}

The advances in artificial intelligence suggest that computerized control of construction and mining equipment has the potential to surpass that of human operators. Fully or semi-autonomous wheel loaders, not relying on experienced operators, can be an important solution to the increasing labor shortage.
Wheel loaders typically operate on construction sites and quarries, repeatedly filling the bucket with soil and dumping it onto load receivers.
They should move mass at a maximum rate with minimal operating cost without compromising safety.
Recent research on autonomous loading control has focused on increasing performance and robustness using deep learning to adapt to soil properties~\cite{dadhich2020adaptation, azulay2021wheel, fernando2020lies, backman2021continuous, ERIKSSON2023104843, halbach2019neural, BORNGRUND2022104013}. However, these studies have been limited to a single bucket filling and have not considered the task of sequential loading from a pile.
%
A challenge with sequential loading is that the pile state changes with every loading action~\cite{singh1992task, Hemami2009, filla2017towards}.
The altered state affects the possible outcomes of the subsequent loading process and, ultimately, the total performance. A greedy strategy of always choosing the loading action that maximizes the performance for a single loading might be sub-optimal over a longer horizon. End-to-end optimization thus requires the ability to predict the cumulative effect of loading actions over a sequence of tasks. This involves having a model of the world and how it changes under the selected actions.
%
%

This paper introduces wheel loader world models for predicting the outcome of a loading action given the shape of the pile. The outcome includes the loaded mass, loading time, and work, as well as the resulting pile shape.
The net outcome of a sequence of loading actions can then be predicted by repeated inferences of the model on the pile state, thus predicting its sequential evolution as well.
The model aims at supporting optimal planning for autonomous wheel loaders, with the best sequence of loading actions computed from the initial pile surface only. We imagine that the plan would be updated with some regularity (e.g., daily, hourly, or after each loading cycle) from a new observation of the pile surface.
Depending on the planning horizon and dimensionality of the action space, the optimizer requires numerous evaluations of the world model in a short time. Although a simulator based on multibody dynamics and the discrete element method (DEM) can predict the outcome of a loading \cite{aoshima2021}, it is far too computationally intensive and slow for this optimization problem. 
Instead, we explore using a simulator to produce ground truth data and trained deep neural networks for ``instantaneous'' prediction of the outcome of particular loading actions on piles of different shapes. The learned model is informed specifically of the physics of the particular wheel loader and the type of soil used in the simulator. 

The main contribution of this paper is a methodology for learning wheel loader world models and an investigation of how sensitive the model accuracy, inference speed, and required amount of training data are to the model complexity.
The models predict the loading outcome, which includes the resulting pile state, loaded mass, time, and work, given the pile's initial state and a selected loading action.
The net performance of a sequence of loading actions can thus be predicted by repeated model inference.
The world models are shown to be differentiable with respect to the action parameters. This enables the use of gradient-based optimization algorithms for planning of action sequences. 
The paper ends with an analysis of the models' computational properties, leaving the solution of the associated optimization problem to future work.

\section{Related work}
\label{sec:related_work}
Previous scientific work on data-driven models for predicting bucket-soil interactions have studied this from the perspective of automatic control \cite{sotiropoulos2021dynamic,sotiropoulos2020autonomous,saku2021spatio,wagner2022model} and bucket motion planning \cite{lindmark2018computational, wang2022prediction}. Model predictive control of earthmoving operations relies on a model to predict the dig force or soil displacement given a control signal acting over some time horizon and the current (and historic) system state \cite{sotiropoulos2020autonomous,sotiropoulos2021dynamic}. A variational autoencoder (VAE) \cite{wagner2022model} and convolutional autoencoder \cite{saku2021spatio} were used for learning a reduced representation of the terrain surface and combined with recurrent long short-term memory and mixture density neural networks, to learn the time-evolution of the surface during an earthmoving task. Unfortunately, the errors grow during rollout, unless intermediate observations are added, making the resulting state unreliable and the method unstable for long horizons. Other researchers have developed models that predict the bucket fill factor and accumulated work from a planned bucket trajectory and initial soil pile shape \cite{wang2022prediction,lindmark2018computational}. This approach is infeasible in practical use-cases, where it is not possible to track a prescribed bucket trajectory \cite{Dadhich2016}. In the present paper, we draw inspiration from the idea of using a VAE for representing the local soil surface but we focus on accurately predicting the end-state after breakout, and on the accumulated mass, time, and work of the loading cycle. Instead of using a parameterized bucket trajectory as input, we use the control parameters of a force-based automatic bucket-filling controller.

\section{Methodology}
\label{sec:method}

\begin{figure} [!htb]
    \centering
    \includegraphics[width=0.7\textwidth]{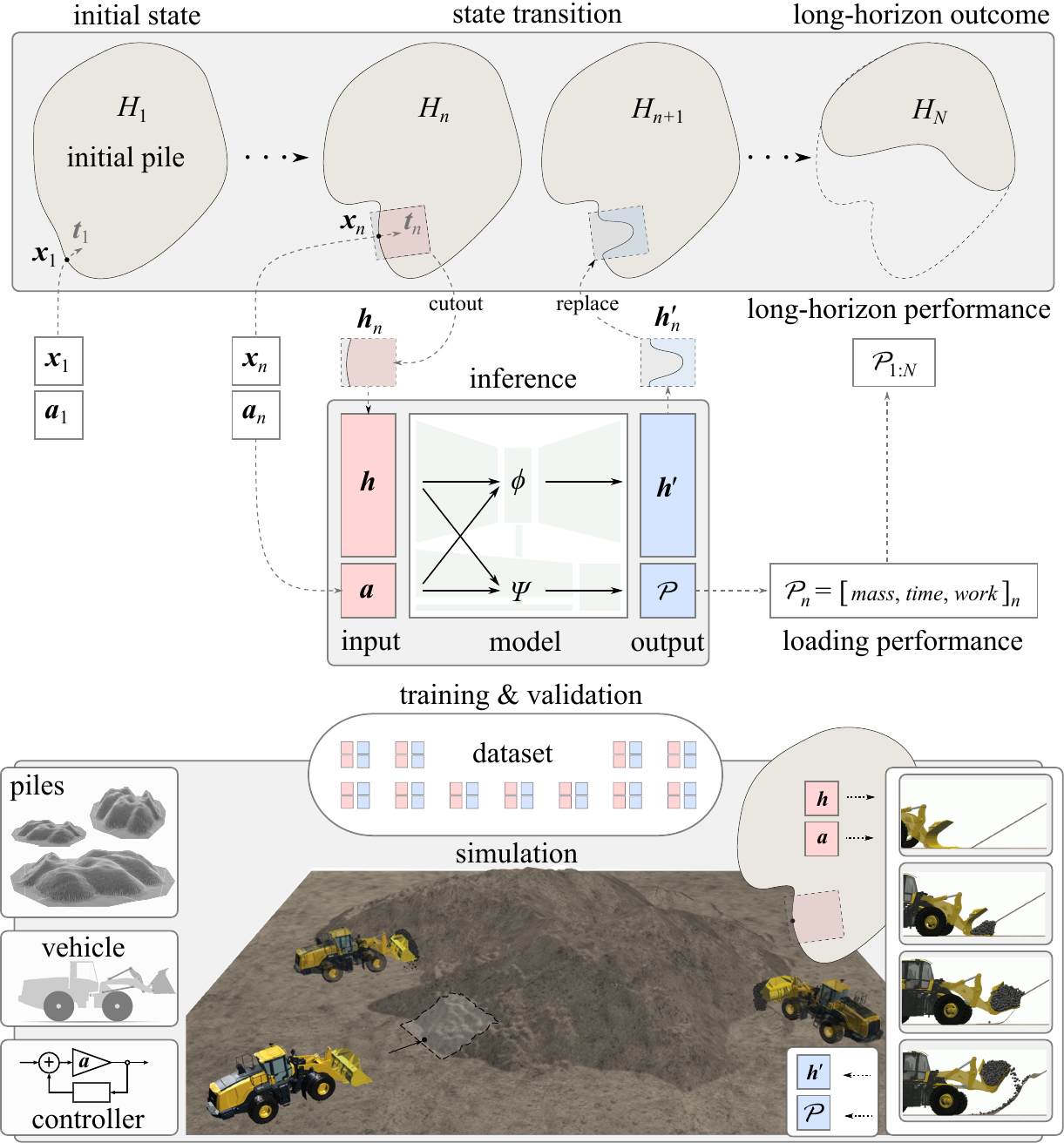}
    \caption{Overview of the wheel loader world model, its development process, and intended usage. First, a dataset of the outcome of parametrized wheel loading actions $\bm{a}$ on a pile with local shape $\bm{h}$ is collected using a simulator. Two models are trained. One model predicts the expected loading performance $\bm{\mathcal{P}}$, in terms of loaded mass, time, and work, given $\bm{a}$ and $\bm{h}$. Another model predicts the new shape $\bm{h}'$ that the pile transitions into after the completed loading. The outcome of a sequence of loading actions on a pile with global state $H$ can be predicted by repeated model inference, given the dig location $\bm{x}$ and heading $\bm{t}$ for each loading.}
    \label{fig:overview}
\end{figure}

The method of developing a wheel loader world model includes creating a simulator~\cite{aoshima2021,aoshima2023sim2real} for the particular machine and its environment. The simulator is based on contacting 3D multibody dynamics with a real-time deformable terrain model that has been shown to produce digging forces and soil displacements with an accuracy close to that of resolved discrete elements and coupled multibody dynamics~\cite{servin2021multiscale}. The simulated wheel loader is equipped with a type of admittance controller \cite{Dobson2017} for automatic bucket-filling, parameterized by four action parameters that determine how the boom and bucket actuation respond to the current dig force. An annotated dataset is created from simulated loading cycles carried out on soil piles of different shapes using different combinations of control parameters. Each simulation results in one data point, consisting of the pile's local heightmap before and after loading, loaded mass, time and work, and the set control parameters that were used. Models are trained to predict how the loading outcome depends on the input values. The models are finally tested on validation data withheld during training. The validated models can then be used to predict the result of new loading operations or be embedded in optimization routines for finding sequences of loading actions that are optimal with respect to productivity or energy efficiency.
The method is illustrated in Fig.~\ref{fig:overview}.

The learned models were investigated for accuracy, inference speed, and the required amount of training data.
In general, model accuracy can be improved by increasing the input and output dimensions and the number of internal model parameters. However, larger models normally require more training data and may have lower inference speeds.
To quantify the difference, we developed both a high-dimensional and a low-dimensional model using different representations of the pile state. 
The high-dimensional model takes a well-resolved heightmap of the pile state as input. This may capture many details of the pile surface at the cost of additional model parameters associated with convolutional layers.
The low-dimensional model takes a heavily reduced representation of the pile surface, involving only four scalar parameters for its slope and curvature.

\section{Wheel loader world models}
\label{sec:models}
 
\subsection{Preface and nomenclature} 
%
%
%
This paper defines the loading cycle as starting with a machine heading in direction $\bm{t}$ to dig location $\bm{x}$ in a pile, represented by the current pile state $H$. The cycle ends when the machine has returned to its initial location after digging in, filling the bucket, breaking out, and leaving the pile in a new state $H'$.
The wheel loader is equipped with an automatic bucket filling controller that is parametrized by some set of \emph{action parameters} $\bm{a}$. The controller is engaged when the machine reaches the dig location $\bm{x}$.
Each loading cycle can be assigned a \emph{performance} $\bm{\mathcal{P}}$, which in this study includes the loaded mass, loading time, and mechanical work. The performance is a consequence of the dynamics of the machine and the soil under the selected action $\bm{a}$ and the initial state of the pile.
The expected performance of a loading cycle, indexed by $n\in\mathbb{N}$, is therefore given by some unknown \emph{performance predictor} function $\bm{\Psi}$.
In other words,
\begin{equation}
    \label{eq:performance_predictor}
    \widehat{\bm{\mathcal{P}}}_n = \bm{\Psi}(H_n, \bm{x}_n, \bm{t}_n, \bm{a}_n),
\end{equation}
where we use the \emph{hat} to distinguish predictions from actual values.
The net effect of a sequence of $N$ loading cycles is the accumulated outcome of a sequence of loading actions, $\bm{a}_1,\ldots,\bm{a}_N$, at pose $(\bm{x}_1, \bm{t}_1),\ldots,(\bm{x}_N, \bm{t}_N)$ applied on a sequence of pile states $H_1,\ldots,H_N$.
Each loading cycle transforms the pile from its previous state to the next, $H_n\to H'_n\equiv H_{n+1}$, according to some unknown \emph{pile state predictor} function $\bm{\Phi}$: 
\begin{equation}
    \label{eq:pile_predictor}
    \widehat{H}'_n = \bm{\Phi}(H_n, \bm{x}_n, \bm{t}_n, \bm{a}_n).
\end{equation}
The net outcome of $N$ sequential loading actions is thus given by the sum
\begin{equation}
    \label{eq:total_outcomes}
    \widehat{\bm{\mathcal{P}}}_{1:N} \equiv\sum_{n=1}^N{\widehat{\bm{\mathcal{P}}}_n(\widehat{H}'_{n-1}, \bm{x}_n, \bm{t}_n, \bm{a}_n)}
\end{equation}
with initial pile state $H'_0 = H_1$. The predictions are associated with some error, that accumulates over repeated loadings from the evolving pile. The \emph{accumulated error} in the pile state and loading performance over a horizon of $N$ cycles is
\begin{align}
    \label{eq:accumulated_pile_error}
    \bm{\mathcal{E}}_{1:N}^{\widehat{H}'} \equiv & \norm{\widehat{H}'_N - H'_N }, \\
    \bm{\mathcal{E}}_{1:N}^{\widehat{\bm{\mathcal{P}}}} \equiv &  \sum_{n=1}^N{ \norm{ \widehat{\bm{\mathcal{P}}}_n - \bm{\mathcal{P}}_n }}.
\end{align}

When solving the problem of finding the sequence of dig locations and loading actions that maximize the net performance $\widehat{\bm{\mathcal{P}}}_{1:N}$, it is beneficial to use gradient-based optimization methods. These require the sought function $\bm{\Psi}$ to be differentiable with respect to $\bm{a}$.

\subsection{Global and local pile state} 
\label{sec:local_pile_state}
We represent the global pile state and the surrounding terrain by a height surface function $z = H(x,y)$ in Cartesian coordinates. 
When making sequential predictions, the aim is to accurately predict the global state of the pile, which may take any shape consistent with the physics of the soil.
To simplify matters, the predictor models take only the \emph{local pile state} as input, assuming that the outcome of a loading depends
only on the local state and not on the global shape of the pile. A dig location, $\bm{x} = [x,y]$ and heading $\bm{t}$ defines a local frame with basis vectors $\{\bm{e}_x, \bm{e}_y, \bm{e}_z\}$,
where $\bm{e}_x$ is aligned with the dig direction, and $\bm{e}_z$ is the vertical direction aligned with the gravitational field.
We represent the local pile state with a discrete heightmap of $I\times J$ grid cells with the side length $\Delta l$. The heightmap, in the local frame, is 
\begin{equation}
    h_{ij} = H(\bm{x} - \delta \bm{e}_x + i \Delta l\bm{e}_x +  j \Delta l \bm{e}_y) - H(\bm{x} - \delta  \bm{e}_x)
\end{equation}
with integers $i\in [0,I]$ and $j \in [-J/2,J/2]$ and constant displacement $\delta$ that ensures a certain fraction of the ground in front of the pile is present in the heightmap. The length and width of the heighmap, $I\Delta l$ and $J\Delta l$, must be large enough to cover the interaction region, see Sec.~\ref{sec:local-heightmap-settings}.

The simplified problem of predicting the loading outcome, Eqs.~\eqref{eq:performance_predictor}-\eqref{eq:total_outcomes}, using the local heightmap is
\begin{align}
    \label{eq:local_performance_predictor}
    \widehat{\bm{\mathcal{P}}}_n & = \psi(\bm{h}_n, \bm{a}_n), \\
    \label{eq:local_pile_predictor}
    \hat{\bm{h}}'_n & = \phi(\bm{h}_n, \bm{a}_n), \\ 
    \label{eq:local_total_outcomes}
    \widehat{\bm{\mathcal{P}}}_{1:N} & \equiv \sum_{n=1}^N{\widehat{\bm{\mathcal{P}}}_n(\hat{\bm{h}}'_{n-1}, \bm{a}_n)}.
\end{align}
with \emph{local} predictor functions $\psi$ and $\phi$ for the performance and pile state. The computational process is described in Algorithm~\ref{alg:long_horizon_prediction} and illustrated in Fig.~\ref{fig:overview}.

\subsection{Local pile characteristics} 
\label{sec:local_characteristics}
As an alternative to representing the local pile state in terms of a heightmap, we introduce a low-dimensional characterization, $\tilde{\bm{h}} \equiv [\theta, \alpha, \kappa_{x}, \kappa_{y}]$ 
, in terms of four scalar quantities: slope angle $\theta$, incidence angle $\alpha$, longitudinal curvature $\kappa_{x}$, and lateral curvature $\kappa_{y}$. These aim to capture the essence of the local pile shape from the perspective of bucket-filling \cite{singh2006factors, singh1998multi}. First, the mean unit normal $\bm{n}$ is computed from the local heightmap. The slope angle relative to the horizontal plane is then computed $\theta \equiv \cos^{-1}\left[ \bm{n} \cdot \bm{e}_z\right]$. 
The incidence angle is the angle between the attack direction $\bm{e}_x = \bm{t}/\norm{\bm{t}}$ and the pile normal projected on the horizontal plane $\bm{n}_{\perp} = \bm{n} - (\bm{n} \cdot \bm{e}_z) \bm{e}_z$, that is, $\alpha \equiv \cos^{-1}(\left[ \bm{n}_{\perp}\cdot \bm{e}_x\right] / \norm{\bm{n}_{\perp}})$. 
%
\begin{figure} [!htb]
    \centering
    \includegraphics[width=0.75\textwidth]{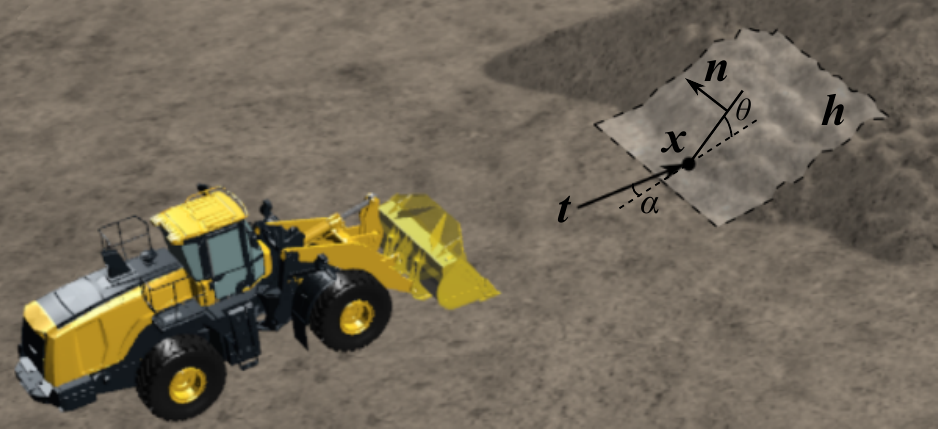}
    \caption{The mean normal $\bm{n}$ of the local heightmap $\bm{h}$ defines a slope angle $\theta$ relative to the horizontal plane.
        The attack angle $\alpha$ is the angle between the dig direction $\bm{t}$ and the normal projected onto the horizontal plane.}
    \label{fig:local_slope}
\end{figure}
%
%
Taking inspiration from \cite{magnusson2015quantitative}, we calculate the local mean curvature in the $\bm{e}_x$ and $\bm{e}_y$ directions by fitting a quadratic surface $h(\bm{x}) \approx b + \bm{c}^\text{T}\bm{x} + \tfrac{1}{2}\bm{x}^\text{T}\bm{Q}\bm{x}$ with surface parameters $\bm{Q} = \mathrm{diag}[-\kappa_x,\enskip -\kappa_y]$, $\bm{c} \in \mathbb{R}^2$, and $b\in \mathbb{R}$. This sign convention makes the curvatures positive for convex pile shape, which is the recommended shape for high-performance loading \cite{singh1998multi}.
%
%

%
\begin{algorithm} [!htb]
\caption{Long-horizon prediction using world models}
\label{alg:long_horizon_prediction}
Initialization: $\widehat{H}'_0 = H_1$\\
\For{$n = 1, \ldots, N$}
{
    $\bm{x}_n, \bm{t}_n \leftarrow$ select dig pose along edge of $\widehat{H}'_{n-1}$\\
    $\bm{h}_n \leftarrow$ get local heightmap of $\widehat{H}'_{n-1}$ at pose $\bm{x}_n, \bm{t}_n$\\
    $\bm{a}_n \leftarrow$ select appropriate dig action for $\bm{h}_n$\\
    \uIf{high\_dim\_model}{
     $\widehat{\bm{\mathcal{P}}}_n := \psihigh(\bm{h}_n, \bm{a}_n)$ \\
     $\hat{\bm{h}}'_n := \phi(\bm{h}_n, \bm{a}_n)$ \\
     $\widehat{H}'_n \leftarrow$ apply new local heightmap $\hat{\bm{h}}'_n$ to $\widehat{H}_n$ at pose $\bm{x}_n, \bm{t}_n$ \\
    }
    \uElseIf{low\_dim\_model}{
     $\bm{\tilde{h}}_n \leftarrow$ compute local pile characteristics of $\bm{h}_n)$ \\
     $\widehat{\bm{\mathcal{P}}}_n := \psilow(\bm{\tilde{h}}_n, \bm{a}_n)$ \\
     $\widehat{H}'_n \leftarrow$ apply cellular automata to $\widehat{H}_n$ after removing mass ($\widehat{\bm{\mathcal{P}}}_n$) at $\bm{x}_n, \bm{t}_n$\\
    }
}
$\widehat{\bm{\mathcal{P}}}_{1:N} = \sum_{n=1}^N{\widehat{\bm{\mathcal{P}}}_n}$ \\
\end{algorithm}

\subsection{Performance predictor model}  
\label{sec:performance_predictor_model}
\begin{figure} [!htb]
    \begin{subfigure}{0.5\textwidth}
    \centering
    \includegraphics[clip,trim=0 0 0 0, width=0.7\textwidth]{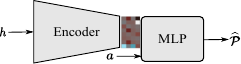}\\
    \scriptsize{\textbf{(a)}}
    \end{subfigure}
    \begin{subfigure}{0.5\textwidth}
    \centering
    \includegraphics[clip,trim=0 0 0 0, width=0.4\textwidth]{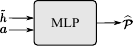}\\
    \scriptsize{\textbf{(b)}}
    \end{subfigure}
    \caption{Illustration of the model architectures. In the high-dimensional model (a), a convolutional neural network encodes the heightmap before being fed with the action vector to an MLP. In the low-dimensional model (b), the pile characteristics and action vector are inputs directly to the MLP.}
    \label{fig:model_structures}
\end{figure}

We developed two different models to predict the loading performance, referred to as the high-dimensional model and the low-dimensional model, respectively.
The \emph{high-dimensional model}, $\psihigh(\bm{h},\action)$, 
takes the local heightmap $\bm{h}$ as input, while the \emph{low-dimensional model}, $\psilow(\tilde{\bm{h}},\action)$,
uses the local pile characteristics $\tilde{\bm{h}}$. Both models take the same action parameters $\bm{a}$ as 
input, and they both output a performance vector $\widehat{\bm{\perf}}$. The name distinction comes from $\text{dim}(\bm{h}) \gg \text{dim}(\tilde{\bm{h}})$. The difference is reflected in the different neural network architectures of the models (Fig.~\ref{fig:model_structures}).
The high-dimensional model uses three convolutional layers and a fully connected linear layer to encode $\bm{h}$ into a latent vector of length 32 before it is concatenated with $\bm{a}$. $\bm{h}$ is interpolated to a size of $32 \times 32$ before encoding. The convolutions use ten filters of $3\times 3$ kernels and unit stride with zero-padding. They are followed by batch normalization and max pooling with window size $2\times2$. The activation function is subject to hyperparameter tuning.
The concatenation steps in both $\psihigh$ and $\psilow$ are followed by multilayer perceptrons (MLP) of identical architectures that are also subject to hyperparameter tuning.


%

\subsection{Pile state predictor model}  
\label{sec:pile_state_predictor_model}
\begin{figure}[!htb]
    \centering
    \includegraphics[width=0.55\textwidth]{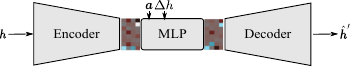}
    \caption{The model architecture of the pile state predictor model. A VAE is paired with an MLP that learns state transition in the latent space.}
\label{fig:pile_state_predictor_model_structure}
\end{figure}

The pile state predictor model combines a VAE architecture with an MLP to make predictions via three steps: First, an encoder network compresses the initial pile state $\bm{h}\in\mathbb{R}^{52\times52}$ into a lower-dimensional, regularized latent representation $\bm{z}\in\mathbb{R}^{64}$. Next, given $\bm{z}$ along with the loading action $\bm{a}$ and a scale factor $\Delta h \equiv \max(\bm{h}) - \min (\bm{h})$, an MLP predicts the regularized latent representation of the new pile state $\bm{z}'$. Last, a decoder network constructs the resulting pile state $\hat{\bm{h}}'\in\mathbb{R}^{52\times 52}$ from $\bm{z}'$. 
Fig.~\ref{fig:pile_state_predictor_model_structure} illustrates the inference process.
%
For the VAE encoder/decoder blocks, we use the same CNN architecture as in \cite{Kuchur2021}. 

The MLP block consists of two hidden layers with 1,024 nodes and uses Leaky ReLU activation. Note that $\bm{h}$ is interpolated to size $64\times 64$ before encoding and back to $52 \times 52$ after decoding to fit the off-the-shelf architecture.

\subsubsection{Post-processing}
\label{ref:post-processing}
\begin{figure}[h]
    \centering
    \includegraphics[width=0.5\textwidth]{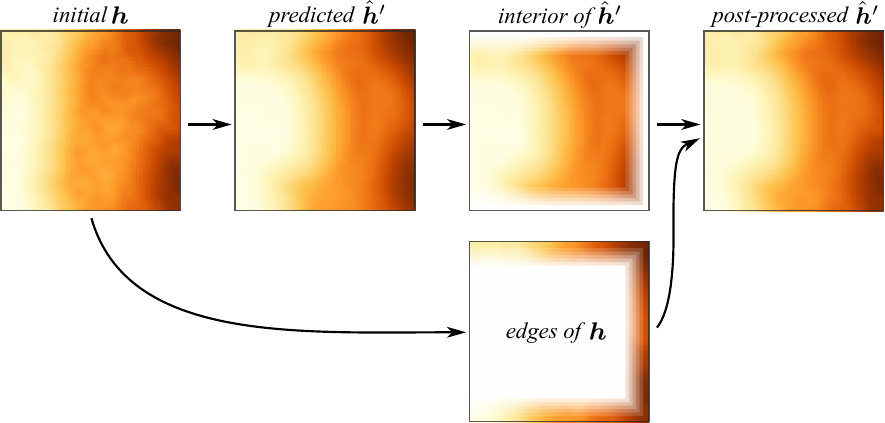}
    \caption{The post-processing interpolates between the interior of the predicted $\hat{\bm{h}}'$ and the edges of the initial heightmap $\bm{h}$
        to preserve the details along the boundary. The pile in the heightmaps, color-coded by height, was dug from the left.}
    \label{fig:post_process}
\end{figure}

VAEs are known to produce somewhat blurry images \cite{kingma2019introduction}.
In our case, this led to a loss of detail in the decoded pile states.
In our preliminary tests, we found that we could improve the prediction accuracy on average by reusing some information from the initial pile state. By interpolating between $\bm{h}$ and $\hat{\bm{h}}'$ along the edges, the details of the outer area are retained. This makes the predicted local pile states blend seamlessly into the global pile state after the substitution.
Fig.~\ref{fig:post_process} illustrates the post-processing.

\subsection{Low-dimensional pile state prediction using cellular automata}  
\label{sec:avalanching_algorithm}
The low-dimensional predictor model did not have direct access to the local heightmap.
Instead, we constructed a pile state predictor method that acts directly on the global pile state. 
It works in two stages. First, the predicted load mass (part of the performance vector $\widehat{\bm{\mathcal{P}}}_n$) is removed from the global pile state at the dig location. 
This is done by computing the corresponding volume and subtracting this from the global pile along a strip as wide as the bucket, starting at $\bm{x}_n$ and stretching along $\bm{t}_n$.
In the second stage, soil mass is redistributed to eliminate any slopes steeper than the set angle of repose. 
The mass-preserving cellular automata algorithm in \cite{pla2004approximation} was employed, with the ``velocity of flowing matter'' set to $z_+=0.2\Delta l$.

\subsection{Delimitations} 
\label{sec:delimitations_assumptions}
The predictor models developed in this paper are limited to a single type of soil, gravel, which was assumed to be homogeneous. However, the process of learning world models is applicable to any other soil type supported by the simulator.
												
This paper does not consider soil spillage on the ground after breaking out with an overfilled bucket. Spillage would affect subsequent performance, either by causing a loss in control precision or traction or by requiring the ground to be cleared.
In this paper, ``loading'' refers to the bucket-filling phase, including approaching the pile straight ahead and reversing in the opposite direction after the breakout, following the definition in \cite{lindmark2018computational}.
To account for the full loading performance, one should incorporate
the V-cycle manoeuver, emptying the bucket into the load receiver, and clearing spillage from the ground.

%
\section{Simulator and loading controller} 
\label{sec:simulator_and_loading_controller}
%
%
\subsection{Simulator} 
\label{sec:simulator}
\begin{figure}[h]
    \centering
    \includegraphics[height=0.1\textwidth, trim = 40 0 30 50, clip]{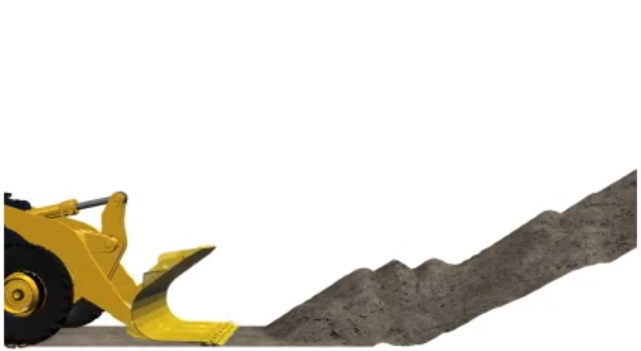}
    \includegraphics[height=0.1\textwidth, trim = 40 0 30 50, clip]{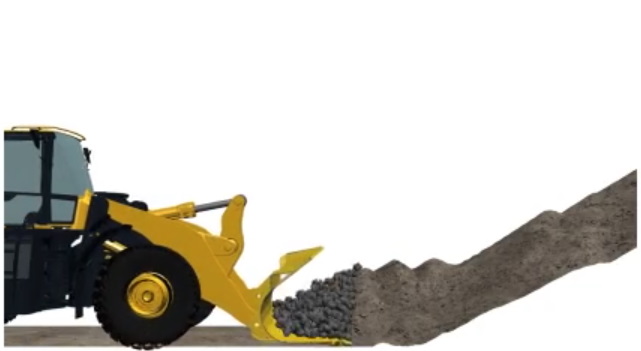}
    \includegraphics[height=0.1\textwidth, trim = 40 0 30 50, clip]{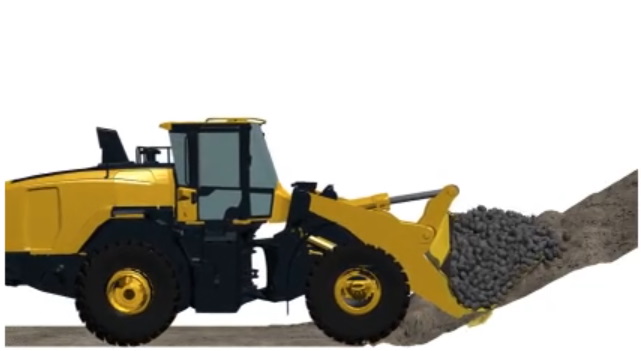}
    \includegraphics[height=0.1\textwidth, trim = 40 0 30 50, clip]{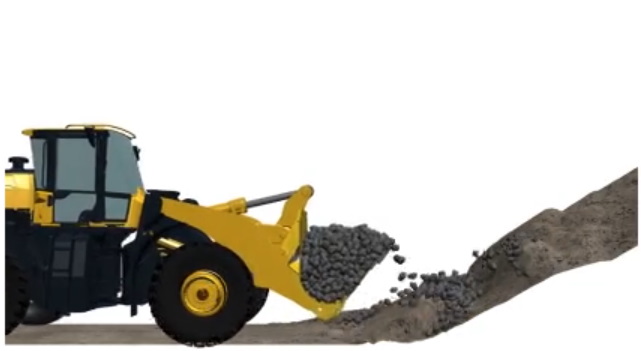}
    \includegraphics[height=0.1\textwidth, trim = 40 0 30 50, clip]{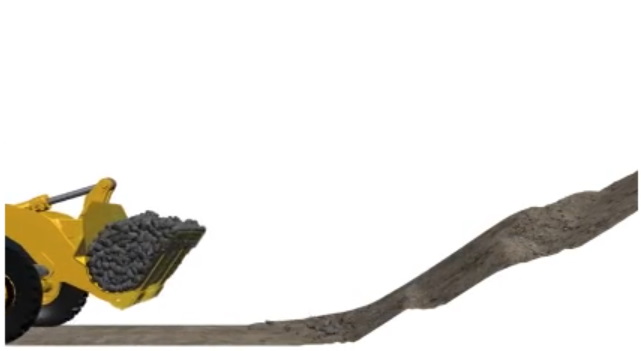}
    \caption{Image sequence of a simulated loading cycle. A video version is found in Supplementary Video 1.}
    \label{fig:simulator}
\end{figure}

We collected data using a simulator developed~\cite{aoshima2021} and validated~\cite{aoshima2023sim2real} in a previous work. Images and videos from the simulator are found in Fig.~\ref{fig:overview}, Fig.~\ref{fig:simulator} and Supplementary Video 1.
The simulator combines a deformable terrain model with a wheel loader model and runs approximately in real-time.
The terrain model, introduced in \cite{servin2021multiscale}, combines the representation of soil as a continuous solid, distinct particles, and rigid multibodies. 
When a digging tool comes in contact with the terrain, the active zone of moving soil is predicted and resolved in terms of particles using DEM. 
When particles come to rest on the terrain surface, they merge back into the solid soil. The computational processes preserve the total mass.
The wheel loader is modeled as a rigid multibody system,
matching roughly the dimensions and physical properties of a Komatsu WA320-7.
The model has five actuated joints, with hinge motors powering the driveline and steering and linear motors for the boom and bucket cylinders.
The motors are controlled by specifying a momentaneous joint target speed and force limits. The actuators will run at the set target speed only if the system dynamics and required constraint force do not exceed their respective force limits.
The limits are set to match the strength of the real machine \cite{wa320brochure}.
The transmission driveline model includes the main, front, and rear differential couplings to the wheels. The force interaction between the vehicle and the terrain occurs through the tire-terrain and bucket-terrain contacts. The resulting forces on the bucket depend on the shape and amount of active soil, its dynamic state, and mechanical properties, including bulk density, internal friction angle, cohesion, and dilatancy angle. As in \cite{aoshima2023sim2real}, these are set to the values \SI{1727}{kg/m^3}, $32^\circ$, $0$ kPa, and $8^\circ$, which are intended to reflect the properties of dry gravel. 
The simulations were run with the physics engine AGX Dynamics \cite{AGX2023} using a $0.01$ s time-step and terrain grid cell-size of $\SI{0.1}{m} \times \SI{0.1}{m}$.
%
%
%

\subsection{Loading controller} 
\label{sec:loading_controller}
\begin{figure}[h]
    \centering
    \includegraphics[width=0.21\textwidth, trim = 0 0 0 0, clip]{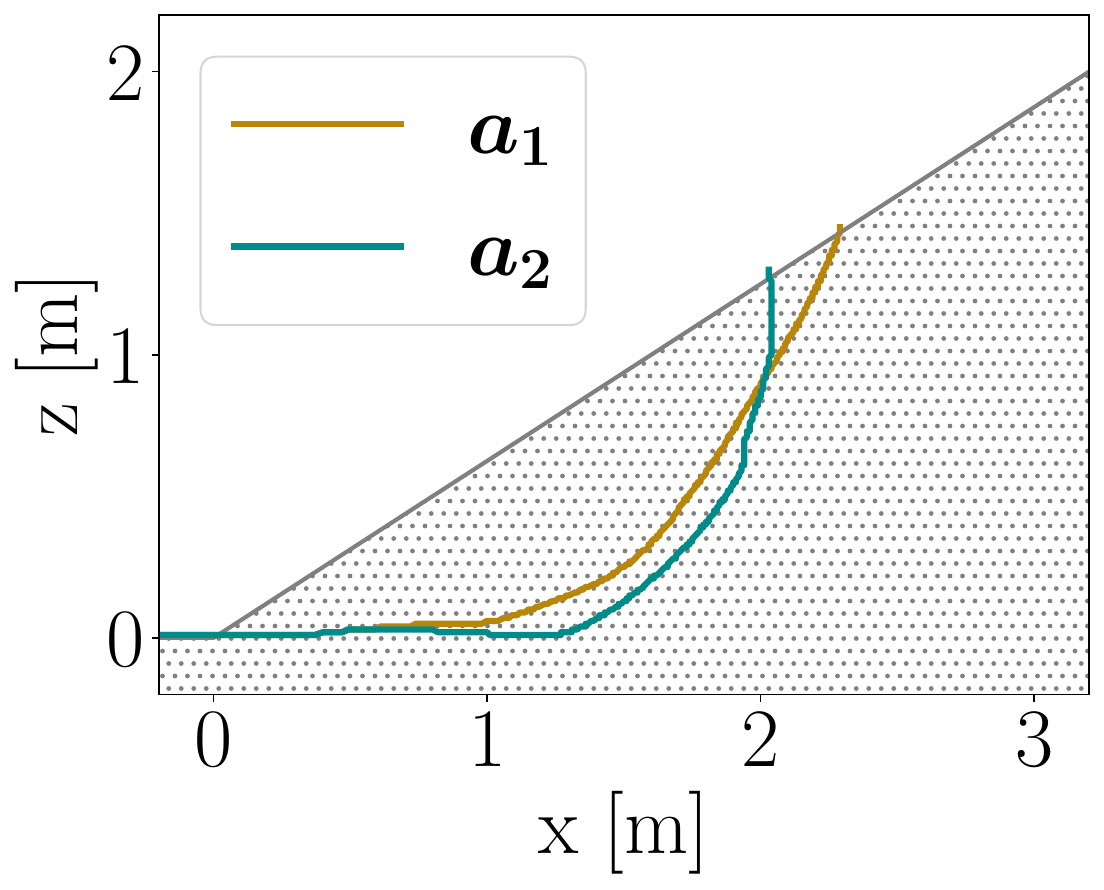}
    \includegraphics[width=0.18\textwidth, trim = 75 0 0 0, clip]{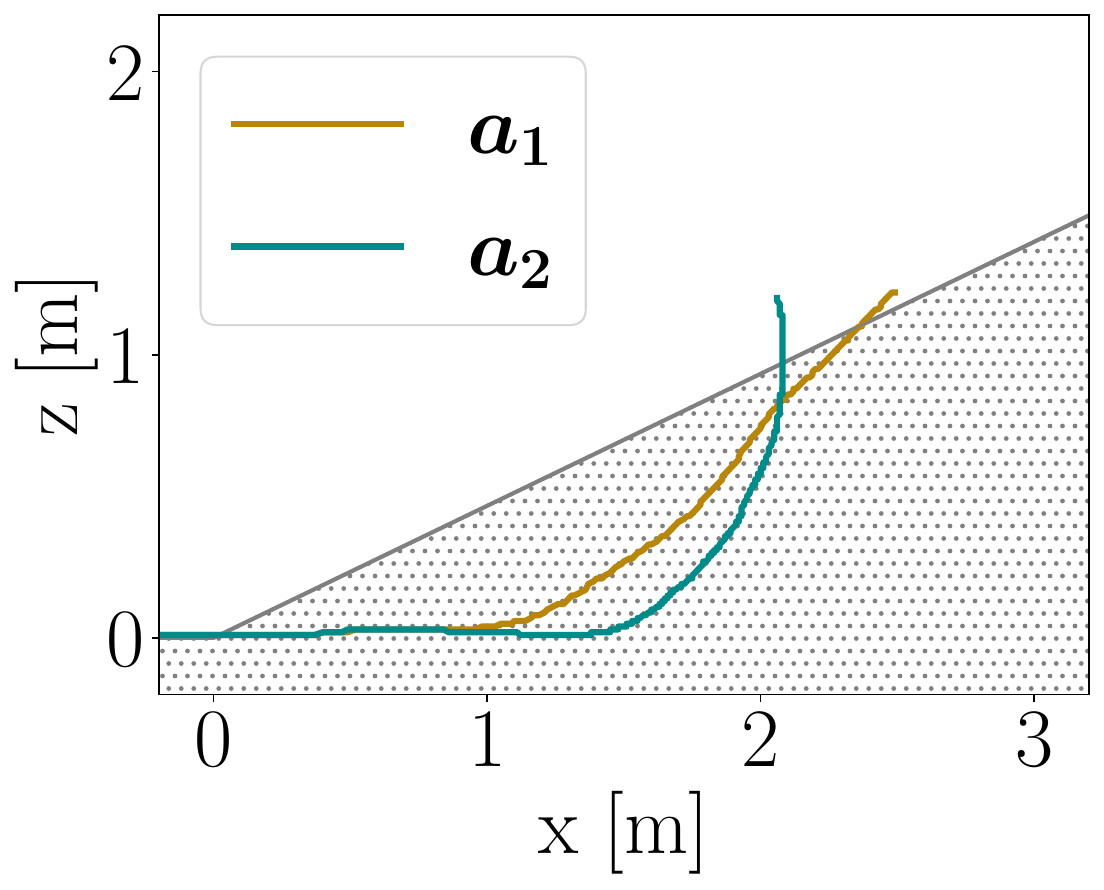}
    \includegraphics[width=0.18\textwidth, trim = 75 0 0 0, clip]{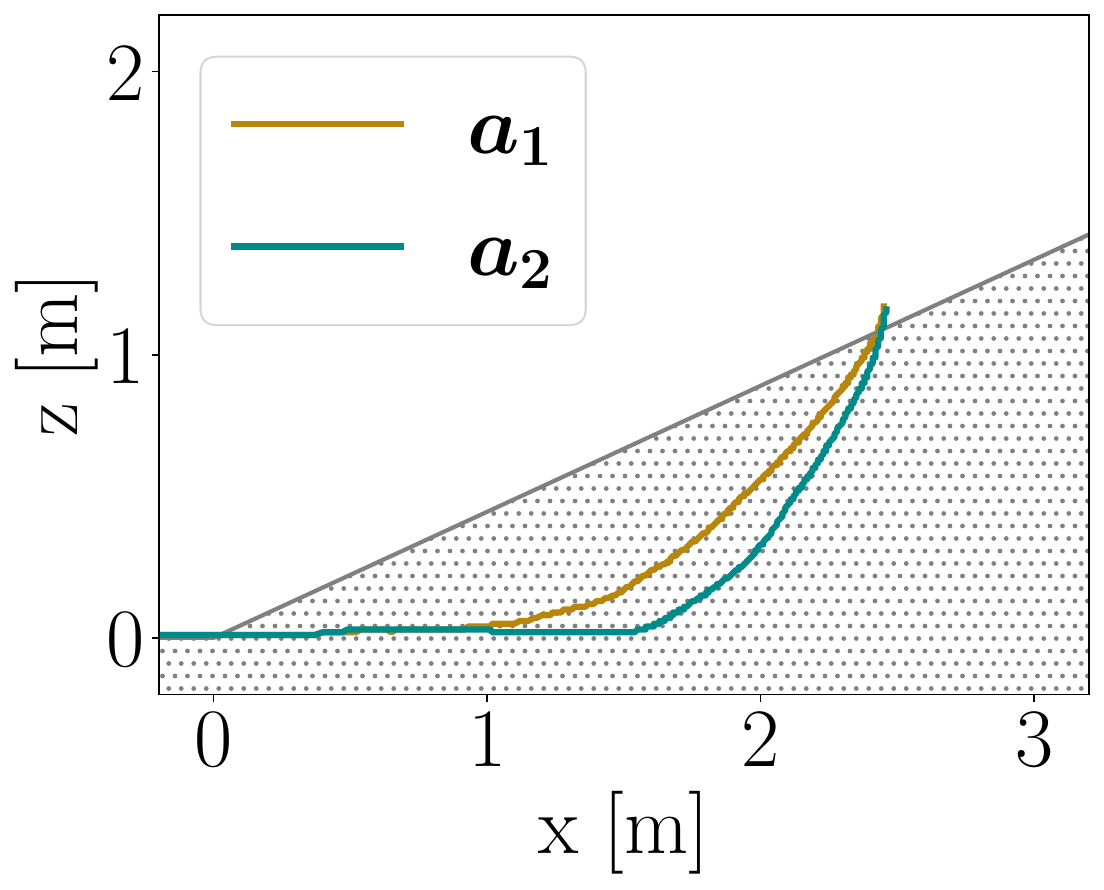}
    \includegraphics[width=0.18\textwidth, trim = 75 0 0 0, clip]{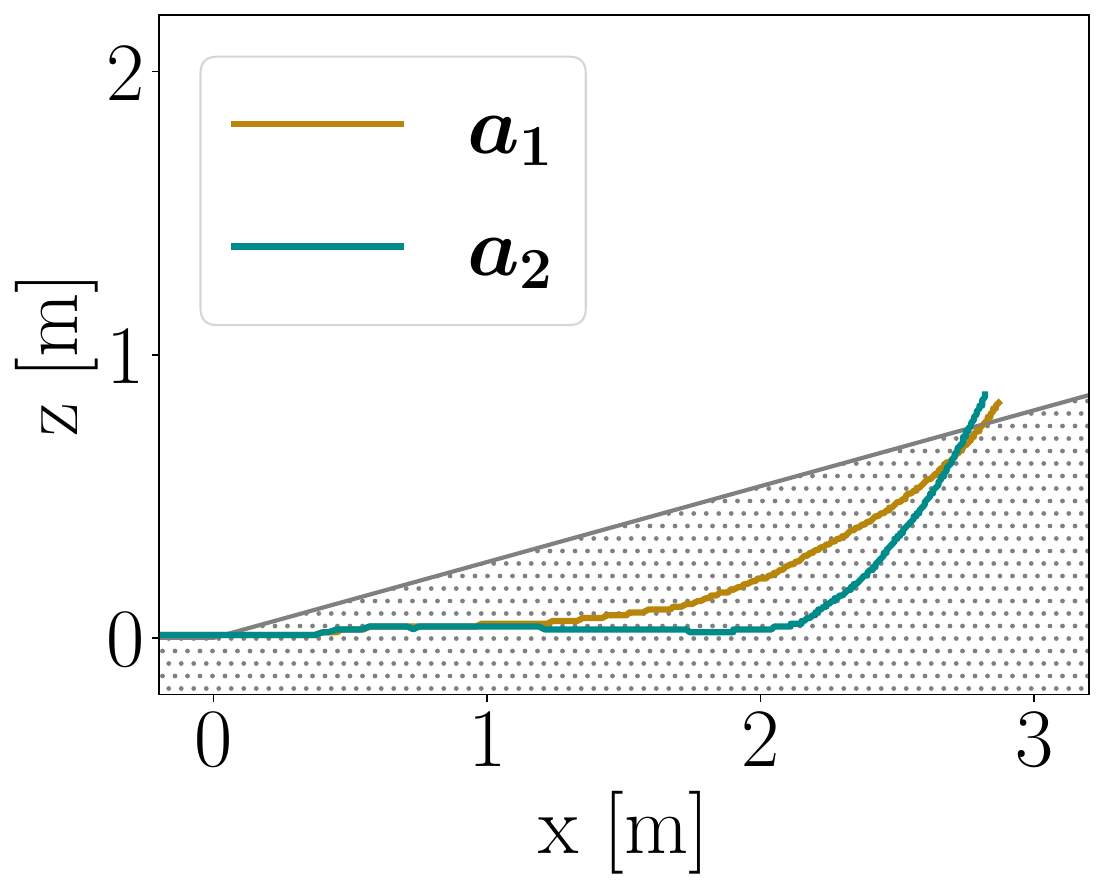}
    \includegraphics[width=0.18\textwidth, trim = 75 0 0 0, clip]{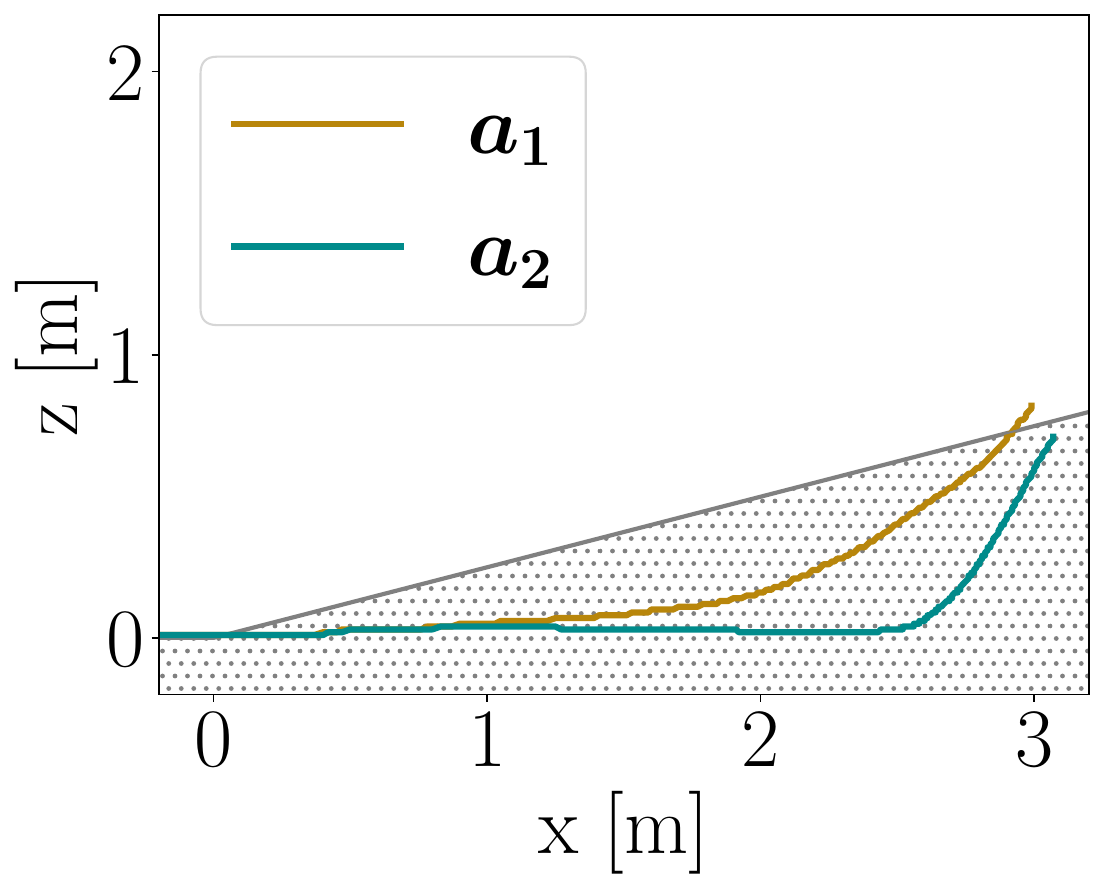}
    \caption{Examples of simulated trajectories using two different sets of action parameters $\bm{a}_1=[0.1, 1.0, 0.1, 2.5]$ and $\bm{a}_2=[0.5, 1.0, 0.4, 4.0]$ on five piles with a slope ranging between $30^\circ$ and $10^\circ$.}
    \label{fig:example_trajectories}
\end{figure}

The loading scenario starts with the machine driving at $\SI{8}{km/h}$ towards a pile $\SI{5}{m}$ from the target dig location.
The bucket is lowered and held level to the ground during the approach.
The target drive velocity is kept constant during the bucket-filling phase, although the actual velocity will vary due to digging resistance.
Throughout the cycle, the machine maintains the same heading.
When the bucket reaches the set dig location, the automatic bucket-filling controller is engaged. 
This controller was inspired by the admittance controller from \cite{Dobson2017}, which regulates the bucket actuator velocity using the measured boom cylinder force. 
Our admittance controller uses the same method but applies it to both the boom and bucket actuators.
The controller determines the target velocities of the boom and bucket cylinders, $v_\text{bm}$ and $v_\text{bk}$, as follows
\begin{align}
        v_\text{bm} & =  \text{clip}\left( k_\text{bm}\left[ f_\text{bm}/f_\text{b0} -\delta_\text{bm} \right],0,1\right) v_\text{bm}^\text{max}, \\
        v_\text{bk} & =  \text{clip}\left( k_\text{bk}\left[ f_\text{bm}/f_\text{b0} -\delta_\text{bk} \right],0,1\right) v_\text{bk}^\text{max},
\end{align}
where $f_\text{bm}$ is the momentaneous boom cylinder force and $\text{clip}(value, min, max)$ limits $value$ to the maximum and minimum values. 
The ramp function has four parameters (velocity gains $k_\text{bm}$ and $k_\text{bk}$, and force thresholds
$\delta_\text{bm}$ and $\delta_\text{bk}$) that parameterize the behavior of the controller. The actuator max speeds, $v_\text{bm}^\text{max}$ and $v_\text{bk}^\text{max}$, and the normalizing boom force $f_\text{b0}$ are set using specifications from the manufacturer.
%
%
The control parameters are collected in the action vector
\begin{equation}
    \bm{a} = \left[\delta_\text{bm}, k_\text{bm}, \delta_\text{bk}, k_\text{bk}\right].
\end{equation}
Parameters $\delta_\text{bm}$ and $\delta_\text{bk}$ regulate what force magnitude is required to trigger the lift and tilt reactions while 
parameters $k_\text{bm}$ and $k_\text{bk}$ control how rapid the reaction is.
Different values of the control parameter thus render different scooping motions, as illustrated by the examples in Fig.~\ref{fig:example_trajectories}. The challenge is to select the parameters most appropriate for the local pile state.
Note that the controller operates only on the lift and tilt actuators. The vehicle keeps thrusting forward to reach the set target drive velocity. If there is insufficient traction, the wheels may slip.

The bucket-filling controller is stopped if the bucket achieves the tilt end position, reaches a penetration depth of \SI{3.2}{m}, breaks out of the initial surface, or if the loading duration exceeds \SI{15}{s}. The bucket is then tilted with maximum speed to its end position (if not there already). The brake is applied for at least $0.5$\,s to let the agitated soil settle.
After that, the vehicle is driven in reverse with a target speed of $8$\,km/h, lift $ v_\text{bm} = 0.6 v^\text{max}_\text{bm}$ and tilt $v_\text{bk} = 0.6 v^\text{max}_\text{bk}$
to reach a boom angle of $-20^\circ$ relative to the horizontal axis, and the bucket tilt ends.
The scenario ends when the vehicle has reversed $5$\,m from the dig location.
To simplify time and energy comparison between simulations, they all use the same start and end distance from the dig location and all end with identical boom and bucket angles.
The target speeds and force measurements are smoothed using a $0.1$\,s moving average to avoid a jerky motion.

%
%
\section{Data collection and model training} 
\label{sec:data_collection_and_model_training}
This section describes how data was collected from the simulator and used for learning and evaluating the predictor models.

\subsection{Loading outcome} 
\label{sec:loading_outcomes}
The models were developed to predict the outcome of a loading cycle in terms of the performance $\bm{\mathcal{P}}$ and the resulting local pile state $\bm{h}'$ given the initial pile $\bm{h}$ and action $\bm{a}$.
The performance was measured using the three essential scalar metrics, namely load mass (tonne), loading time (s), and work (kJ). Hence, $\bm{\mathcal{P}} \in \mathbb{R}^3$. 
The load mass is measured as the amount of soil the bucket carries at the end of each loading cycle.
The loading time is the time elapsed between each loading cycle's start and end.
																																	   
The work is the energy consumed by the boom and bucket actuators and the forward drive. It includes the energy required to fill the bucket, break out, raise the bucket, and accelerate the vehicle and soil. Much of the work is lost to frictional dissipation internally in the soil and at the bucket-soil interface. The physics-based simulator accounts for this dissipation. Energy dissipation in the vehicle engine and hydraulics is not included, and should be added if needed.
\subsection{Data collection} 
\label{sec:data_collection}

\begin{figure} [!htb]
        \centering
        \includegraphics[clip,trim=0 0 0 0, width=0.8\textwidth]{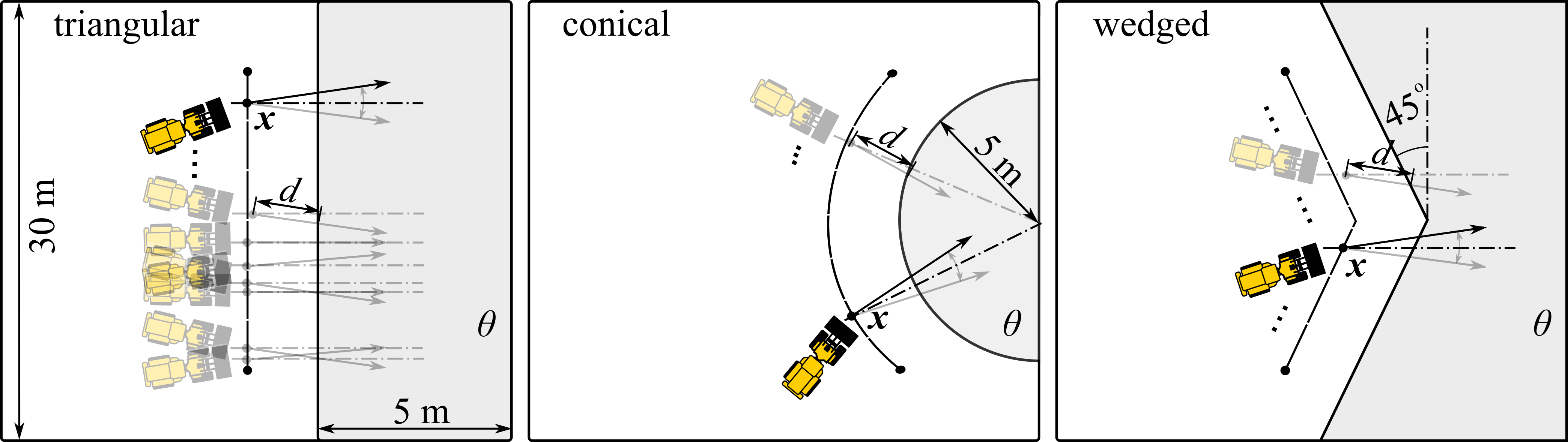}
        \caption{Six initial seed piles are used, triangular, conical, and wedged, with slope angle $\theta$. The dig location $\bm{x}$ and heading are randomized at each loading.}
    \label{fig:pile_seeds}
\end{figure}

We collected a total of $10,718$ samples, $\{\bm{h}_n, \bm{a}_n, \bm{h}'_n, \bm{\mathcal{P}}_n\}_{n=1}^{10718}$,
by repeating the loading scenario (Sec.~\ref{sec:loading_controller}) in the simulation.
First, six different initial piles were prepared: two triangular, two conical, and two wedged with respective slope angles of $20^\circ$ and $30^\circ$. These are illustrated in Fig.~\ref{fig:pile_seeds}. Perlin noise \cite{perlin1985image} was applied to the surfaces to increase the variability in pile shape.
For each of the six initial piles, 30 consecutive loading cycles were simulated. Each loading cycle produced one data point $(\bm{h}_n, \bm{a}_n, \bm{h}'_n, \bm{\mathcal{P}}_n)$. 
Since the resulting pile state was used as the initial state in the next loading cycle, a variety of piles of different shapes was achieved. 
This process was repeated 60 times for each of the six seed piles.
%
In each simulation, a random dig location was selected, and the wheel loader was positioned at $\bm{x}_n$, a distance of 5\,m from the pile, and the heading was given a random disturbance of $\pm 10^\circ$, see Fig.~\ref{fig:pile_seeds}. 
Each loading used a set of action parameters, $\bm{a}_n$, randomly sampled by Latin Hypercube Sampling (LHS)
with the ranges $0.0 \leq \delta_\text{bm}, \delta_\text{bk} \leq 0.7$ and $0.0 \leq k_\text{bm}, k_\text{bk} \leq 5.0$.
Note that the actual slope angle $\theta$ and incidence angle $\alpha$ varied according to the dig location because of the Perlin noise.
%
%
\subsection{Local heightmap settings} 
\label{sec:local-heightmap-settings}
We found that the performance and pile state predictor models benefited from using different sizes of the local heightmap and displacement $\delta$ in the dig direction. 
We used a \SI{3.6}{m} sided quadratic heightmap discretized by $36\times 36$ grid cells for the performance predictor. The size leaves a margin of $0.5$\,m on the sides of the bucket.
Due to avalanching the pile state predictor needed a larger heightmap to capture the state that the pile finally settles into after the breakout.
For this, we used a heightmap with 5.2\,m side lengths, discretized in $52\times 52$ cells, and displaced the heightmap 1.0 m towards the approach direction.

\subsection{Model training} 
\label{sec:model_training}
%

This section describes the training processes of
the performance predictor (Sec.~\ref{sec:performance_predictor_model_training}) and the pile state predictor model (Sec.~\ref{sec:pile_state_predictor_model_training}), respectively.
The models were implemented using PyTorch.
\subsubsection{Performance predictor model} 
\label{sec:performance_predictor_model_training}
%
%
The models were trained until convergence with a mean squared error (MSE) loss using the Adam optimizer using a learning rate of $10^{-5}$.
Hyperparameter tuning was done via grid search over the number of hidden layers (1, 2, or 3), the number of units ($2^3,2^4,\ldots,2^{12}$) in each hidden layer with 0.1 dropout rate,
as well as the activation function in fully connected and convolutional layers (Leaky ReLU or Swish \cite{ramachandran2018swish}).
The dataset was first min-max normalized.
The training and validation set (split 90/10) was sequentially increased in size from 100 to 9,646 samples, while the test set size was fixed at 1,072 samples.
\subsubsection{Pile state predictor models} 
\label{sec:pile_state_predictor_model_training}
The pile state predictor model was trained in a two-stage process. 
First, we trained the VAE to perform heightmap reconstruction. 
Before training, the heightmaps were re-scaled by subtracting the average slope and applying min-max scaling to them individually,
\begin{align}
    \bm{h}_n & := \bm{h}_n - \frac{1}{N}\sum_{m=1}^N \bm{h}_{m} \label{eq:scaling_1}\\
    \bm{h}_{n} & := \frac{\bm{h}_{n} - \min(\bm{h}_{n})}{\max(\bm{h}_{n}) - \min{\bm{h}}_{n}} \label{eq:scaling_2}.
\end{align}
The standardization in Eq.~\eqref{eq:scaling_1} centers each height distribution around 0 across the entire data set before applying the min-max scaling in Eq.~\eqref{eq:scaling_2}. We found this helps to preserve the pile-ground boundary in our reconstructions. 
The normalization in Eq.~\eqref{eq:scaling_2} makes the VAE agnostic to scale, which we found increases its overall performance.
The VAE was trained using the Adam optimizer with a learning rate of $10^{-3}$.
We used a weighted sum of the element-wise MSE and the Kullback-Leibler Divergence (KLD), $\mathcal{L}_\mathrm{VAE} = \mathcal{L}_\mathrm{MSE} + 0.1 \mathcal{L}_\mathrm{KLD}$ for the loss function.

In the second training phase, we used the encoder trained in the first phase to encode sampled pile state pairs $(\bm{h}, \bm{h}')$ into latent pairs $(\bm{z}, \bm{z}')$.
We then trained an MLP on the latent pairs with a 0.1 dropout rate for the hidden layers and used the Adam optimizer with a learning rate $10^{-5}$.

The full dataset was used with a split ratio of 80/10/10 in training, validation, and test data for the pile state predictor. Since the wheel loader geometry and actions are left-right symmetric, we applied random reflections to the heightmaps to augment the dataset.
\section{Results}  
\label{sec:results}
The models for predicting the loading performance and resulting pile state were evaluated separately on the hold-out test data. The best models were selected and used in combination to test their ability to predict the outcome of sequential loadings. An overview of the full dataset is given in Appendix~\ref{appendix:dataset_overview}.

\subsection{Performance predictor model}
\label{sec:performance_predictor_model_result}
In total, we developed 480 model instances during the hyperparameter tuning for both the high and low-dimensional performance predictors, $\psi^\text{high}$ and $\psi^\text{low}$.
The models were evaluated using three metrics: mean relative error (MRE), training time, and inference speed. The MRE is relative to the simulation ground truth and is calculated from the average of five distinct training results.
The training time was counted as the time per epoch (number of epochs ranging up to about $2,000$), calculating the median over the entire training.
The inference speed is measured as the model execution time, taking the average of 1,000 model executions.

We selected the MLPs with two hidden layers with $2,048$ units as models of special interest. 
These are denoted $\psi^\text{high}_{\diamond}$ and $\psi^\text{high}_{\star}$, where the difference is the use of Swish and Leaky ReLU for activation functions. 
These models have roughly $10^7$ model parameters, including the convolutional layers. 
The loading performance MRE, listed in Table \ref{tab:extreme_models}, ranges between 3.5 and 7\,\% with insignificant dependence on the activation function.
For the low-dimensional model, the selected models of special interest, $\psi^\text{low}_{\diamond}$ and $\psi^\text{low}_{\star}$, are smaller with two hidden layers and $512$ units per layer.
This amounts to roughly $5\times10^5$ parameters (no convolutional network is involved).
It was trained on a smaller dataset with only 3,000 samples, as adding additional samples did not significantly improve the performance.
The loading performance MRE for the low-dimensional model saturates in the range of about 5.5\--8.5\,\%, in other words, with 20\--70\% larger errors than for the high-dimensional model.

The training time and inference time are found in Table \ref{tab:extreme_models}. The low-dimensional model trains about seven times faster, and inference runs three times faster than the selected high-dimensional model. The computational overhead of using Swish over Leaky ReLU is marginal.
The effect of hyperparameters on the model performance is found in Appendix~\ref{appendix:hyper_parameters}.

\begin{table} [!htb]
    \centering
    {\small
    \caption{Key properties of the selected performance predictor models.}
    \label{tab:extreme_models}
    \begin{tabular}{r c c c c}
    \dtoprule
    & \multicolumn{2}{c}{$\psi^\text{high}$} & \multicolumn{2}{c}{$\psi^\text{low}$} \\
    \cmidrule(r){2-3} \cmidrule(r){4-5}
    & L ReLU & Swish & L ReLU & Swish \\ \dtoprule
    mass MRE [\%] & $\bm{4.47}$ & $\bm{5.70}$ & 7.77 & 7.58 \\
    time MRE [\%] & $\bm{3.61}$ & $\bm{4.43}$ & 5.66 & 5.41 \\
    work MRE [\%] & $\bm{5.90}$ & $\bm{6.98}$ & 8.51 & 8.47 \\
    \cmidrule(r){1-5}
    training time [s] & 1.9 & 1.92 & $\bm{0.27}$ & $\bm{0.27}$ \\
    inference time [ms] & 1.13 & 1.27 & $\bm{0.31}$ & $\bm{0.36}$ \\
   \dbottomrule
    \end{tabular}
    }
\end{table}

\begin{figure} [!htb]
     \centering
    \includegraphics[clip,trim=0 0 0 0, width=1.0\textwidth]{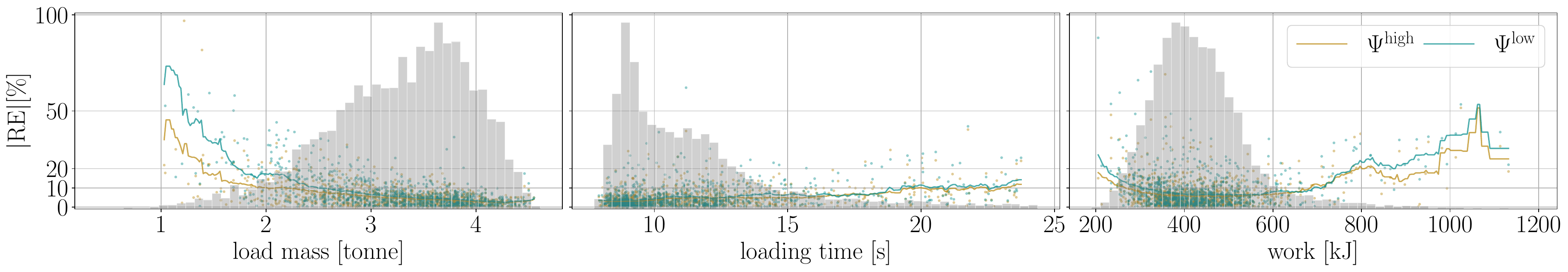}
    \caption{The distribution of relative errors for the performance predictor visualized by scatter plots and moving averages (curves).
    The gray histograms show the data distribution.}
    \label{fig:error_distribution}
\end{figure}

\begin{figure} [!htb]
    \centering
    \includegraphics[clip,trim=270 80 250 130, width=0.18\textwidth]{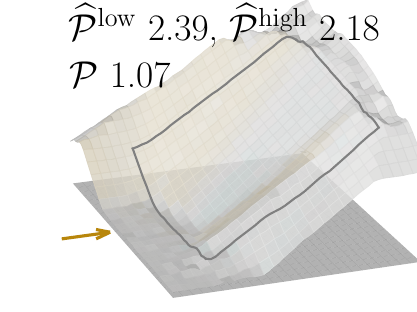}
    \includegraphics[clip,trim=270 80 250 130, width=0.18\textwidth]{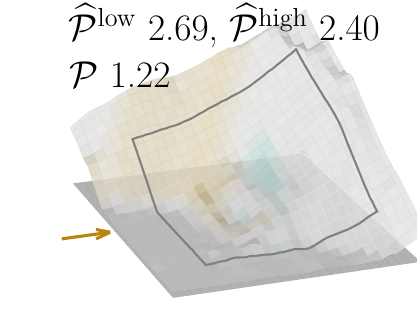}
    \includegraphics[clip,trim=270 80 250 130, width=0.18\textwidth]{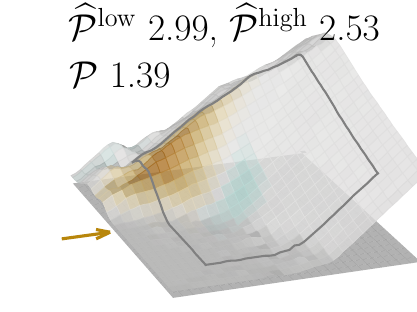}
    \includegraphics[clip,trim=270 80 250 130, width=0.18\textwidth]{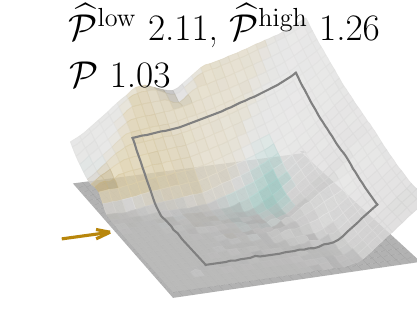}
    \includegraphics[clip,trim=270 80 250 130, width=0.18\textwidth]{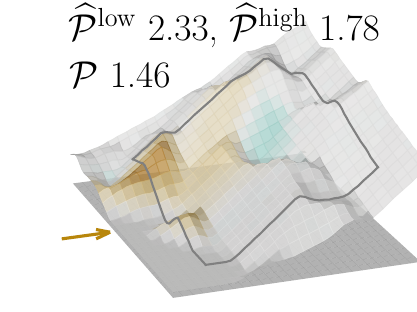} \\
    \includegraphics[clip,trim=270 80 250 130, width=0.18\textwidth]{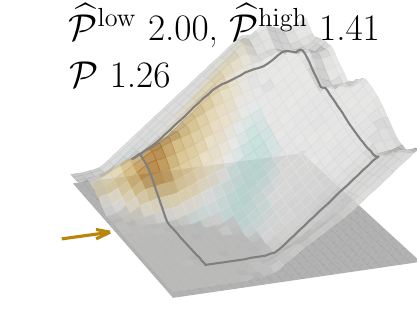}
    \includegraphics[clip,trim=270 80 250 130, width=0.18\textwidth]{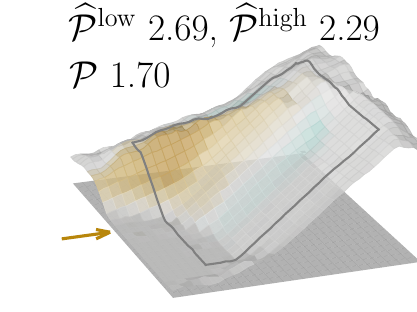}
    \includegraphics[clip,trim=270 80 250 130, width=0.18\textwidth]{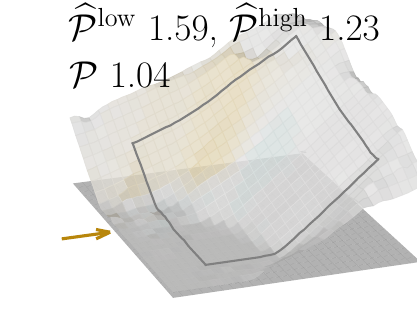}
    \includegraphics[clip,trim=270 80 250 130, width=0.18\textwidth]{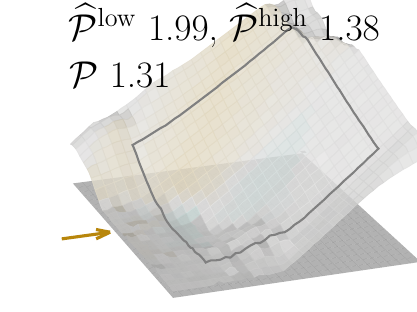}
    \includegraphics[clip,trim=270 80 250 130, width=0.18\textwidth]{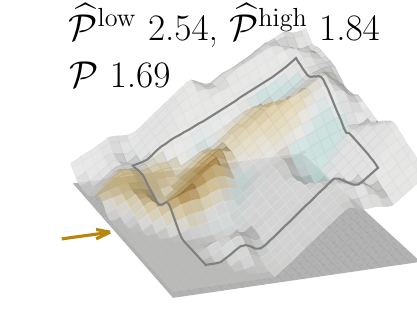}
    \caption{The ten worst load mass predictions of $\psi^\text{low}_{\diamond}$.
        The listed values are the errors for the low and high-dimensional models and the ground truth load mass.
        The heightmap of the initial pile state is shown with the grid cells color-coded by the decrease (gold) or increase (cyan) after loading. Some of the heightfields have been mirrored for illustrative purposes.}
    \label{fig:worst10}
\end{figure}

It is interesting to see how the errors are distributed in the space of the predictions (Fig.~\ref{fig:error_distribution}).
We observe that the relative error (RE) remains comparatively small for high load masses while it increases for the smallest load masses.
This suggests that the model is more reliable for high-performing loading actions, which is the general goal, than for low-performing ones.
It is interesting to understand why the model fails occasionally.
Therefore, we identified the test samples with the ten worst load mass predictions.
These are displayed in Fig.~\ref{fig:worst10} with the local heightmaps.
The common factors are a) that the loaded mass is small in the ground truth samples while overestimated by the model
and b) that the heightmap is skewed, with most mass distributed on either the left or right side.
It is understandable that the low-dimensional model, with only four parameters characterizing the
pile state, has more difficulty with these piles as it cannot distinguish between uniform and irregular pile surfaces.
That makes accurate load mass prediction difficult for complex pile surfaces.
%
%

\subsection{Pile state predictor model}
\label{sec:pile_state_predictor_model_result}
\begin{figure} [!htb]
    \centering
    \begin{subfigure}{0.49\textwidth}
        \centering
        \scriptsize{\textbf{(a)}}\\
        \includegraphics[clip,trim=0 0 371 0, width=0.6\textwidth]{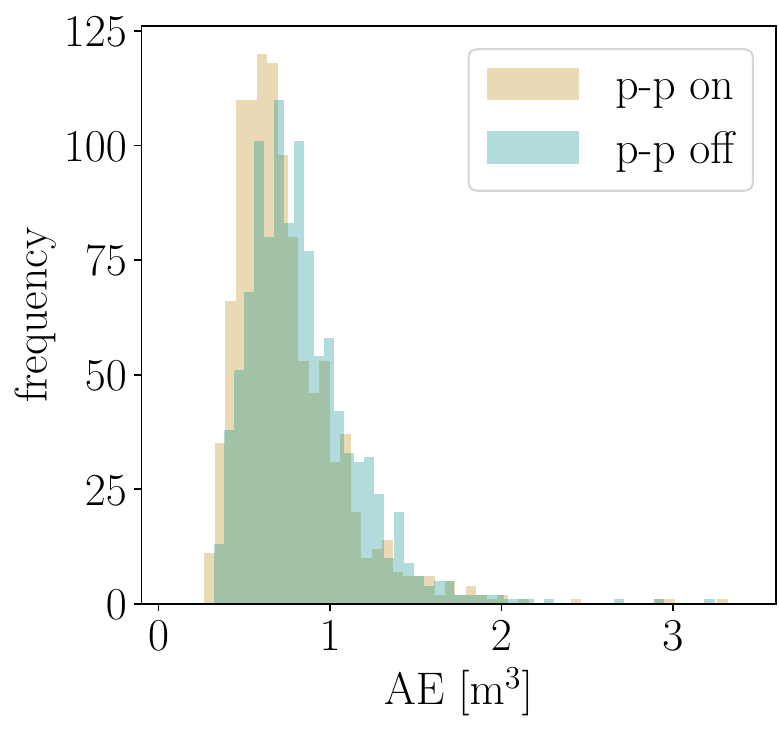}
				  
        \label{fig:vaemlp_error_1}
    \end{subfigure}
    \begin{subfigure}{0.49\textwidth}
        \centering
        \scriptsize{\textbf{(b)}}\\
        \includegraphics[clip,trim=374 0 0 0, width=0.6\textwidth]{fig/hist_pred_error_vol_n_plot_error_vol_vs_error_vol_pp.pdf}
        \label{fig:vaemlp_error_2}
    \end{subfigure}
    \caption{The distribution of the prediction test error (a) and AE correlation plot with and without post-processing (p-p) (b). The five points of interest, marked by diamonds ($\diamond$) in (b), are visualized in Fig.~\ref{fig:pile_state_predictions}.}
    \label{fig:vaemlp_error}
\end{figure}

\begin{figure} [!htb]
    \scriptsize{
         \begin{subfigure}{1.0\textwidth}
                 \centering
                 \stackinset{c}{}{t}{-.15in}{$\bm{h}$}{
                 \includegraphics[clip,trim=95 60 80 130, width=0.175\textwidth]{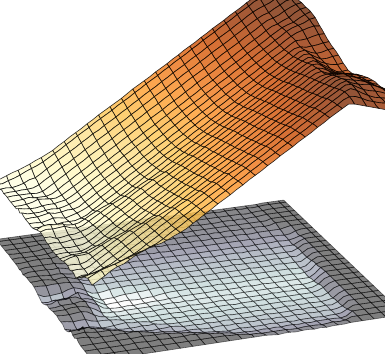}}
                 \stackinset{c}{}{t}{-.15in}{reconstr. $\bm{h}$}{
                 \includegraphics[clip,trim=95 60 80 130, width=0.175\textwidth]{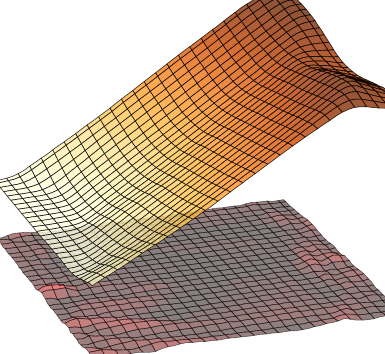}}
                 \stackinset{c}{}{t}{-.15in}{$\bm{h}'$}{
                 \includegraphics[clip,trim=95 60 80 130, width=0.175\textwidth]{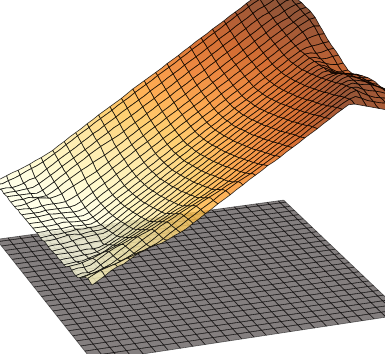}}
                 \stackinset{c}{}{t}{-.15in}{$\hat{\bm{h}}'$ p-p off}{
                 \includegraphics[clip,trim=95 60 80 130, width=0.175\textwidth]{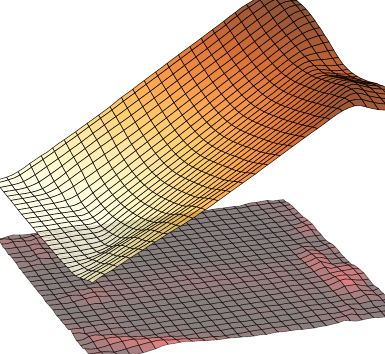}}
                 \stackinset{c}{}{t}{-.15in}{$\hat{\bm{h}}'$ with p-p on}{
                 \includegraphics[clip,trim=95 60 80 130, width=0.175\textwidth]{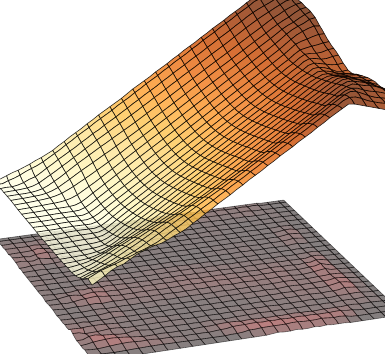}}
                 \vspace{-2mm}
                 \caption*{\hspace{10mm}\scriptsize{(a) test data \#288, AE (p-p on):0.28, AE (p-p off):0.38.}}
         \end{subfigure}
         \begin{subfigure}{1.0\textwidth}
                 \centering
                 \includegraphics[clip,trim=95 60 80 130, width=0.18\textwidth]{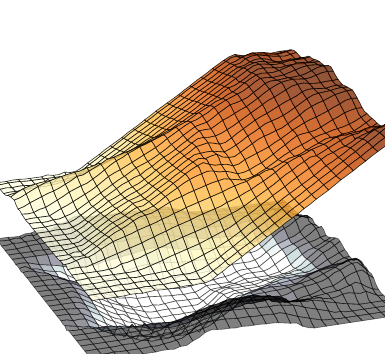}
                 \includegraphics[clip,trim=95 60 80 130, width=0.18\textwidth]{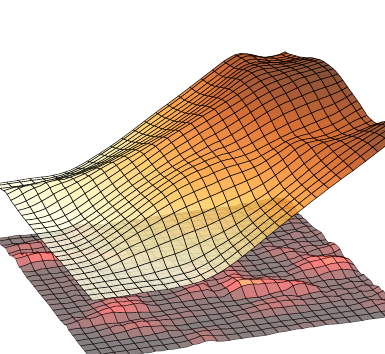}
                 \includegraphics[clip,trim=95 60 80 130, width=0.18\textwidth]{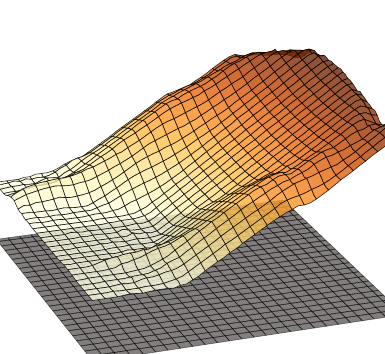}
                 \includegraphics[clip,trim=95 60 80 130, width=0.18\textwidth]{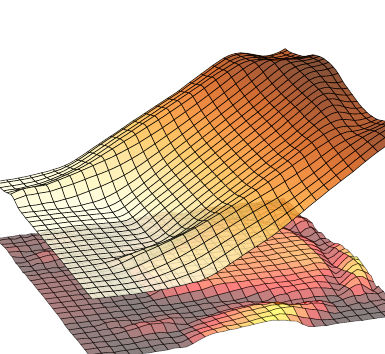}
                 \includegraphics[clip,trim=95 60 80 130, width=0.18\textwidth]{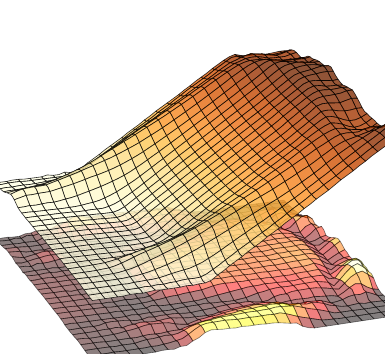}
                 \vspace{-2mm}
                 \caption*{\hspace{10mm}\scriptsize{(b) test data \#646, AE (p-p on):2.13, AE (p-p off):1.92.}}
         \end{subfigure}
         \begin{subfigure}{1.0\textwidth}
                 \centering
                 \includegraphics[clip,trim=95 60 80 130, width=0.18\textwidth]{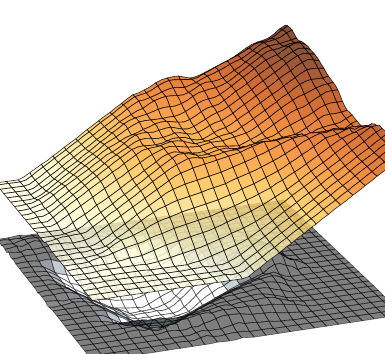}
                 \includegraphics[clip,trim=95 60 80 130, width=0.18\textwidth]{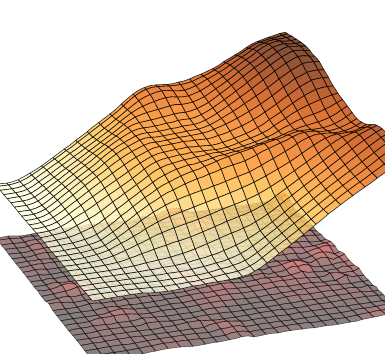}
                 \includegraphics[clip,trim=95 60 80 130, width=0.18\textwidth]{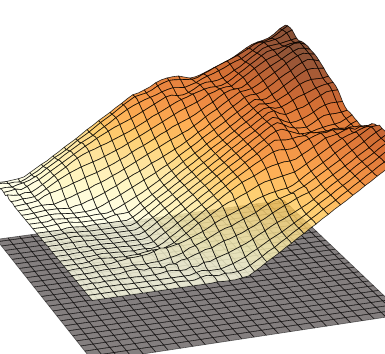}
                 \includegraphics[clip,trim=95 60 80 130, width=0.18\textwidth]{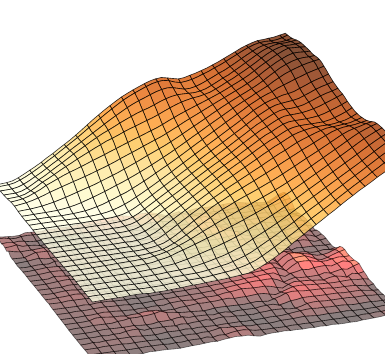}
                 \includegraphics[clip,trim=95 60 80 130, width=0.18\textwidth]{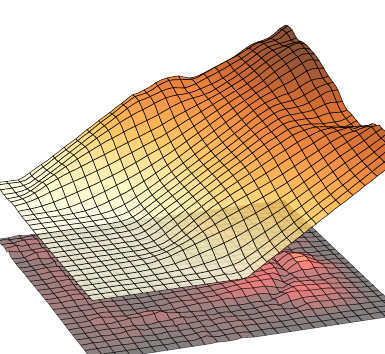}
                 \vspace{-2mm}
                 \caption*{\hspace{10mm}\scriptsize{(c) test data \#388, AE (p-p on):0.68, AE (p-p off):0.88.}}
         \end{subfigure}
         \begin{subfigure}{1.0\textwidth}
                 \centering
                 \includegraphics[clip,trim=95 60 80 130, width=0.18\textwidth]{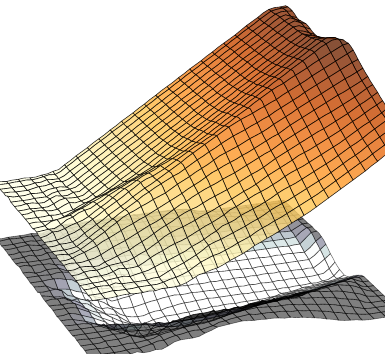}
                 \includegraphics[clip,trim=95 60 80 130, width=0.18\textwidth]{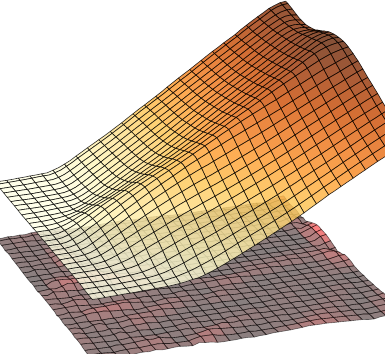}
                 \includegraphics[clip,trim=95 60 80 130, width=0.18\textwidth]{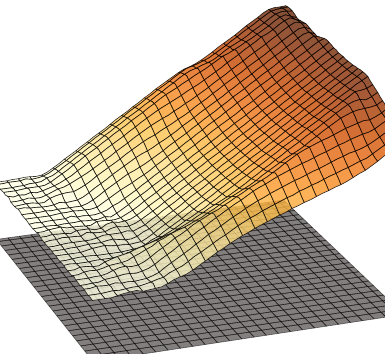}
                 \includegraphics[clip,trim=95 60 80 130, width=0.18\textwidth]{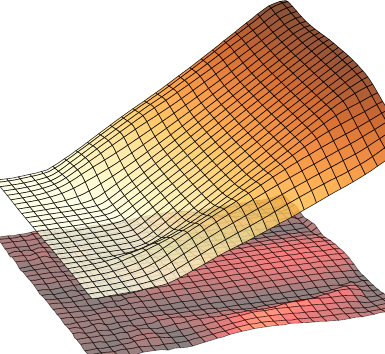}
                 \includegraphics[clip,trim=95 60 80 130, width=0.18\textwidth]{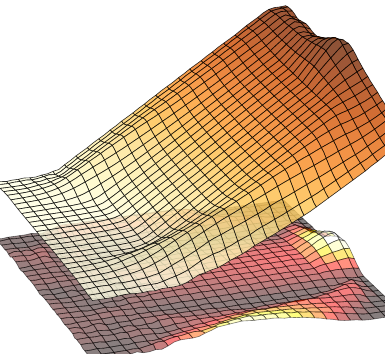}
                 \vspace{-2mm}
                 \caption*{\hspace{10mm}\scriptsize{(d) test data \#142, AE (p-p on):1.57, AE (p-p off):1.22.}}
                 \label{fig:pile_state_predictions_d}
         \end{subfigure}
         \begin{subfigure}{1.0\textwidth}
                 \centering
                 \includegraphics[clip,trim=95 60 80 130, width=0.18\textwidth]{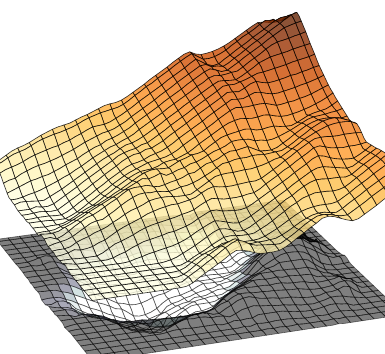}
                 \includegraphics[clip,trim=95 60 80 130, width=0.18\textwidth]{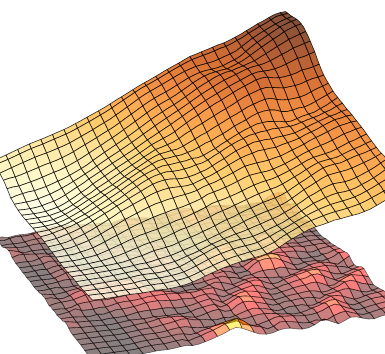}
                 \includegraphics[clip,trim=95 60 80 130, width=0.18\textwidth]{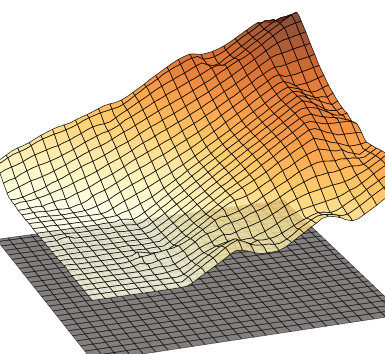}
                 \includegraphics[clip,trim=95 60 80 130, width=0.18\textwidth]{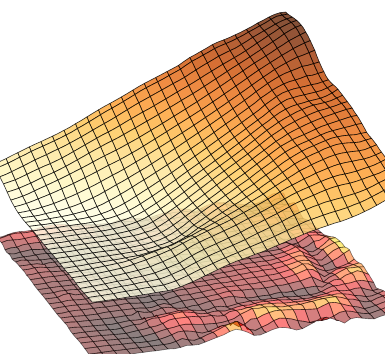}
                 \includegraphics[clip,trim=95 60 80 130, width=0.18\textwidth]{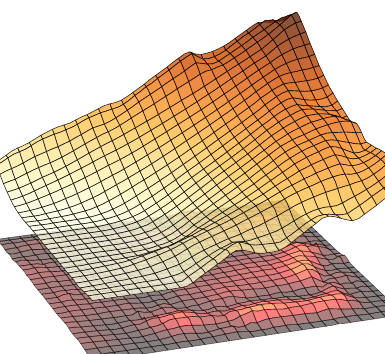}
                 \vspace{-2mm}
                 \caption*{\hspace{10mm}\scriptsize{(e) test data \#1034, AE (p-p on):0.98, AE (p-p off):1.45.}}
                 \vspace{-2mm}
                 \label{fig:pile_state_predictions_e}
         \end{subfigure}
    }
    \caption{Five selected samples of initial pile state $\bm{h}$ and the corresponding resulting ground truth pile state $\bm{h}'$ and model predictions $\hat{\bm{h}}'$, with and without post-processing. The VAE reconstruction capability is included in the second column.
    Samples (a) have small errors in the test data set, (b) have large errors, and (c) are close to the median.
    In samples (d) and (e), the post-processing increases and decreases the errors, respectively.
    The surface under the pile shows the absolute difference compared to the ground truth final pile state, except that only in the second column is it compared to the initial pile state.}
    \label{fig:pile_state_predictions}
\end{figure}

The developed pile state predictor model is evaluated by the mean absolute error (MAE) and mean relative error (MRE) of a prediction $\hat{\bm{h}}'$ compared to the simulated ground truth heightmap $\bm{h}'$.
MAE and MRE are calculated in terms of the volume difference of the enclosing surfaces. In detail, the MRE is computed as
\begin{equation}
    \text{MRE} = \cfrac{1}{N}\sum_n^N \varepsilon_n^\text{rel} = \cfrac{1}{N}\sum_n^N\cfrac{\sum_{i,j} \lvert \hat{h}_{ij}^{\prime\, n}-h_{ij}^{\prime\, n} \rvert \Delta l ^2}{\sum_{i,j} \lvert h_{ij}^{\prime\, n} \rvert \Delta l ^2},
\end{equation}
where the volume of each cell, indexed by $ij$, is calculated between the surface and the extended ground plane in the grid.
The MAE is calculated the same way, excluding the volume normalization.
The results are summarized in Table~\ref{tab:pile_state_model_accuracy}.
The MAE of misplaced volume constitutes roughly 25\,\% of the bucket capacity but only 3\,\% of the volume under the local heightmap.
The post-processing step (Sec.~\ref{ref:post-processing}) reduces the MAE from 0.84 to 0.75\,m$^3$ 
, with insignificant overhead in inference time. 
The improvement by the post-processing can also be seen from the distribution of the errors in Fig.~\ref{fig:vaemlp_error}.
%
\begin{table}[!htb]
    \centering
    {\small
    \caption{Test results of the pile state predictor model, with and without the post-processing (p-p).}
    \label{tab:pile_state_model_accuracy}
    \begin{tabular}{c c c c c} \hline
        p-p & MAE [m$^3$] & MRE [\%] & inference time [ms] \\ \hline
        on  & 0.75 & 3.04 & 4.51 \\
         off  & 0.84 & 3.41 & 4.48 \\ \hline
    \end{tabular}
    }
\end{table}
%
Fig.~\ref{fig:pile_state_predictions} visualizes five selected test predictions, marked by diamonds in Fig.~\ref{fig:vaemlp_error}(b).
Overall, the predicted pile surfaces look reasonable compared to the ground truth. We find that the model performance tends to be better when the initial pile has a smooth, regular shape, which is to be expected as the VAE decoder tends to produce smooth outputs.
Examples of the smoothness issue can be seen in the reconstruction column in Fig.~\ref{fig:pile_state_predictions}.

Post-processing is most beneficial in cases where the initial state is irregular and when there is little change at the boundary (Fig.~\ref{fig:pile_state_predictions_e}).
When the initial pile is smooth with a uniform slope close to the angle of repose, loading will induce avalanches that affect the pile state at the rear boundary. This explains why the AE is sometimes increased by post-processing (Fig.~\ref{fig:pile_state_predictions_d}).

\subsection{Sequential loading predictions}
\label{sec:sequential_loading-predictions_result}
\begin{figure} [!htb]
    {\scriptsize
	\medskip
	\begin{minipage}{0.33\textwidth}
            \begin{center}{\scriptsize (a) Ground truth}\end{center}\vspace{-3mm}
            \includegraphics[clip,trim=150 140 180 290, width=0.95\textwidth]{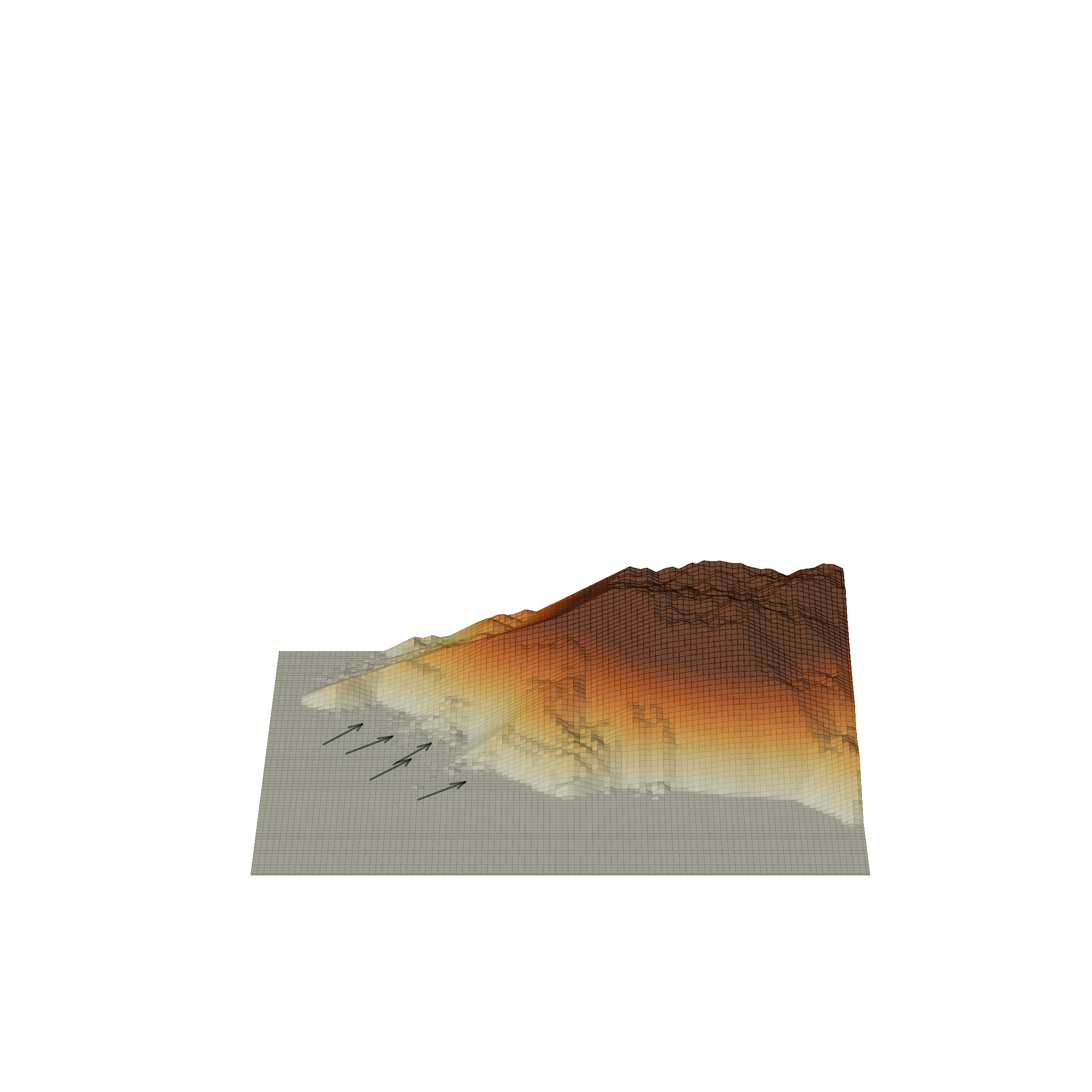} \\
            \includegraphics[clip,trim=150 140 180 290, width=0.95\textwidth]{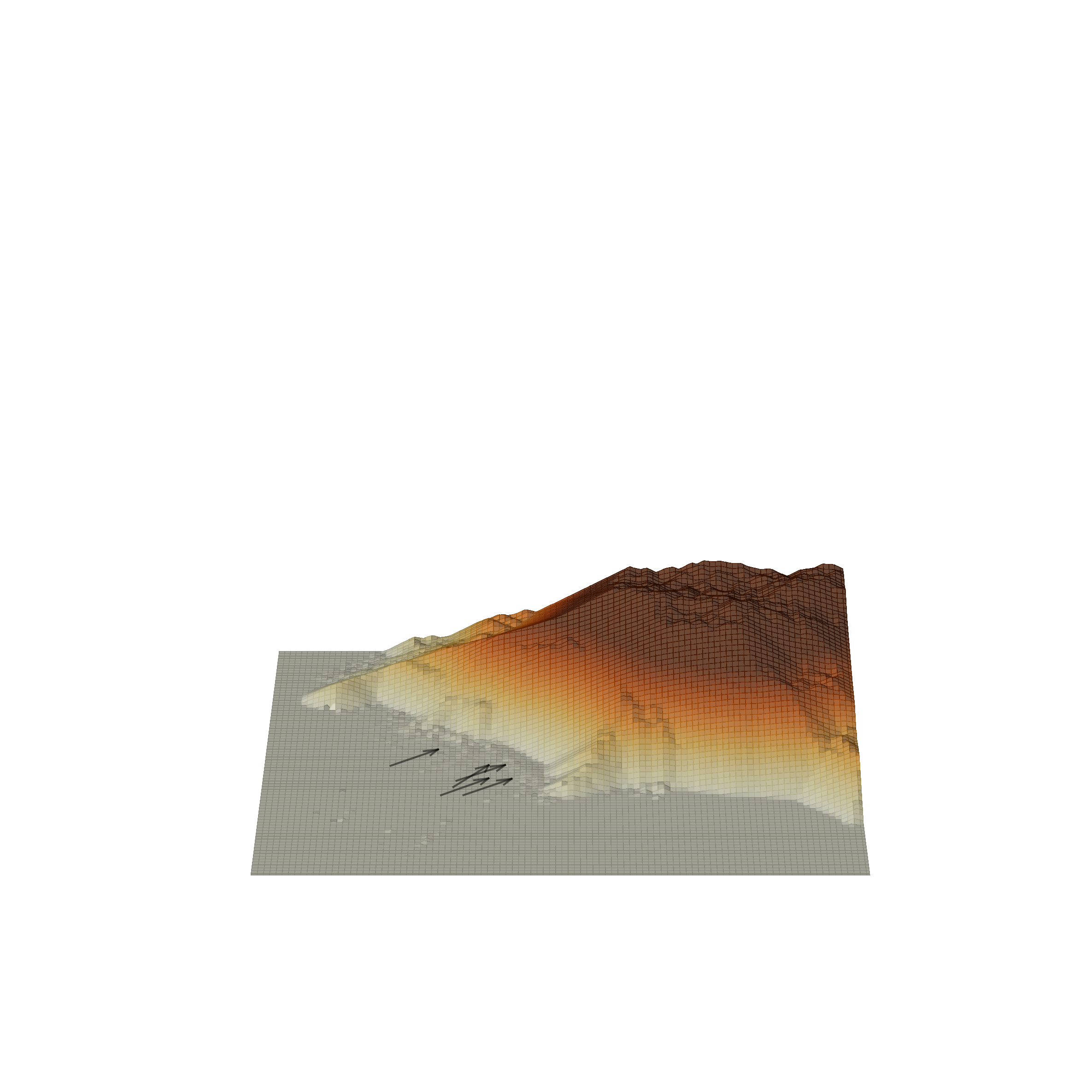}\\
            \includegraphics[clip,trim=150 140 180 290, width=0.95\textwidth]{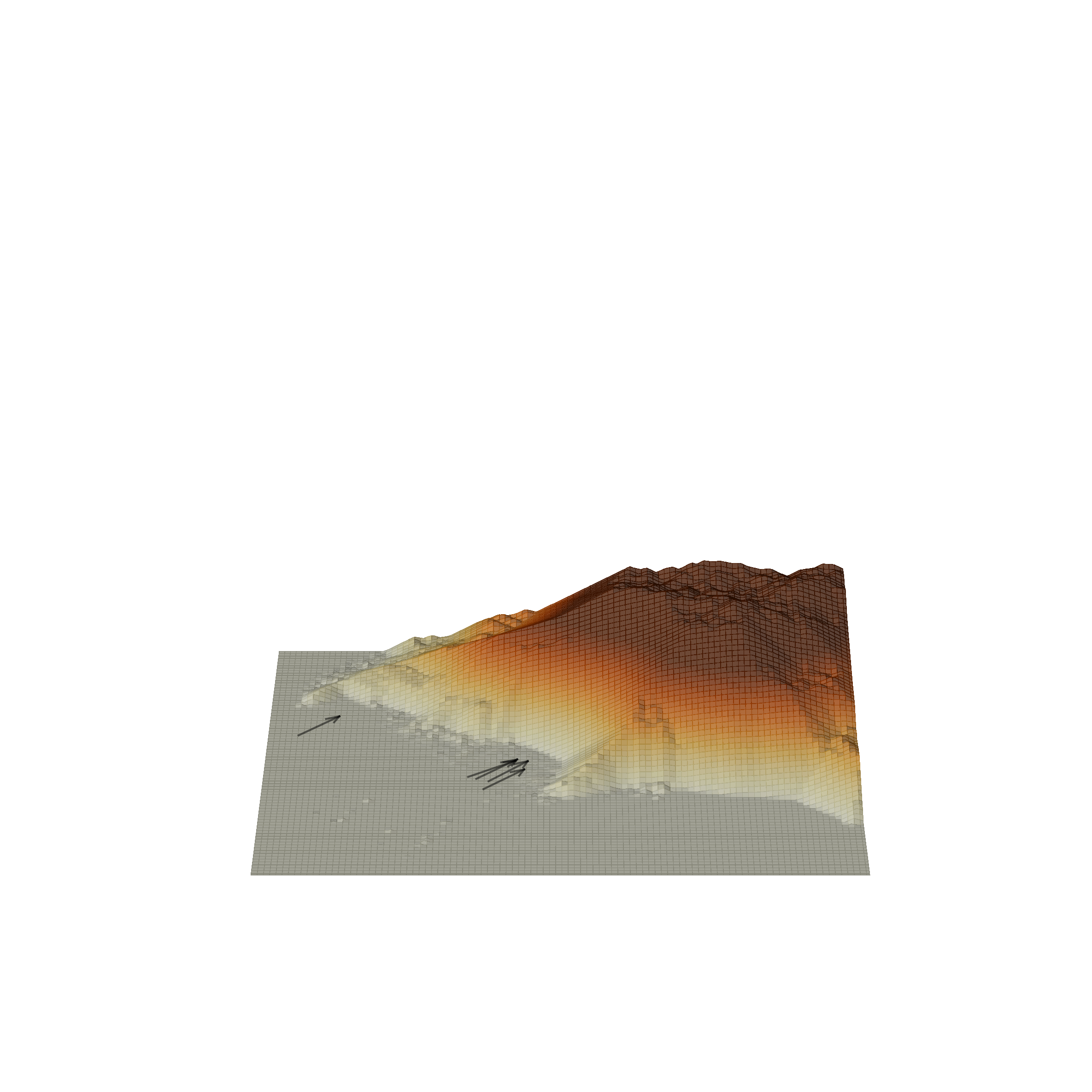}\\
            \includegraphics[clip,trim=150 140 180 290, width=0.95\textwidth]{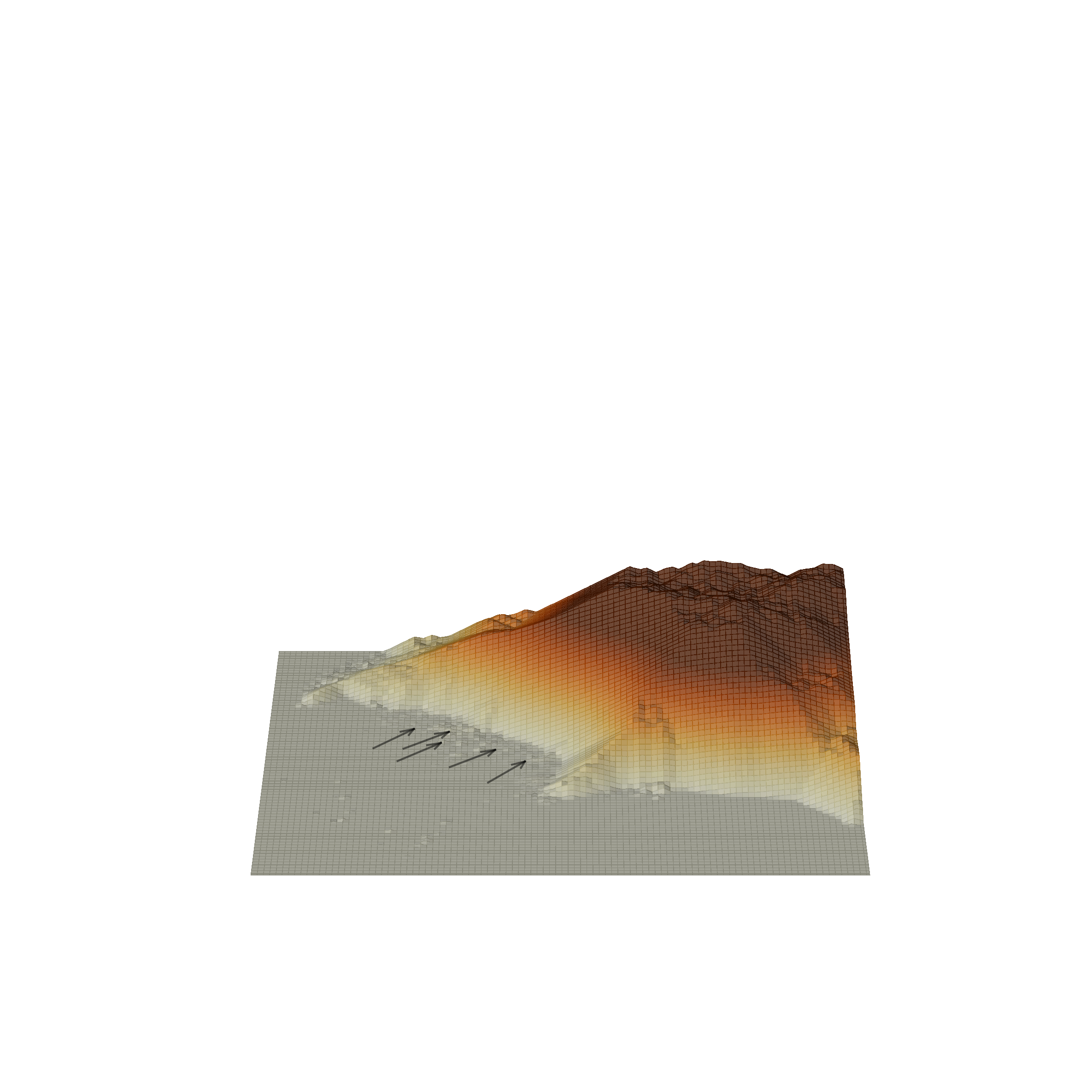}\\
            \includegraphics[clip,trim=150 140 180 290, width=0.95\textwidth]{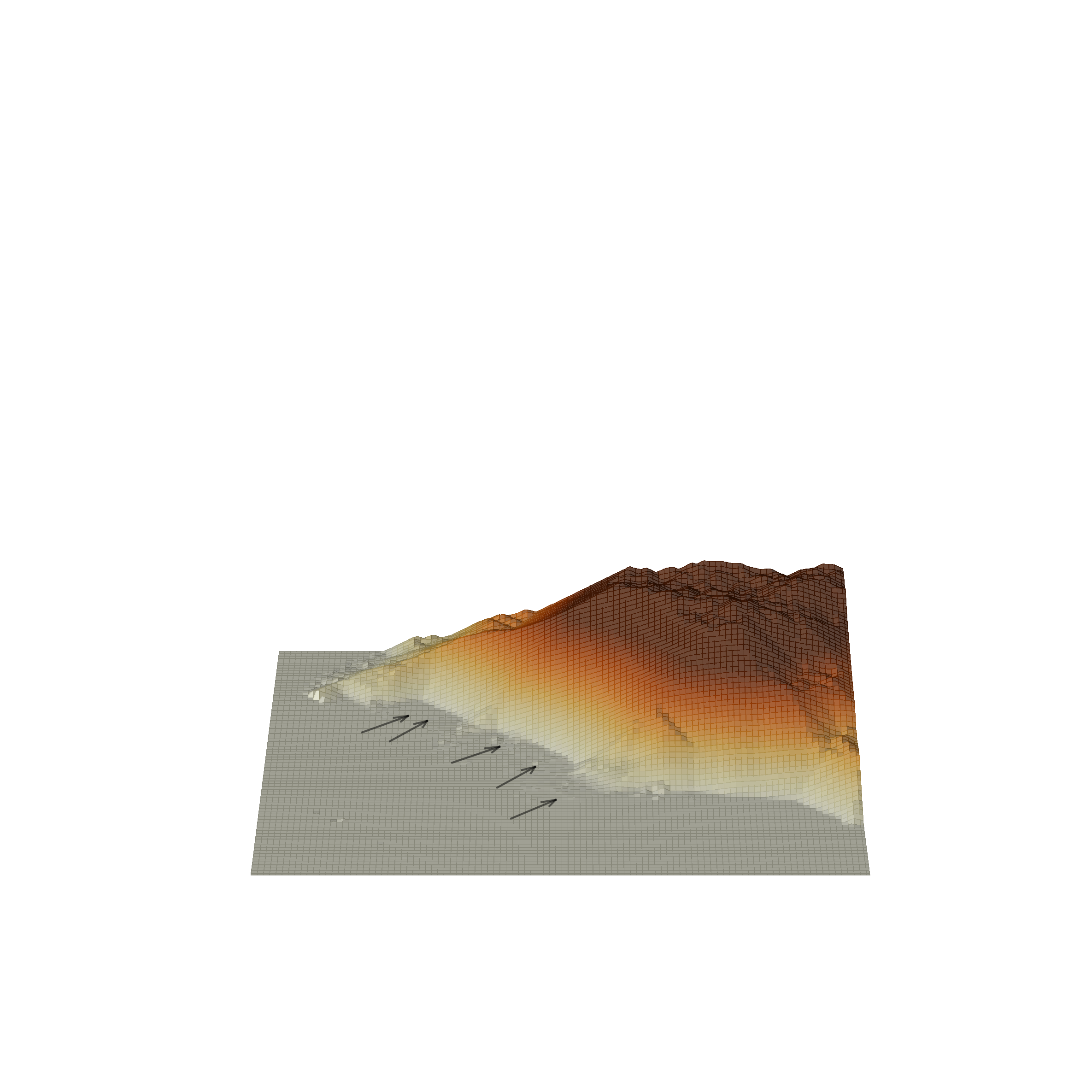}\\
            \includegraphics[clip,trim=150 140 180 290, width=0.95\textwidth]{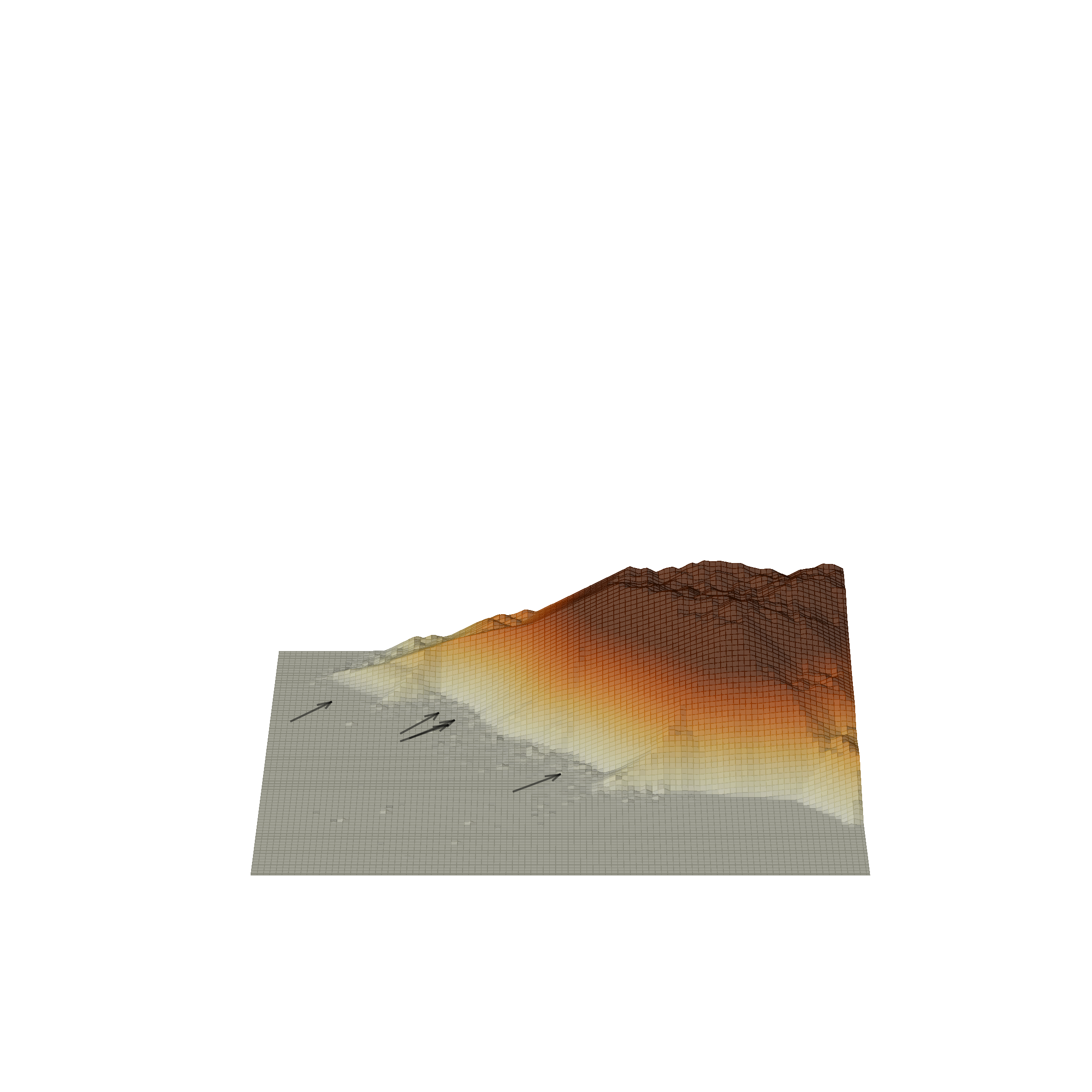}\\
            \includegraphics[clip,trim=150 140 180 290, width=0.95\textwidth]{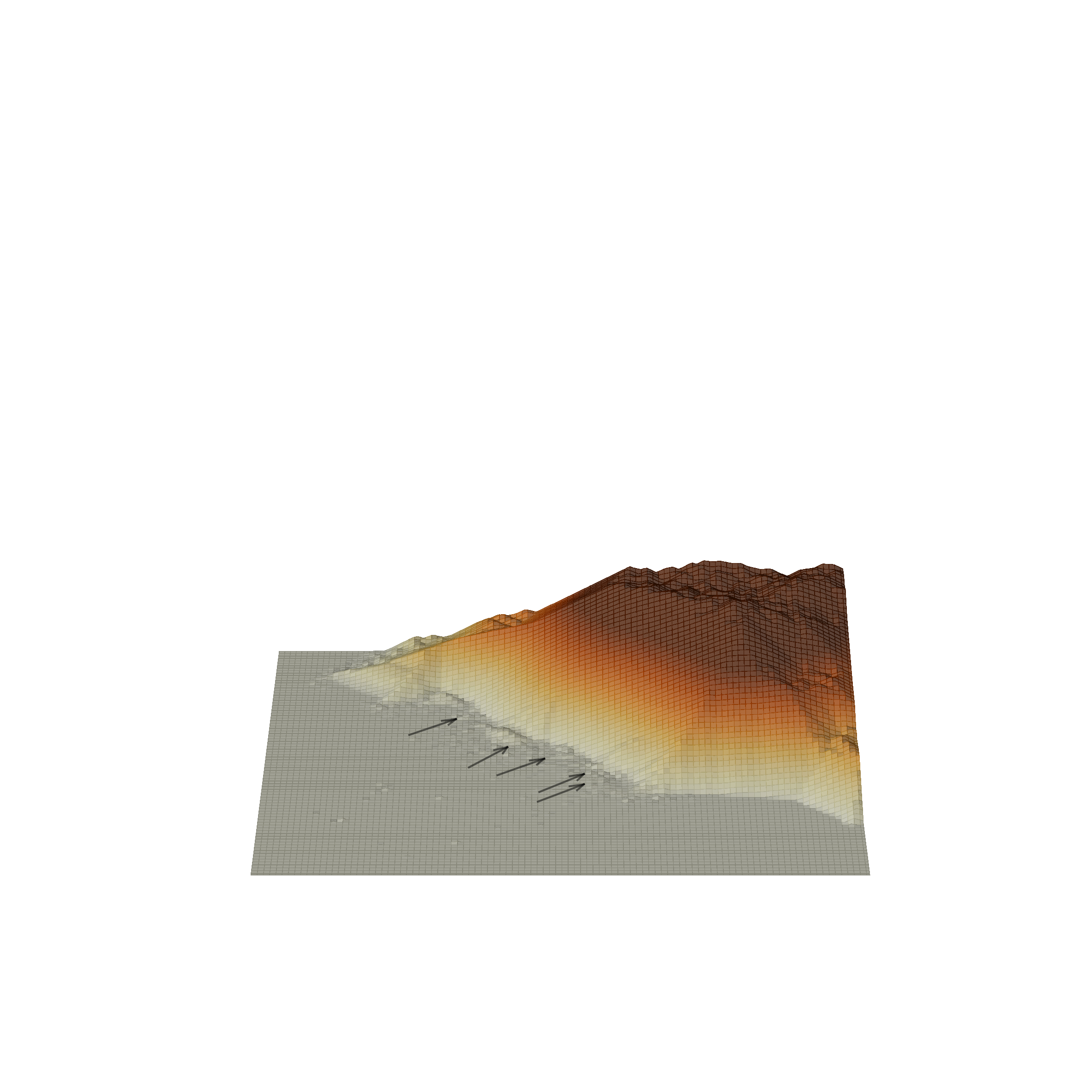}\\
            \includegraphics[clip,trim=150 140 180 290, width=0.95\textwidth]{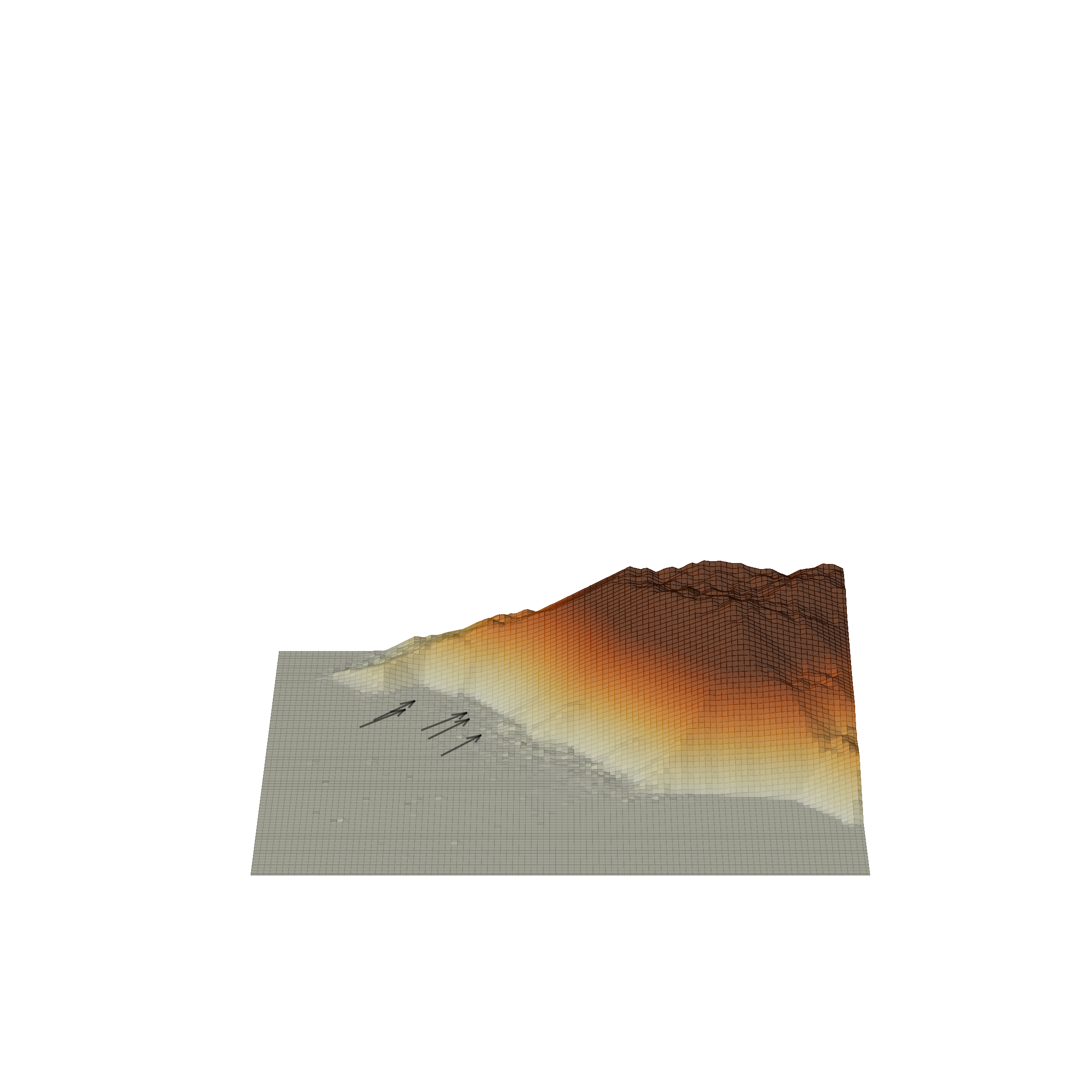}
	\end{minipage}\hfill
	\begin{minipage}{0.33\textwidth}
            \begin{center}{\scriptsize (b) High-dimensional model}\end{center}\vspace{-3mm}
            \includegraphics[clip,trim=150 140 180 290, width=0.95\textwidth]{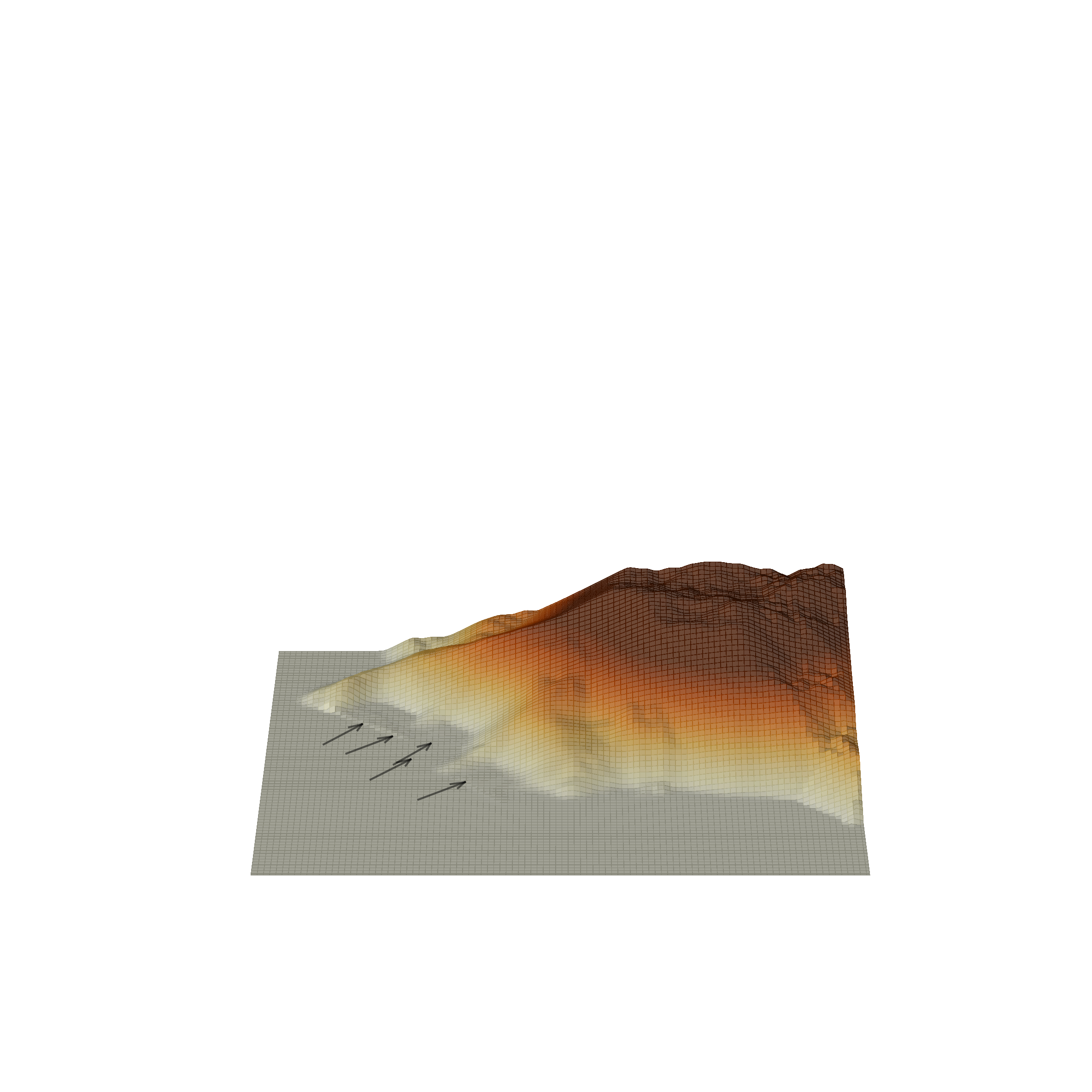}\\
            \includegraphics[clip,trim=150 140 180 290, width=0.95\textwidth]{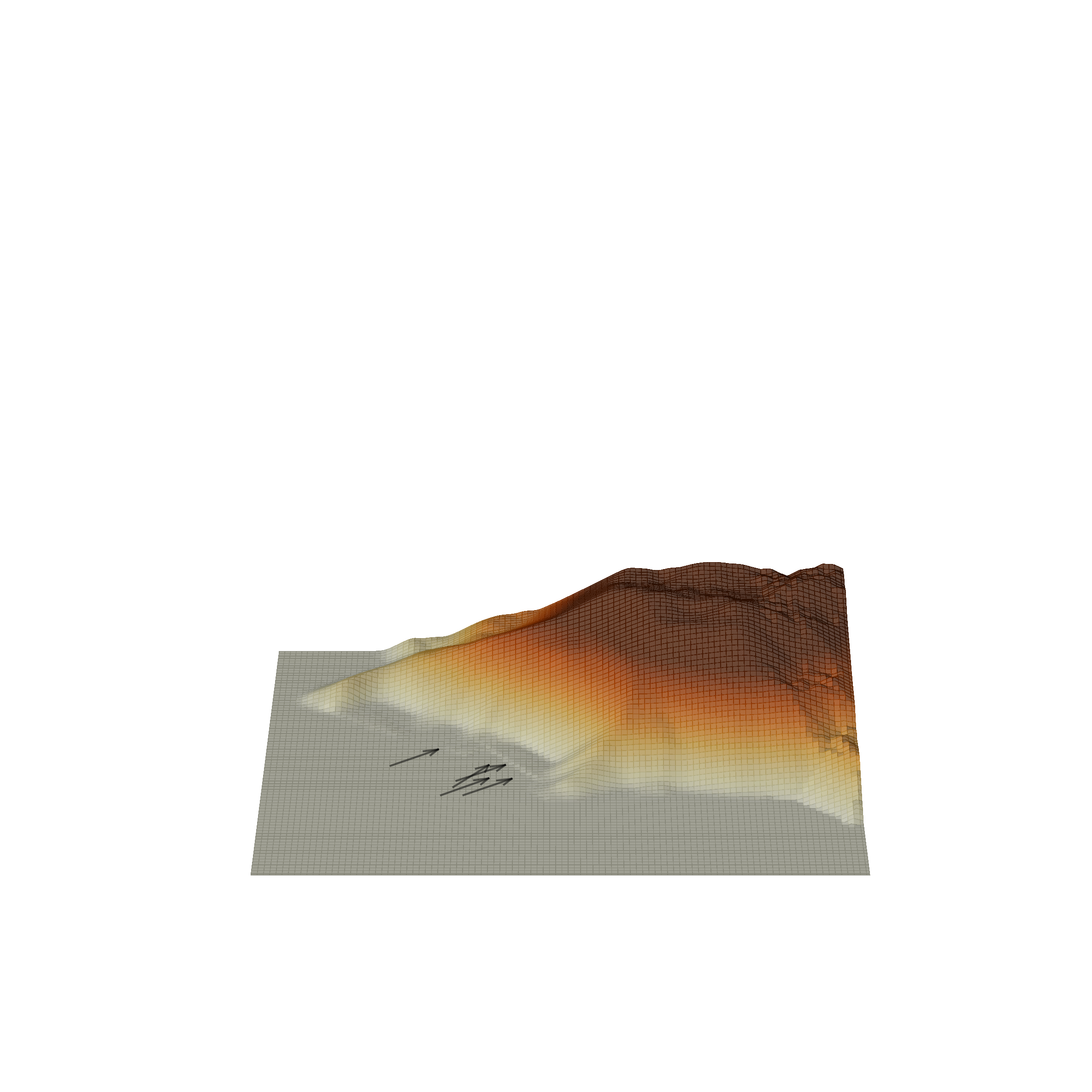}\\
            \includegraphics[clip,trim=150 140 180 290, width=0.95\textwidth]{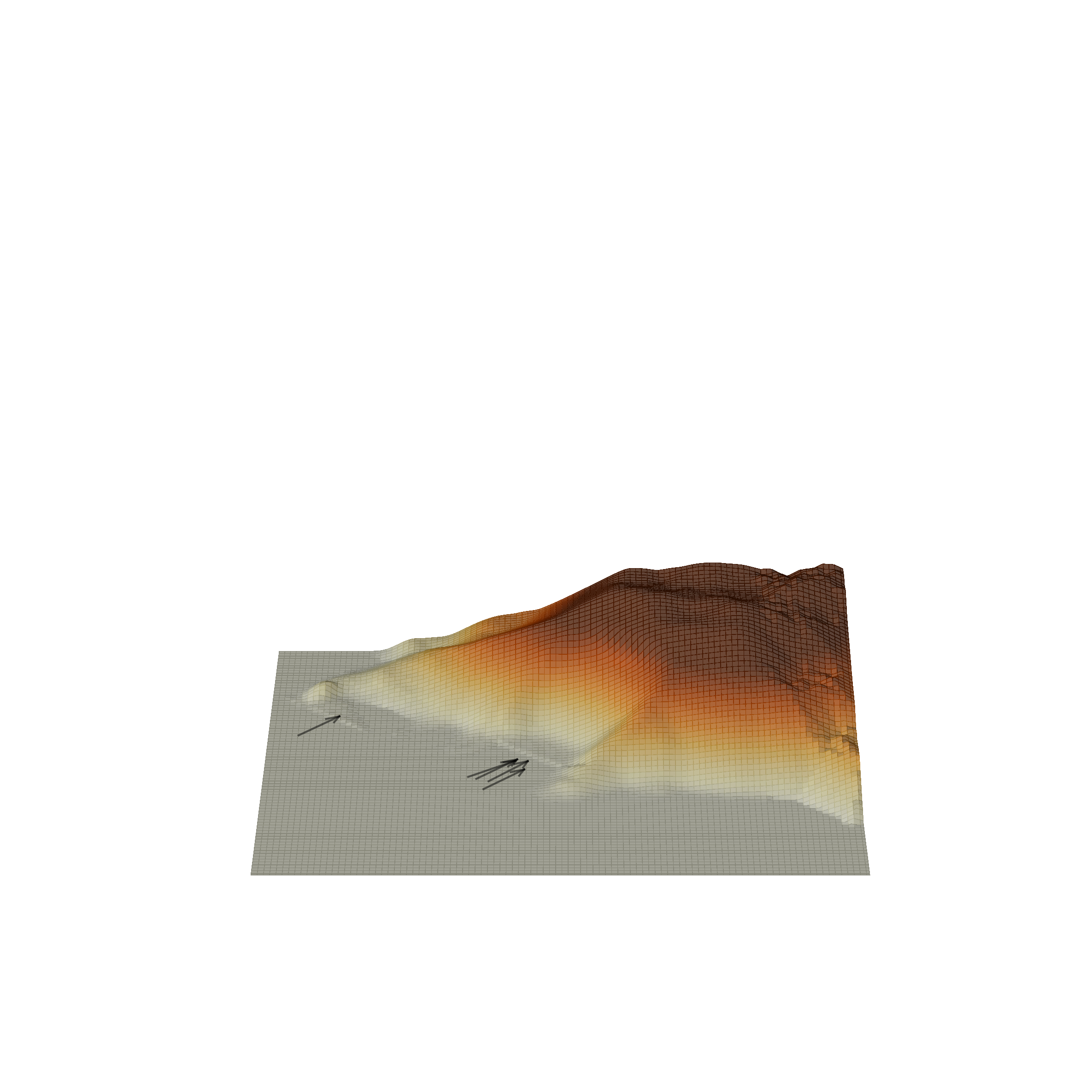}\\
            \includegraphics[clip,trim=150 140 180 290, width=0.95\textwidth]{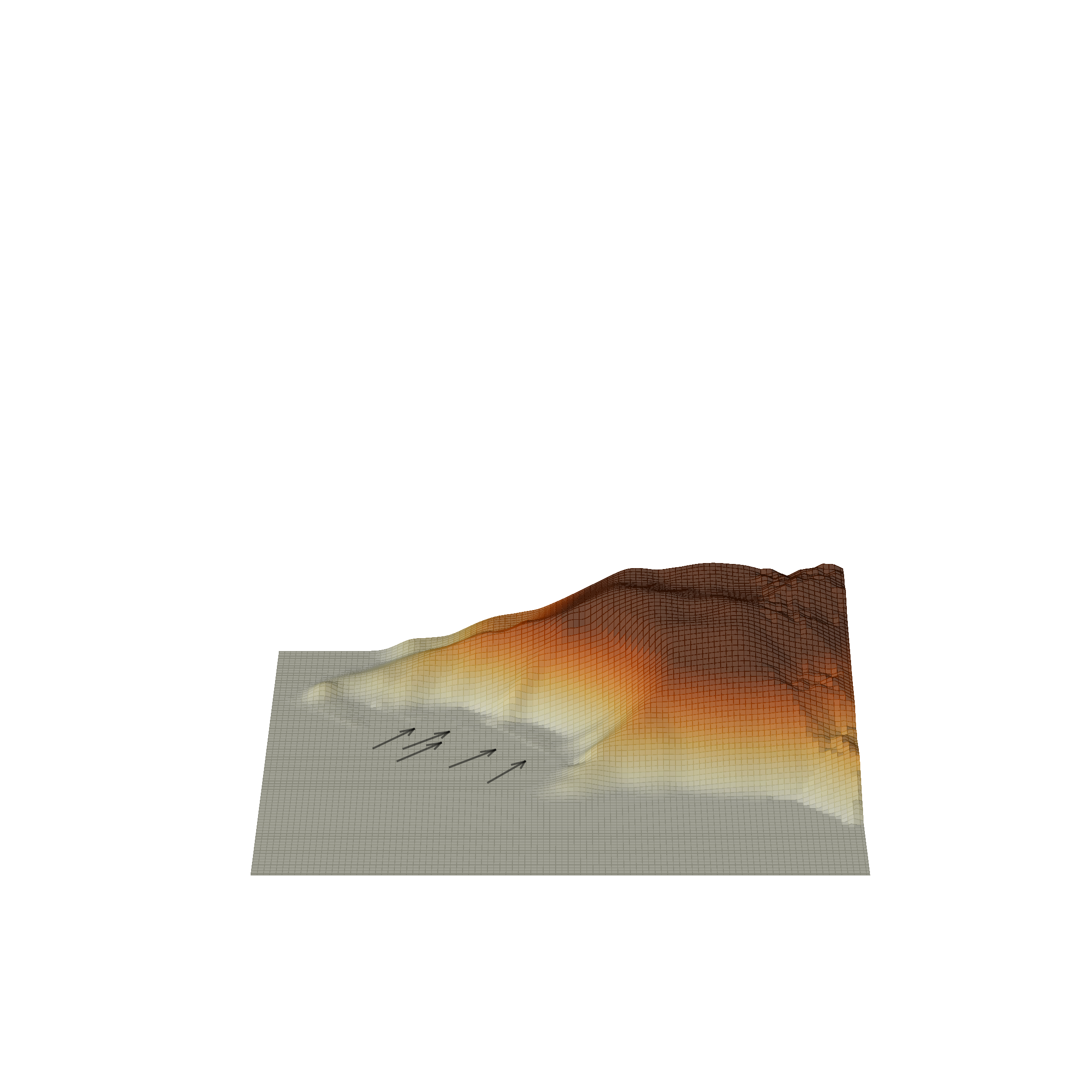}\\
            \includegraphics[clip,trim=150 140 180 290, width=0.95\textwidth]{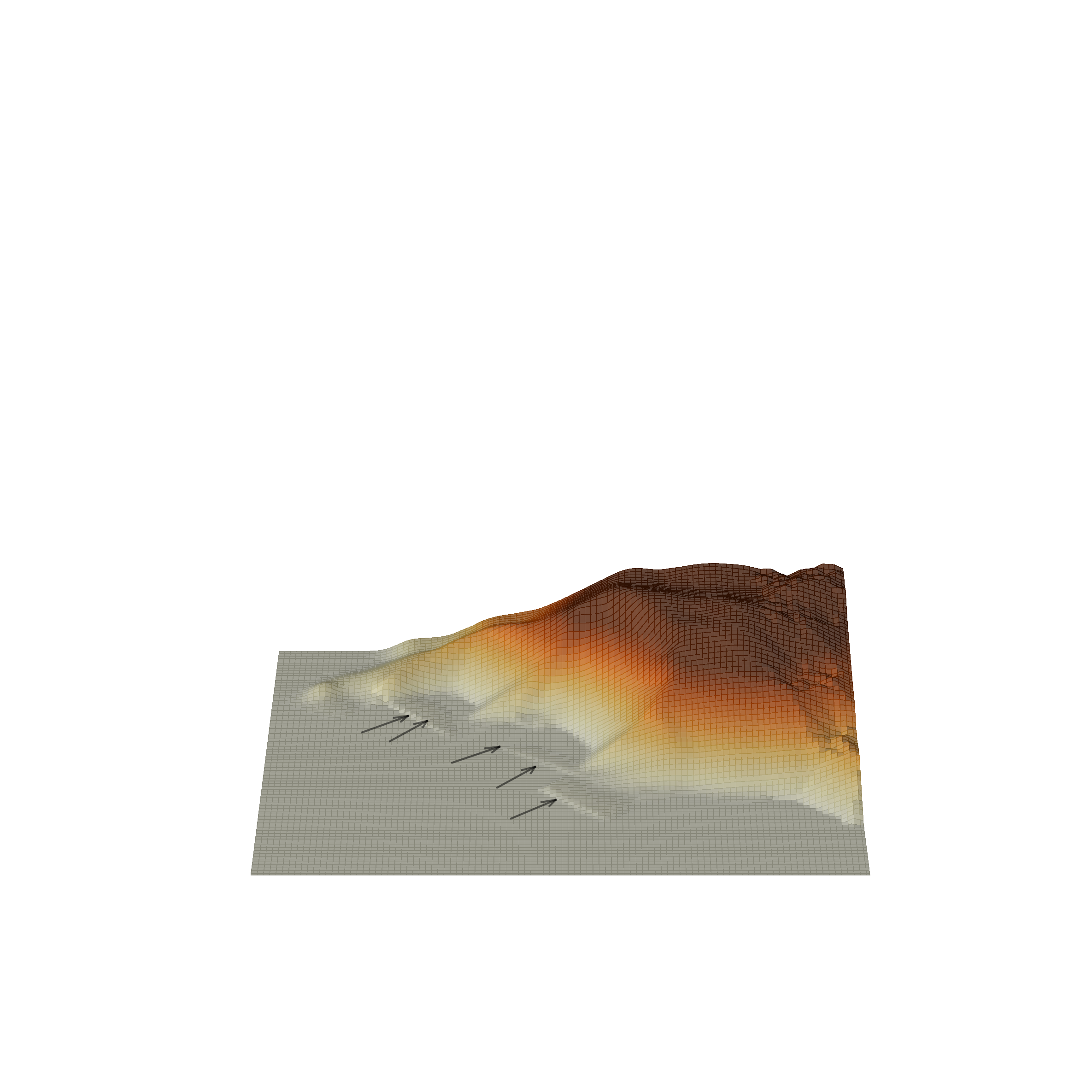}\\
            \includegraphics[clip,trim=150 140 180 290, width=0.95\textwidth]{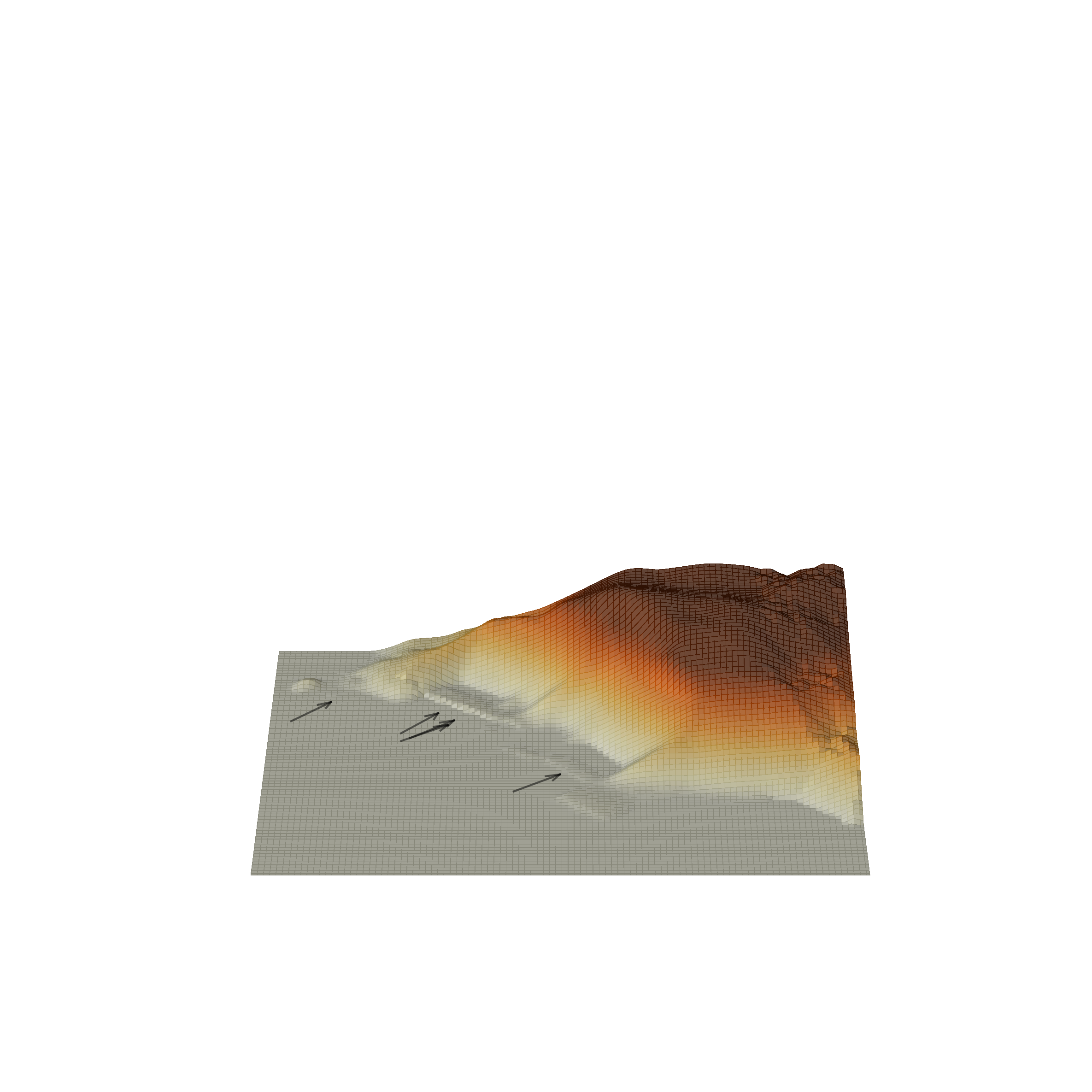}\\
            \includegraphics[clip,trim=150 140 180 290, width=0.95\textwidth]{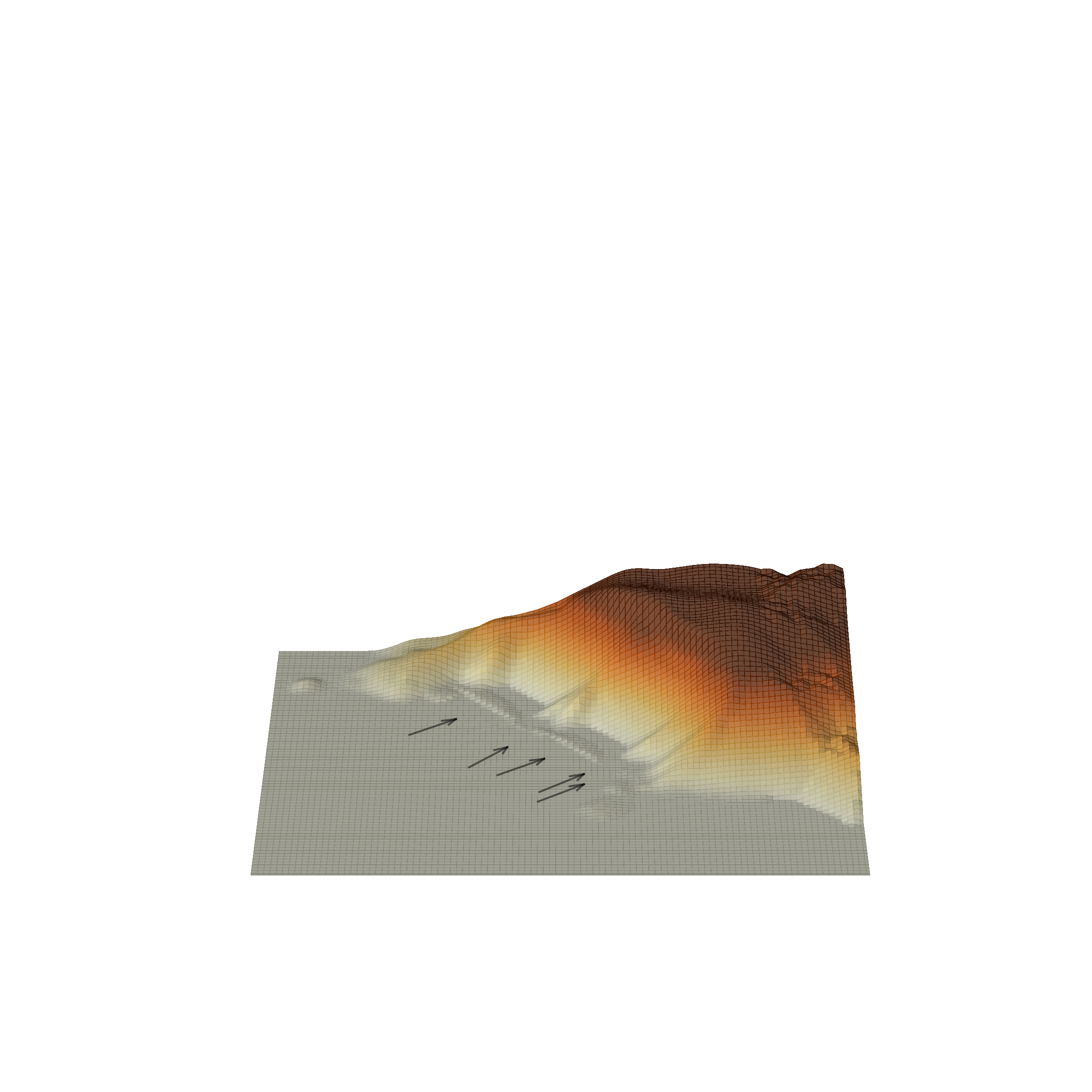}\\
            \includegraphics[clip,trim=150 140 180 290, width=0.95\textwidth]{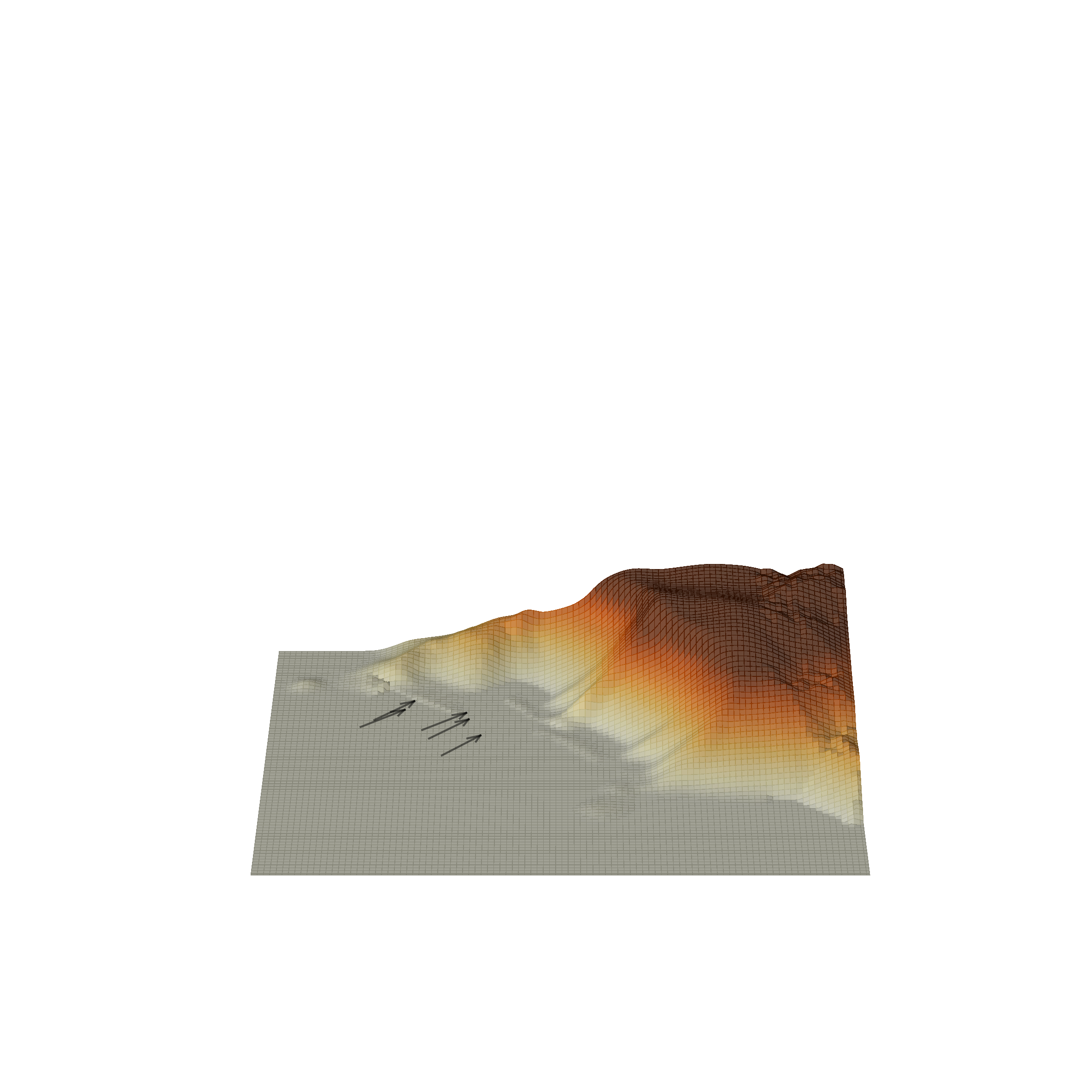}
	\end{minipage}\hfill
	\begin{minipage}{0.33\textwidth}
            \begin{center}{\scriptsize (c) Low-dimensional model}\end{center}\vspace{-3mm}
            \includegraphics[clip,trim=150 140 180 290, width=0.95\textwidth]{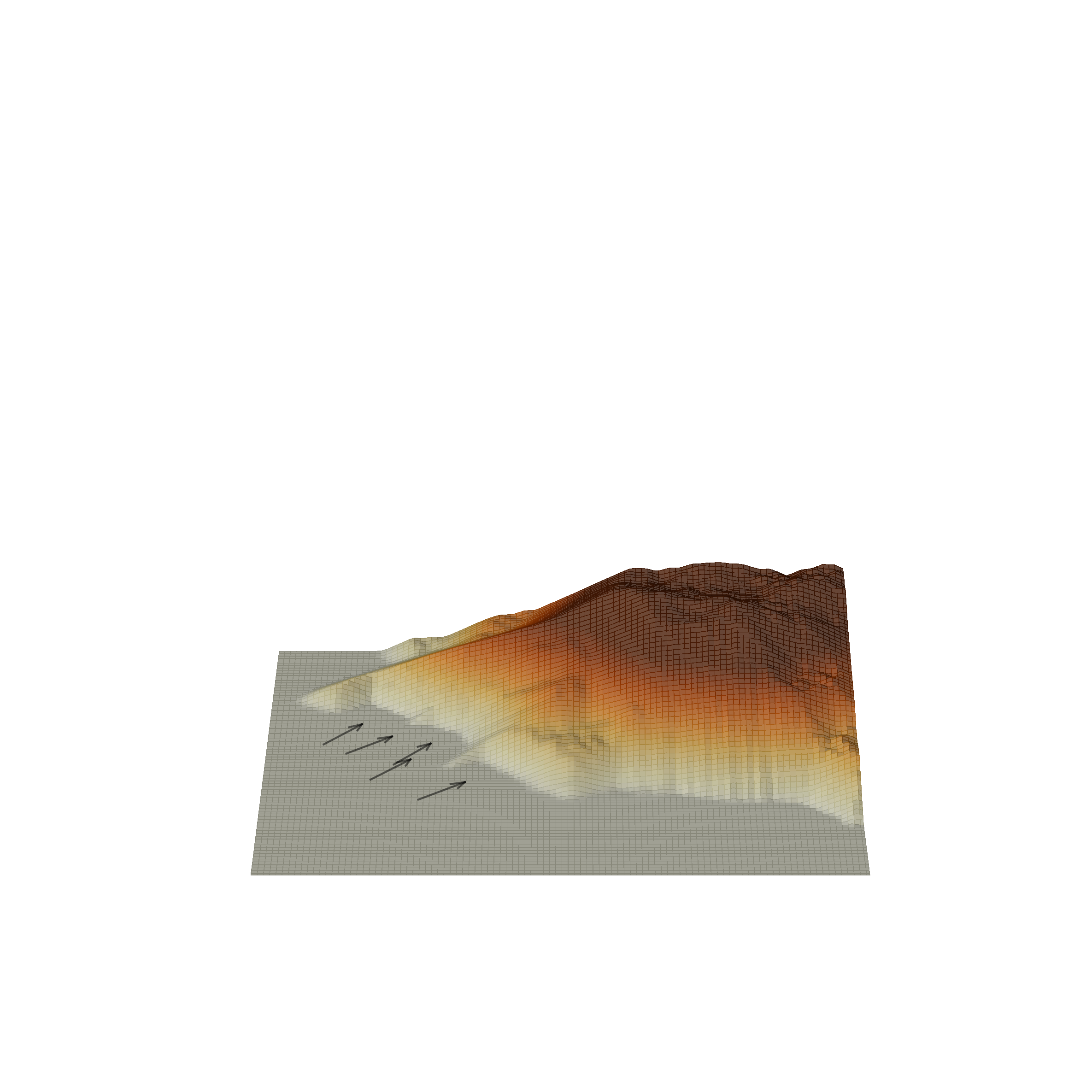}\\
            \includegraphics[clip,trim=150 140 180 290, width=0.95\textwidth]{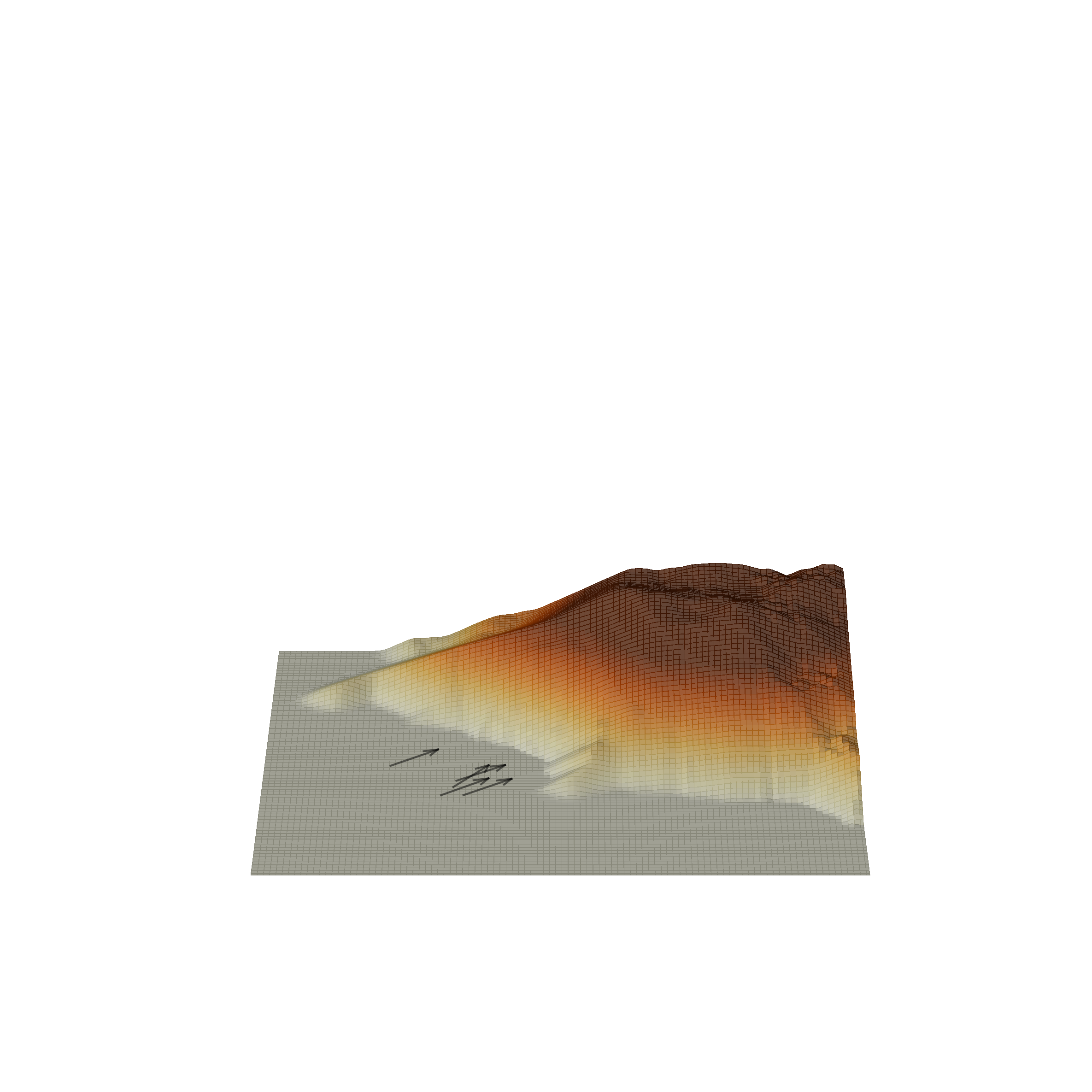}\\
            \includegraphics[clip,trim=150 140 180 290, width=0.95\textwidth]{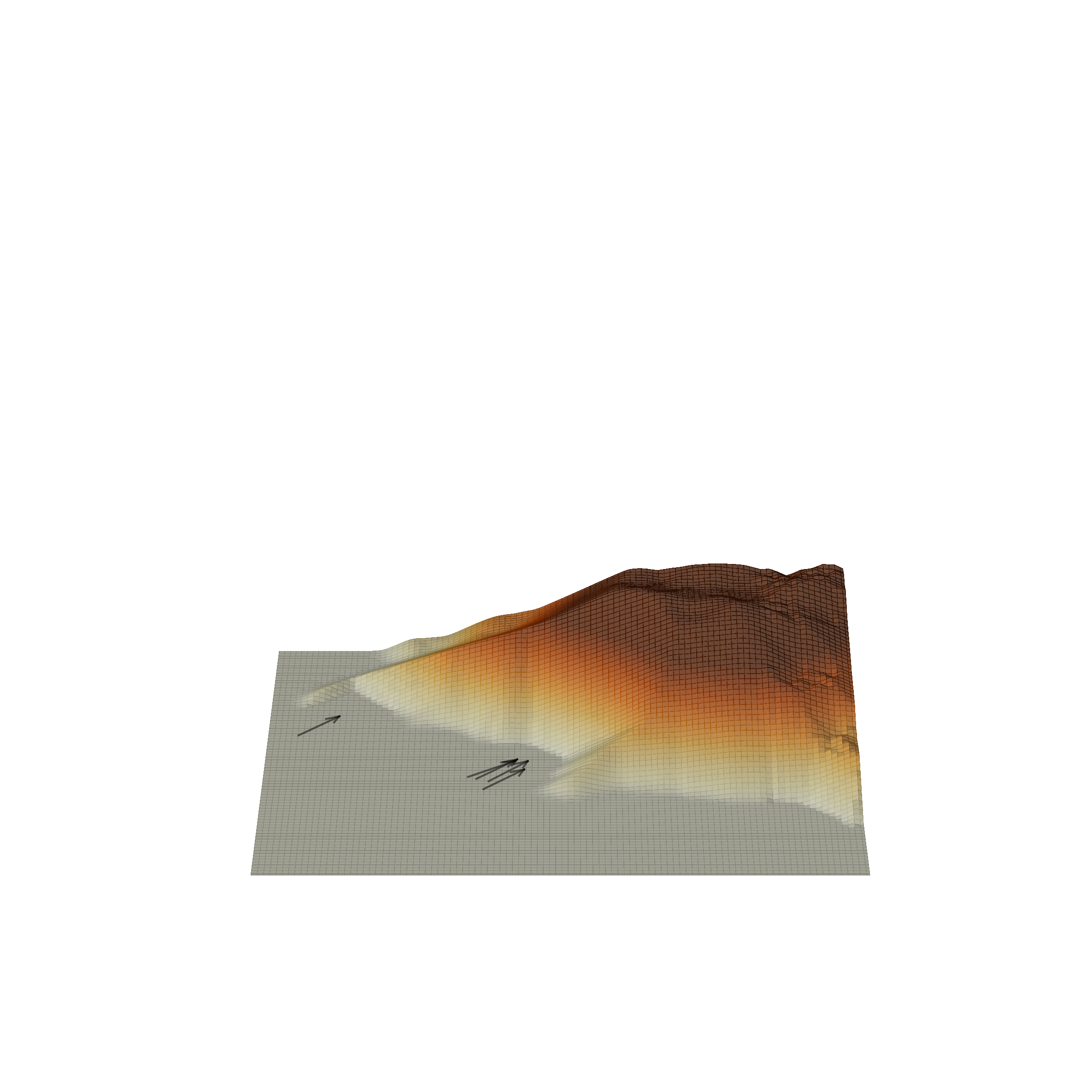}\\
            \includegraphics[clip,trim=150 140 180 290, width=0.95\textwidth]{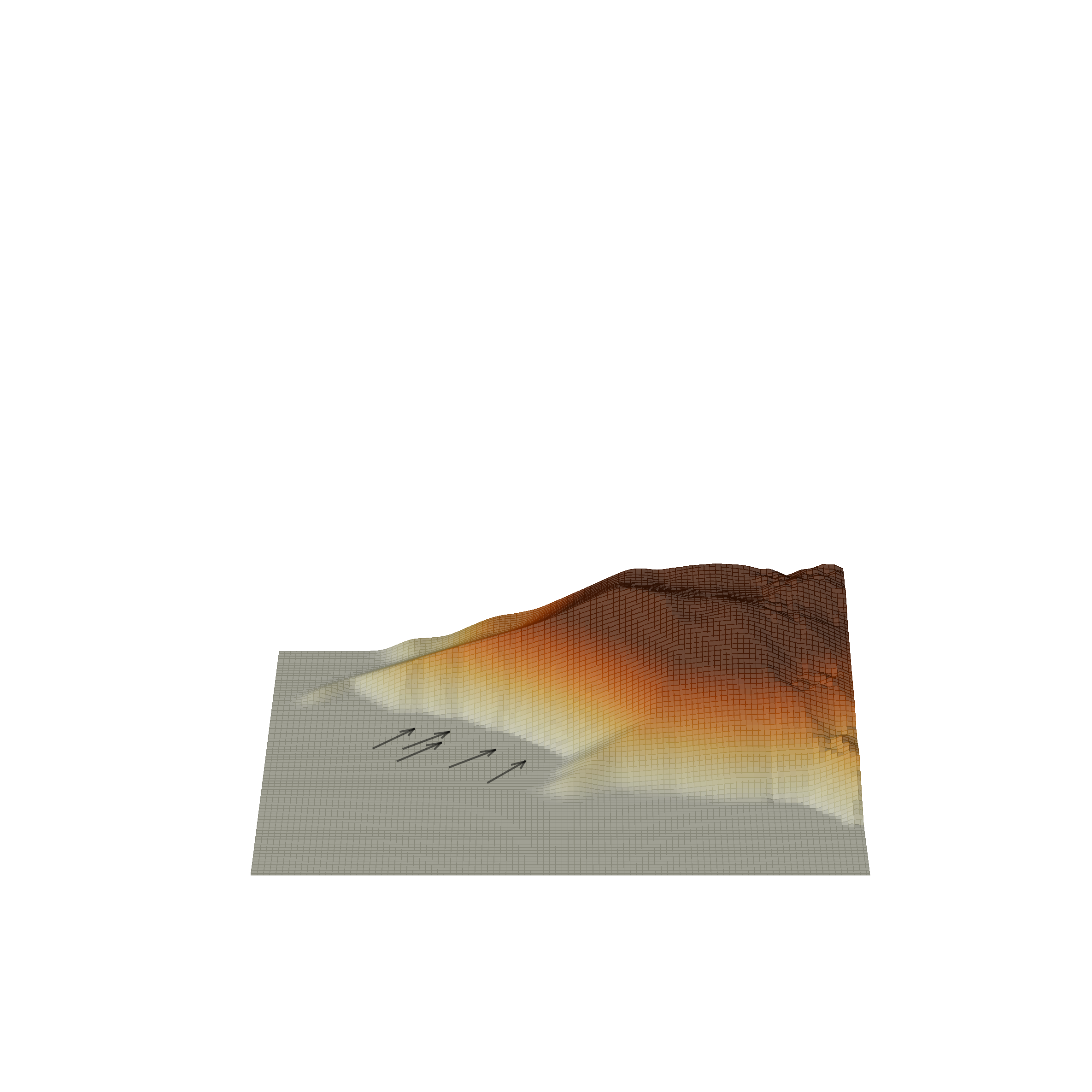}\\
            \includegraphics[clip,trim=150 140 180 290, width=0.95\textwidth]{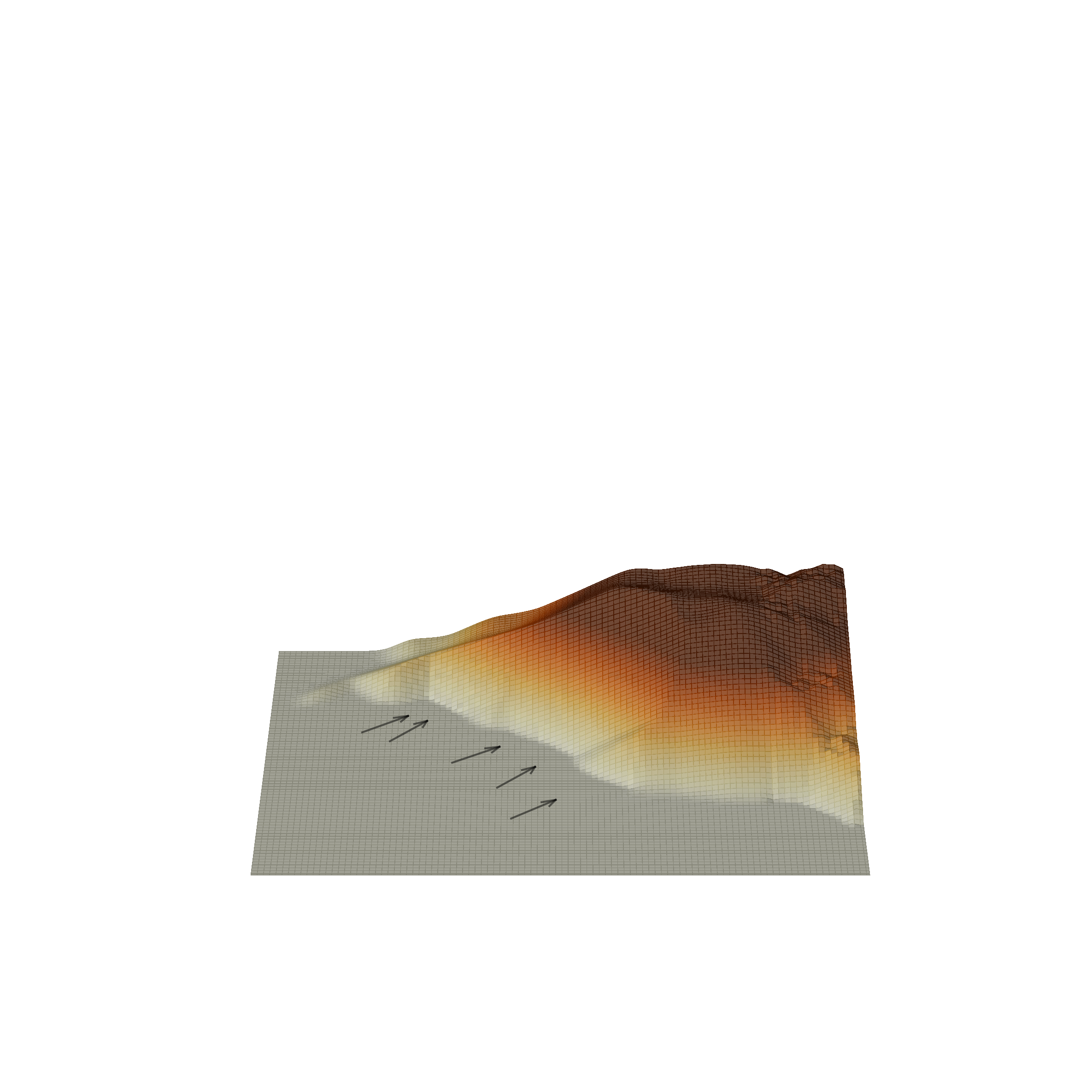}\\
            \includegraphics[clip,trim=150 140 180 290, width=0.95\textwidth]{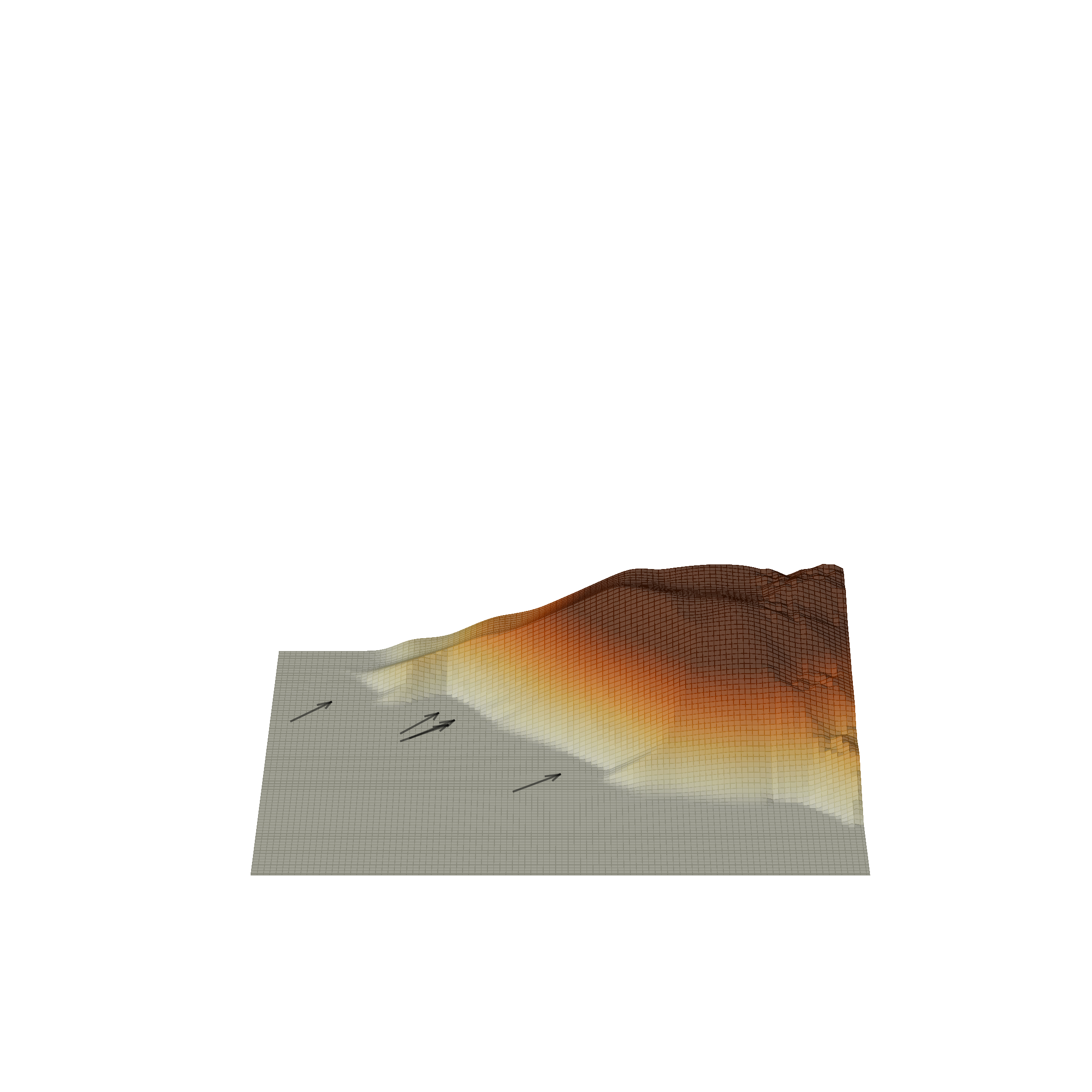}\\
            \includegraphics[clip,trim=150 140 180 290, width=0.95\textwidth]{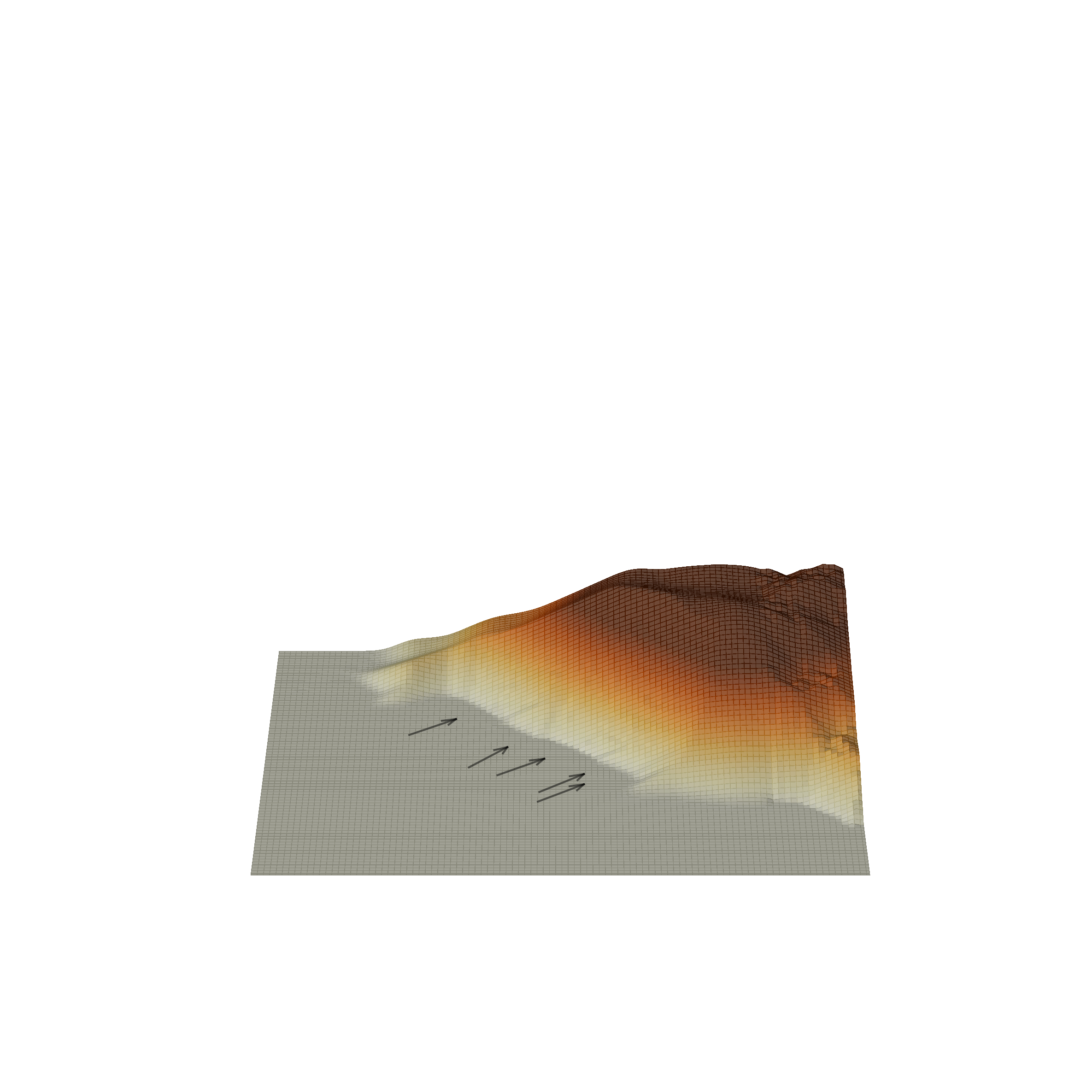}\\
            \includegraphics[clip,trim=150 140 180 290, width=0.95\textwidth]{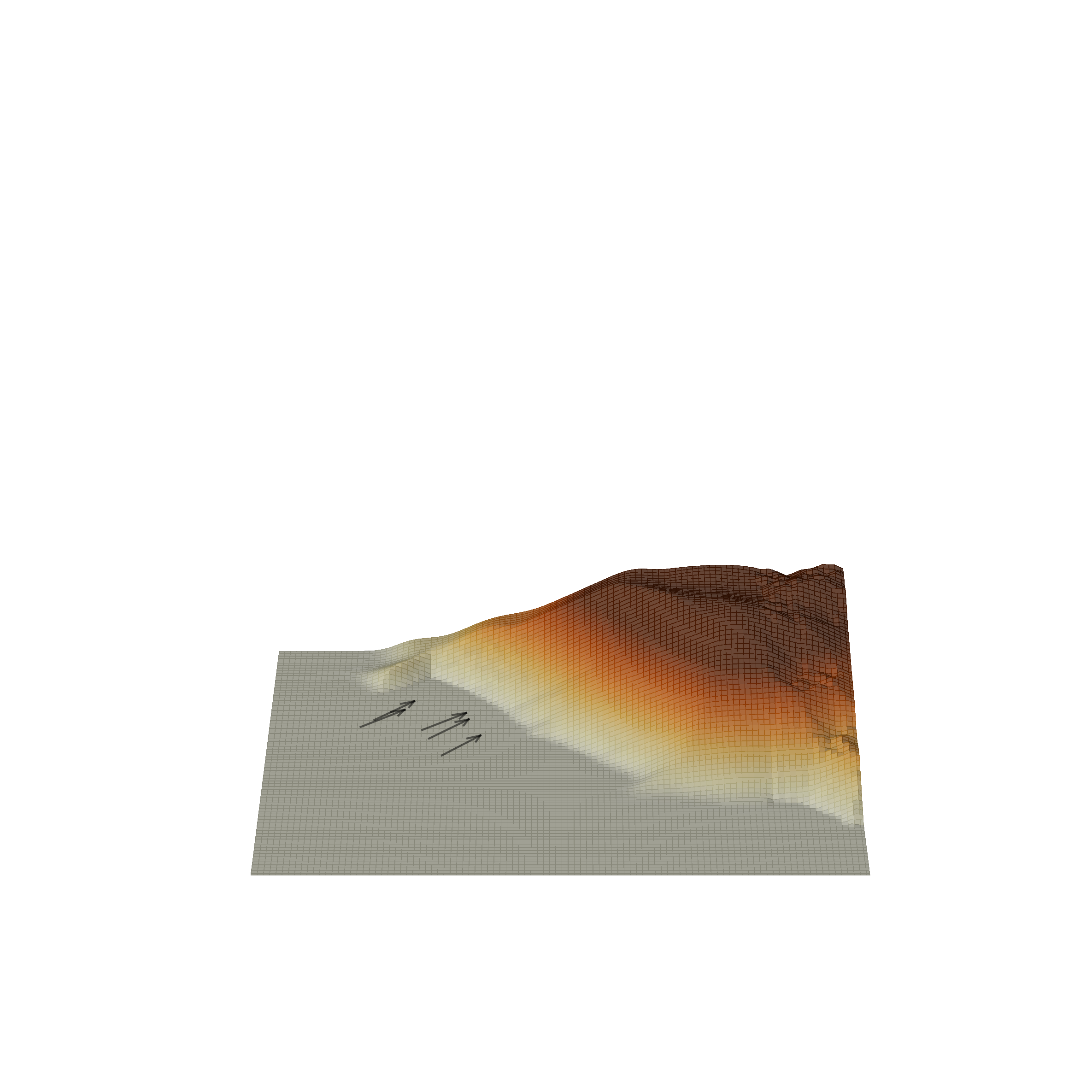}
	\end{minipage}
    }
    \caption{Pile state evolution after every five sequential loadings for (a) ground truth simulation, (b) high-dimensional model, and (c) low-dimensional model combined with cellular automata. Identical dig poses (indicated by arrows) and action parameters are used in the three cases.}
    \label{fig:Sequential_predicted_pile_state}
\end{figure}

This section compares the evolution of 40 sequential loading cycles using the high and low-dimensional world models to the simulation ground truth.
The simulation and the predictor models start with the same initial pile state, $H_1$, and use identical sequences of the dig location, heading, and action parameters, $\{\bm{x}_n, \bm{t}_n, \bm{a}_n\}_{n=1}^{40}$.
The initial pile is the one shown in Fig.~\ref{fig:overview}. The selected headings and dig locations are shown in Fig.~\ref{fig:Sequential_predicted_pile_state}.
The actions were picked randomly from the training data set.
The simulated sequential loading can be seen in Supplementary Video 2.
The performance predictor $\psi^\text{high}_\diamond$ is used together with the pile state predictor $\phi$ for high-dimensional model predictions.
The low-dimensional performance predictor $\psi^\text{low}_\diamond$ is combined with the cellular automata model to predict the next pile state.

\begin{figure} [!htb]
    \centering
    \includegraphics[clip,trim=0 55 0 0, width=0.40\textwidth]{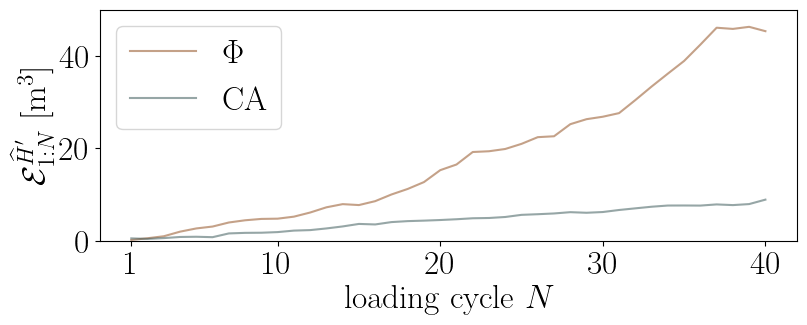} \\
    \includegraphics[clip,trim=0 55 0 0, width=0.40\textwidth]{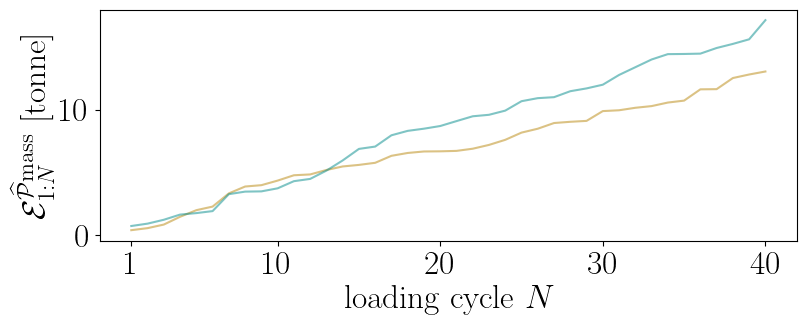} \\
    \includegraphics[clip,trim=0 55 0 0, width=0.40\textwidth]{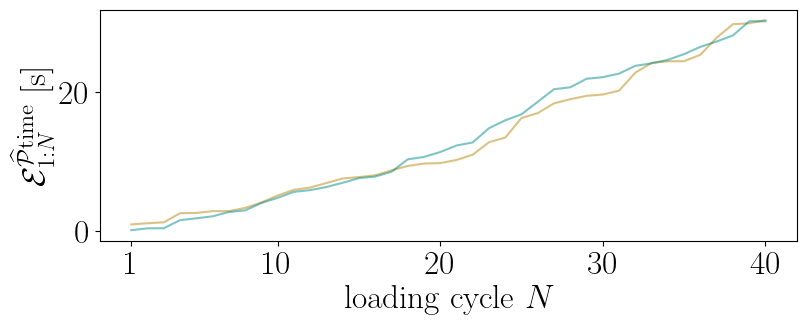} \\
    \includegraphics[clip,trim=0 0 6.0 0, width=0.387\textwidth]{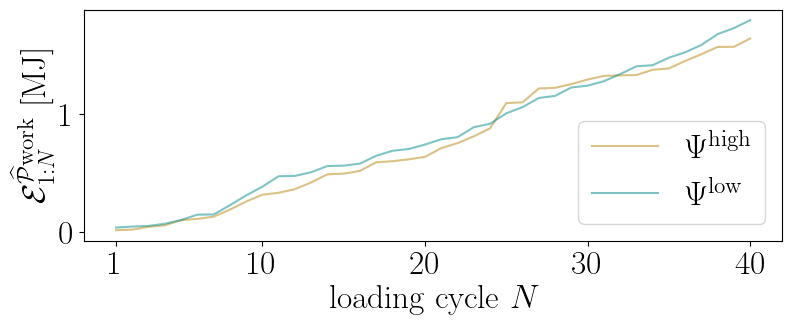}
    \caption{The evolution of the prediction errors during sequential loading.}
    \label{fig:accumulated_errors}
\end{figure}

The evolution of the pile shape is shown in Fig.~\ref{fig:Sequential_predicted_pile_state} and in Supplementary Video 2. The accumulated load mass, time, work, and residual pile volume, evolving with the number of loadings $n$, are summarized in Table~\ref{tab:sequential_total_outcomes}.
\begin{table}[!htb]
    \centering
    {\footnotesize
    \caption{The accumulated load mass, time, work, and residual pile volume during sequential loading. The number of loading cycles is denoted by $n$, and GT stands for ground truth.}
    \label{tab:sequential_total_outcomes}
    \begin{tabular}{c c c c c c c c c c c c c}
    \dtoprule
    & \multicolumn{9}{c}{$\widehat{\mathcal{P}}_{1:N}$} & \multicolumn{3}{c}{$\widehat{H}'_{N}$}\\
    \cmidrule(r){2-10} \cmidrule(r){11-13}
     & \multicolumn{3}{c}{load mass [tonne]} & \multicolumn{3}{c}{loading time [s]} & \multicolumn{3}{c}{work [MJ]} & \multicolumn{3}{c}{pile volume [m$^3$]}\\
    \cmidrule(r){2-4} \cmidrule(r){5-7} \cmidrule(r){8-10} \cmidrule(r){11-13}
    $N$ & GT & $\Psi^\text{high}$ & $\Psi^\text{low}$ & GT & $\Psi^\text{high}$ & $\Psi^\text{low}$ & GT & $\Psi^\text{high}$ & $\Psi^\text{low}$ & GT & $\Phi$ & \texttt{cell.aut.} \\ \dtoprule
5 & 14.7 & 15.7 & 16.1 & 55.8 & 53.2 & 54.5 & 2.0 & 2.0 & 2.0 & 768.6 & 765.9 & 767.7 \\
10 & 28.2 & 31.5 & 31.3 & 106.1 & 101.0 & 101.9 & 4.1 & 3.9 & 3.8 & 760.7 & 755.9 & 758.8 \\
15 & 40.8 & 45.3 & 47.0 & 161.1 & 155.4 & 154.5 & 6.5 & 6.1 & 6.0 & 753.3 & 745.6 & 749.7 \\
20 & 56.8 & 62.0 & 64.4 & 218.0 & 210.8 & 208.1 & 8.9 & 8.5 & 8.3 & 744.0 & 728.7 & 739.5 \\
25 & 71.0 & 75.4 & 80.6 & 273.7 & 265.7 & 260.2 & 11.1 & 10.6 & 10.4 & 735.7 & 714.7 & 730.1 \\
30 & 86.1 & 90.5 & 96.6 & 327.3 & 318.7 & 308.9 & 13.2 & 12.8 & 12.4 & 726.9 & 700.1 & 720.7 \\
35 & 101.4 & 106.3 & 114.3 & 389.4 & 385.1 & 369.9 & 15.7 & 15.3 & 14.6 & 718.0 & 679.1 & 710.4 \\
40 & 116.0 & 118.6 & 130.9 & 443.0 & 441.9 & 418.8 & 17.9 & 17.4 & 16.6 & 709.5 & 664.2 & 700.6 \\
    \dbottomrule
    \end{tabular}
    }
\end{table}
%
%
%
The error accumulation over time is shown in Fig.~\ref{fig:accumulated_errors}.
While the pile state prediction error for the low-dimensional model grows at a nearly constant rate, 
the growth rate of the high-dimensional model is more irregular, starting at a lower rate but growing more rapidly after 15 sequential loadings. The error in loaded mass grows nearly linearly and roughly at the same rate for both models. Also, the errors in loading time and work grow mainly linearly. Both models follow the same trend.
The high-dimensional model uses $\bm{\Phi}$ for the local pile state prediction, making no change outside the local heightmap as discussed in Sec.~\ref{sec:pile_state_predictor_model}.
The accumulated error eventually leads to a predicted pile state with a slope steeper than the angle of repose (Fig.~\ref{fig:Sequential_predicted_pile_state}).

The high- and low-dimensional models' accumulated inference time was also measured over the 40 loading cycles. The high-dimensional model took 0.3\,s in total. The low-dimensional model took 0.01\,s. When the cellular automata is included, this amounts to 44.3\,s, but it should be noted that the implementation of the cellular automata was not optimized or adapted for GPU computing.
%

To summarize, the high-dimensional model was better for shorter time horizons, but the low-dimensional model was more stable over longer horizons.
The stability might be the same if the high-dimensional model used the cellular automata in place of the pile state predictor, but this would come with an additional computational cost.

\subsection{Diggability map}
\label{sec:diggability_map}
\begin{figure} [!htb]
    \begin{subfigure}{0.32\textwidth}
        \centering
        \includegraphics[clip,trim=165 170 195 250, width=1.0\textwidth]{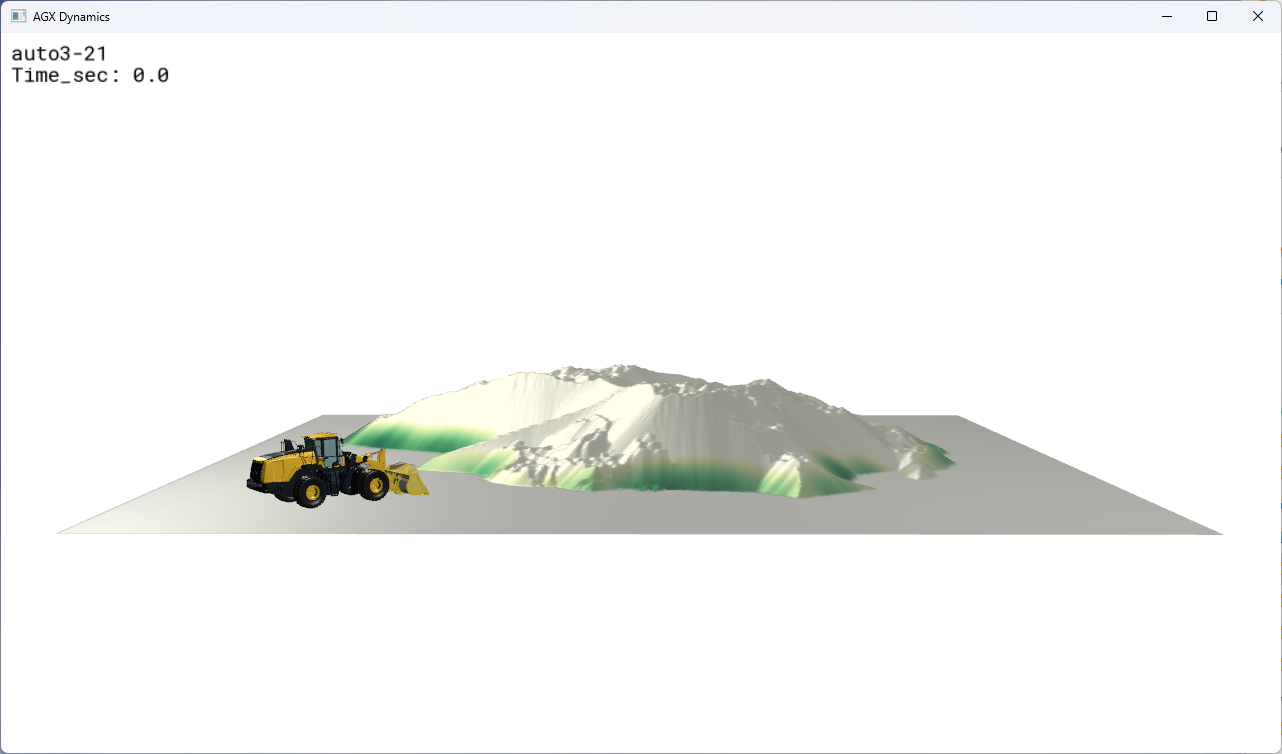} \\
        \includegraphics[clip,trim=215 205 205 205, width=1.0\textwidth]{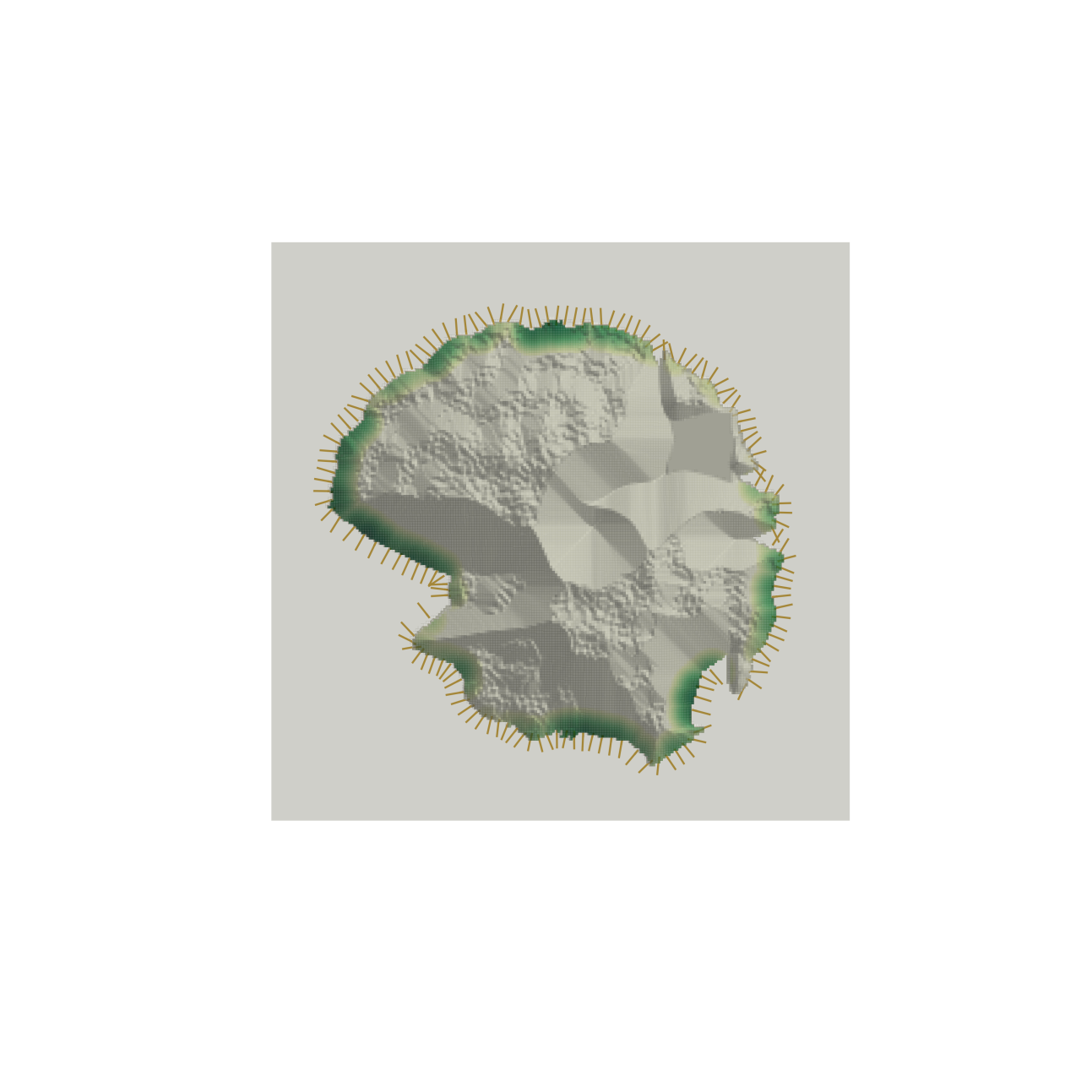} \\
        \includegraphics[clip,trim=20 15 20 20, width=1.0\textwidth]{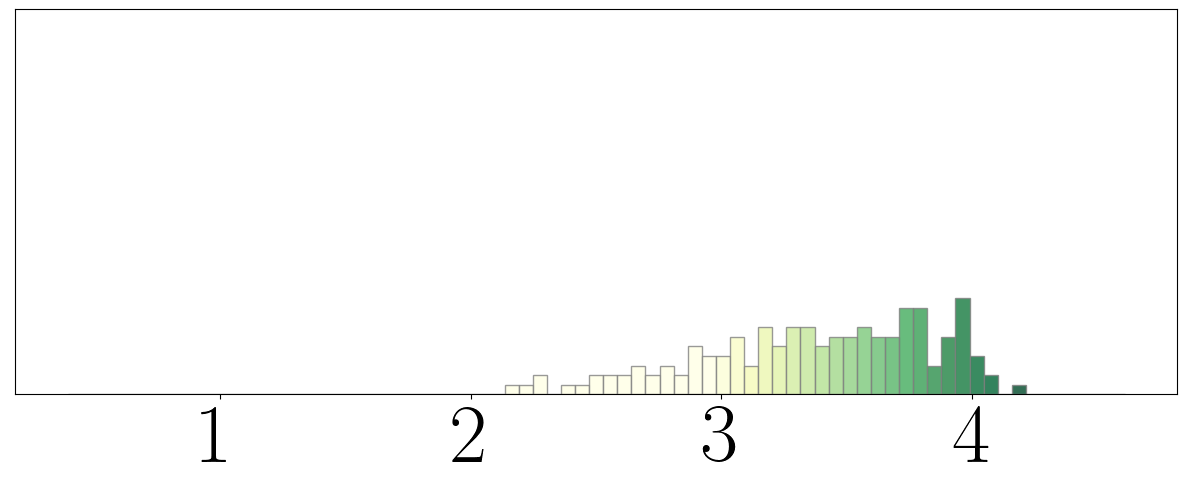}
        \caption{Load mass [tonne].}
    \end{subfigure}
    \begin{subfigure}{0.32\textwidth}
        \centering
        \includegraphics[clip,trim=165 170 195 250, width=1.0\textwidth]{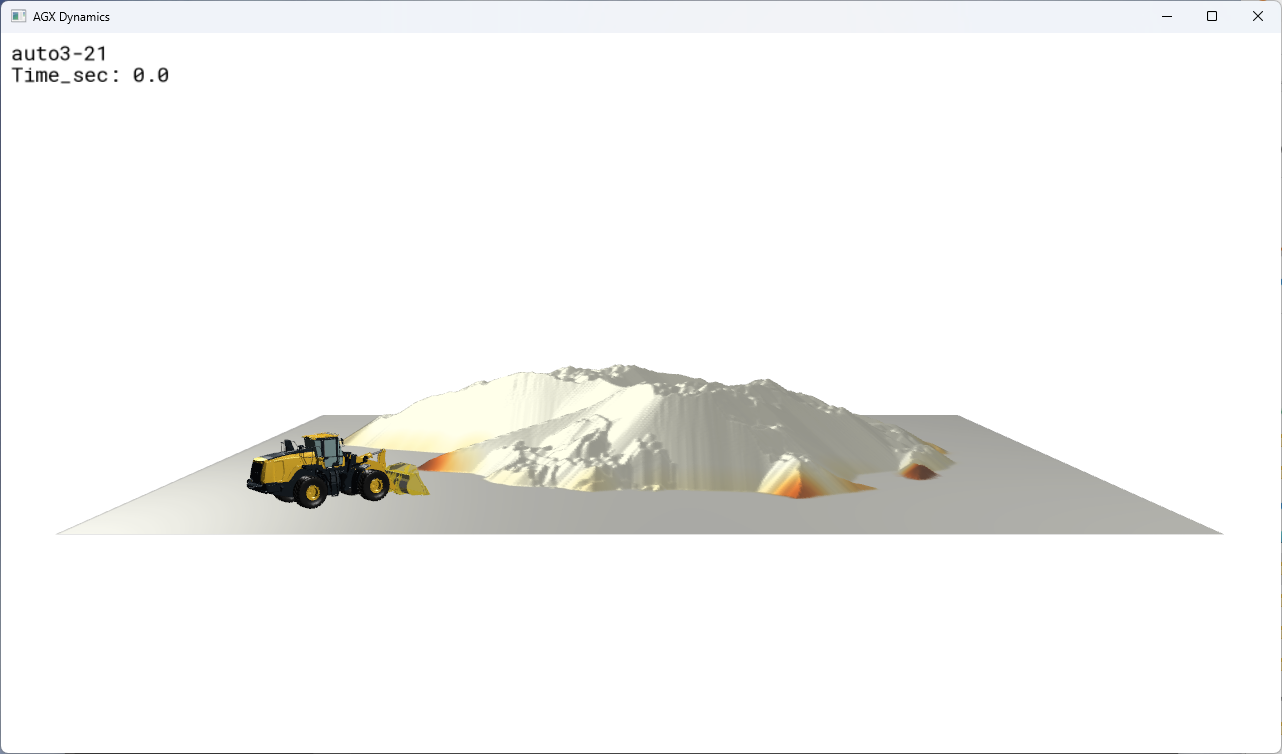} \\
        \includegraphics[clip,trim=215 205 205 205, width=1.0\textwidth]{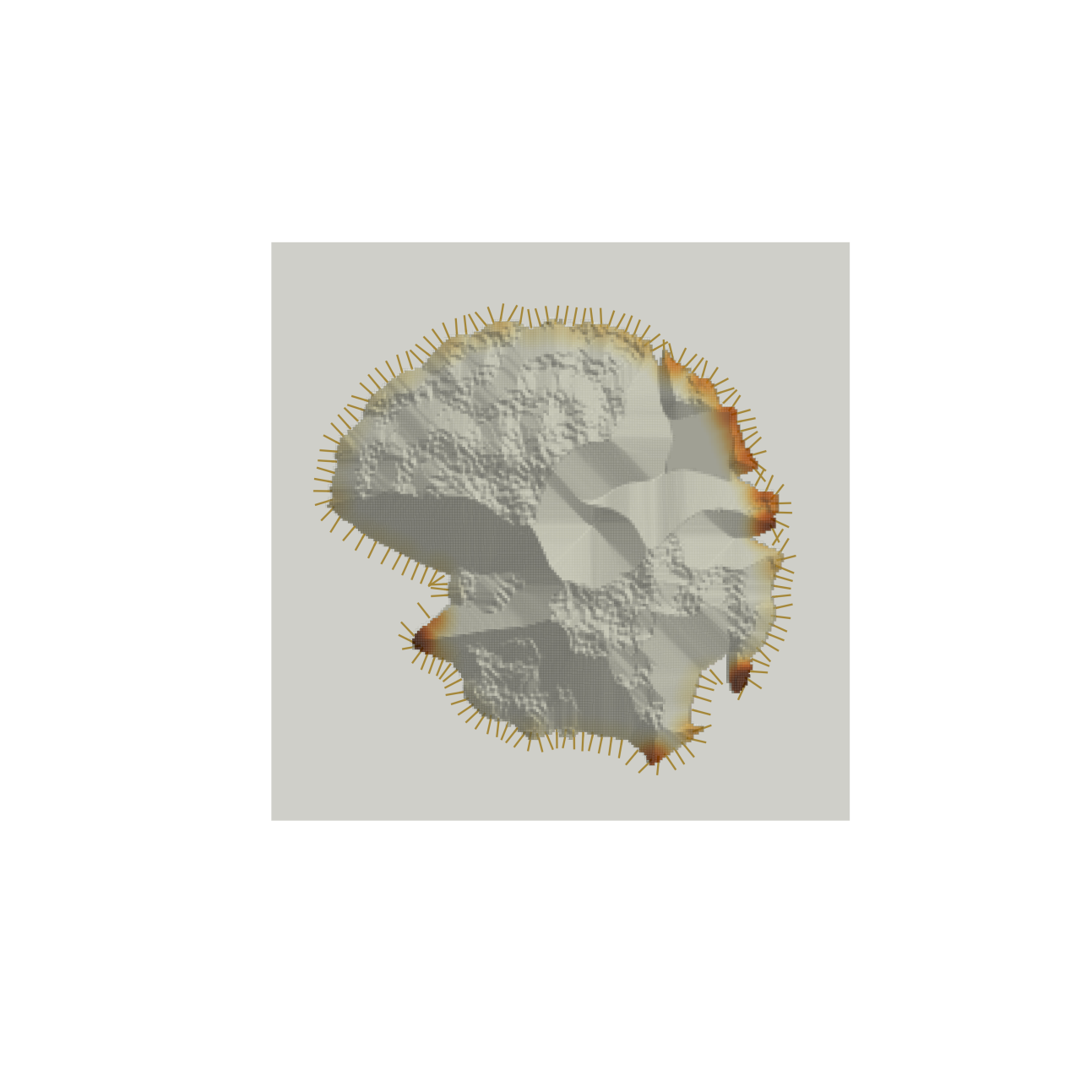} \\
        \includegraphics[clip,trim=20 15 20 20, width=1.0\textwidth]{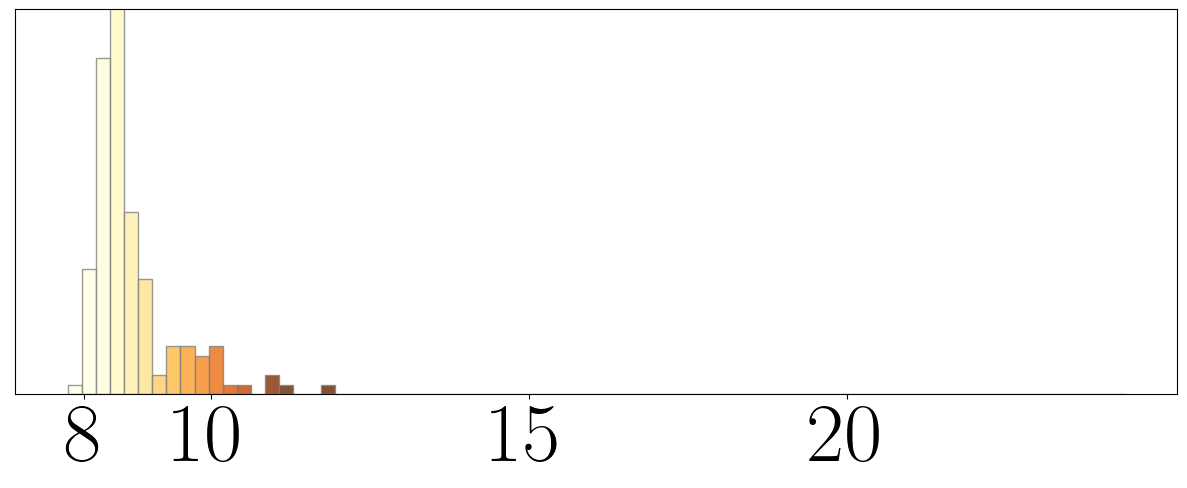}
        \caption{Time [s].}
    \end{subfigure}
    \begin{subfigure}{0.32\textwidth}
        \centering
        \includegraphics[clip,trim=165 170 195 250, width=1.0\textwidth]{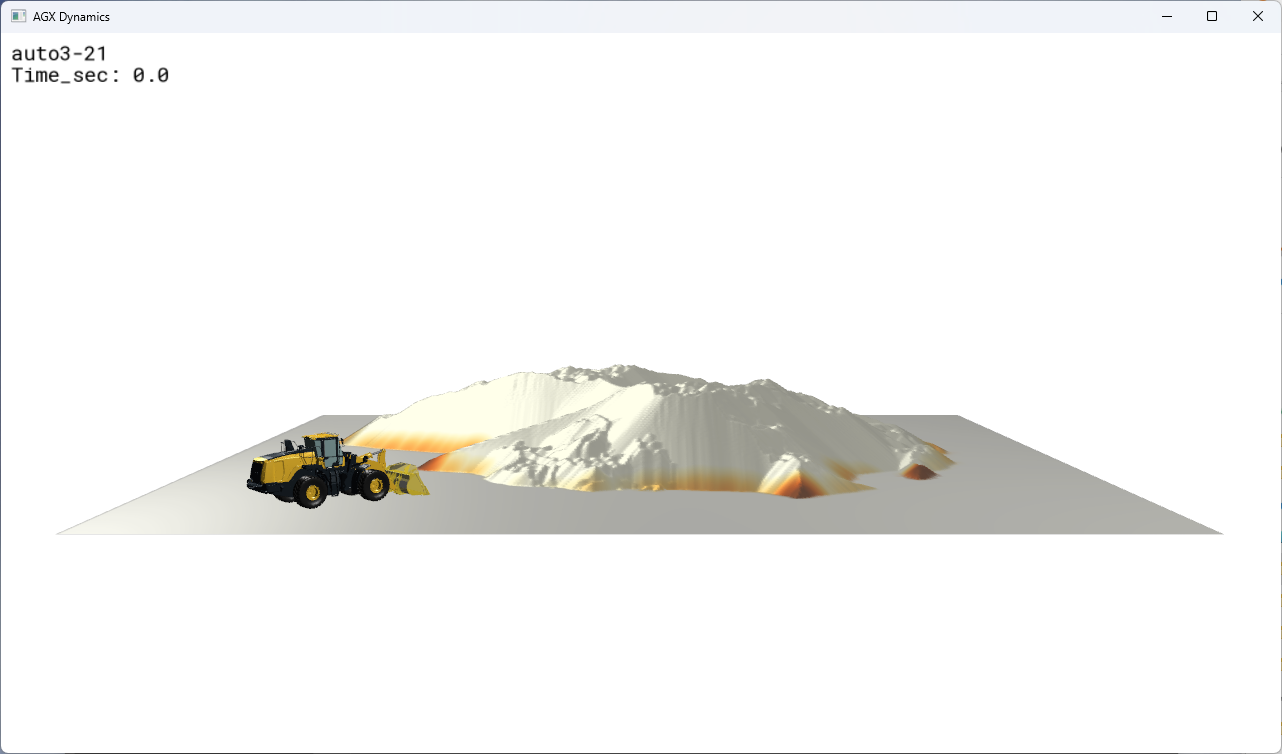} \\
        \includegraphics[clip,trim=215 205 205 205, width=1.0\textwidth]{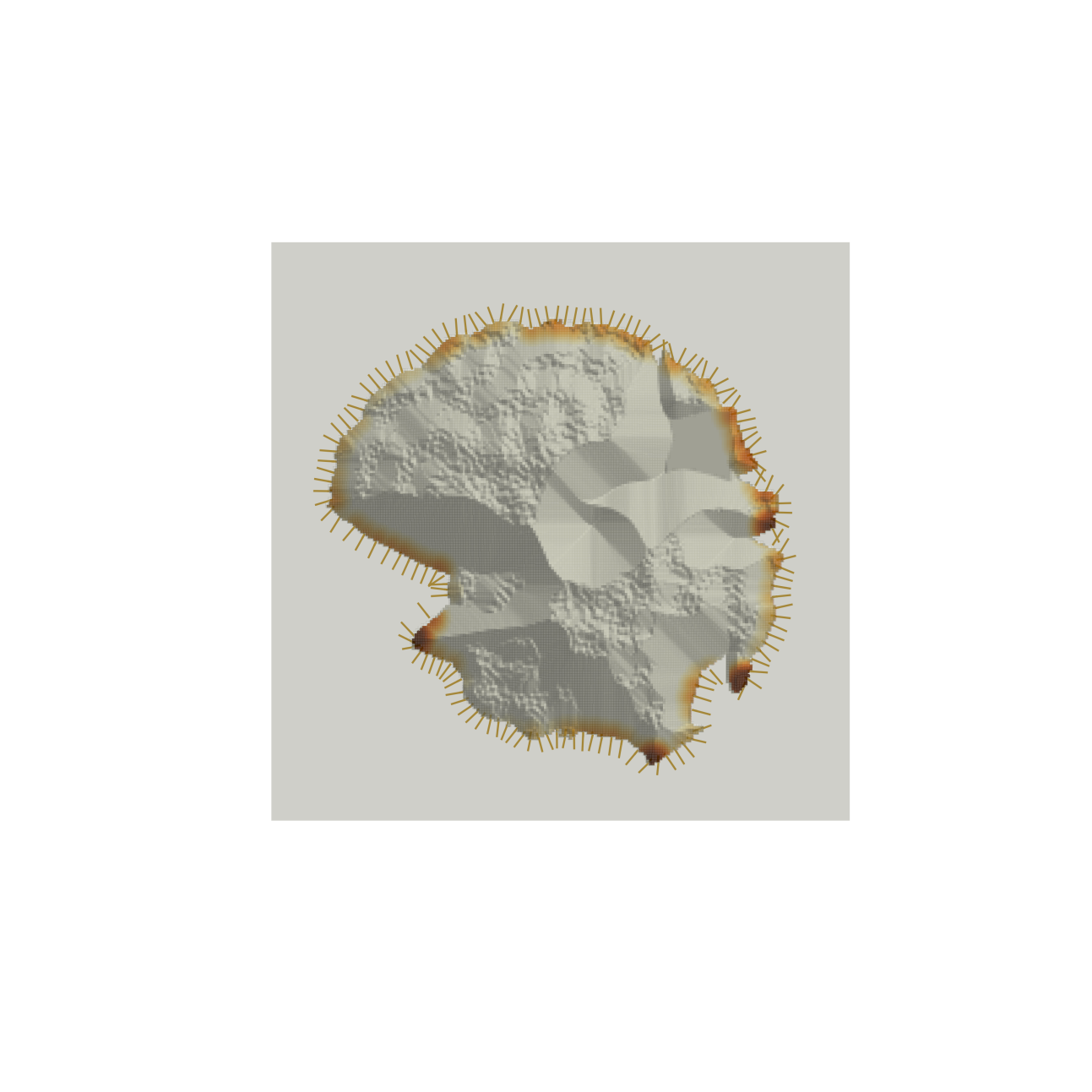} \\
        \includegraphics[clip,trim=15 15 25 20, width=1.0\textwidth]{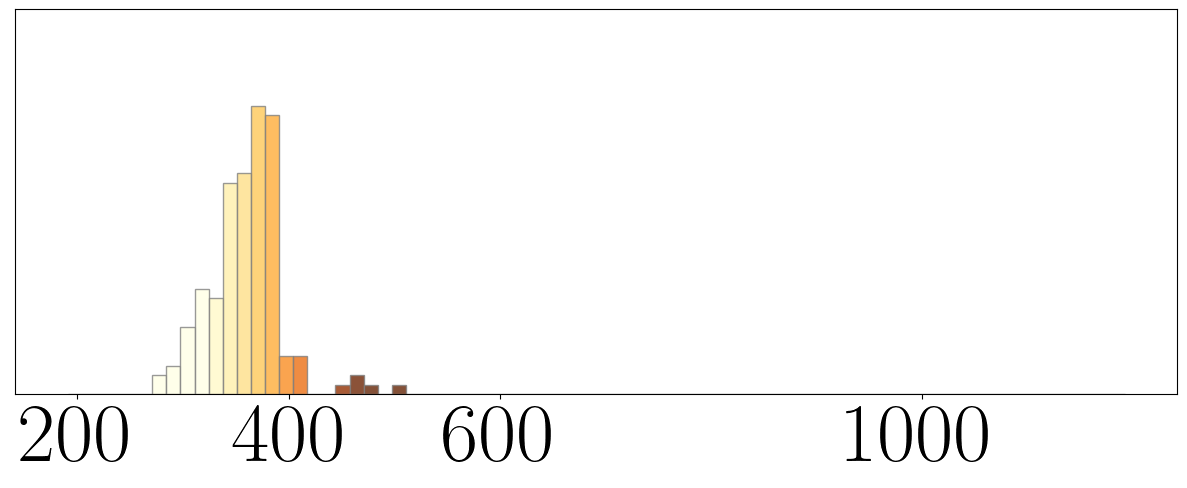} %
        \caption{Work [kJ].}
    \end{subfigure}
    \caption{Diggability maps that show the predicted loaded mass, time, and work at 150 locations with normal headings, indicated by the lines around the edge. The color coding is explained by the histograms for each of the predicted quantities.}
    \label{fig:diggability_map}
\end{figure}

The predictor model can be used to search for the most favourable dig location around the pile.
We demonstrate this by creating \emph{diggability maps}, where the model has been inferred around 
the entire edge of a pile. 
Fig.~\ref{fig:diggability_map} shows a map created by using $\bm{\Psi}^\text{high}_\diamond$ to predict the performance using a fixed action $\bm{a}=[0.68, 4.51, 0.17, 4.46]$ at 150 dig locations along the local normal direction. 
The map highlights the regions with higher pile angles where the loaded volume would be relatively high and would be dug with lower energy cost. 
Low-performance regions are visible where the pile bulges.
The diggability maps can be further improved by searching also for the locally optimal heading and loading action.

\section{Discussion}
\label{sec:discussion}
\subsection{Feasibility}
The relatively small errors of single loading prediction, 3--8.5 \%, are encouraging. There are, however, several practical limitations with the the method in its current form. The model is particular to the simulator's specific wheel loader and soil, which was homogeneous gravel. Consequently, a new model needs to be trained for each combination of wheel loader model and soil type unless this is made part of the model input. There is a greater need for automation solutions for more complex soils, such as blasted rock or highly cohesive media, than for homogeneous gravel. Although these materials can be numerically simulated, the problem of learning predictor models may be harder and require different data, as discussed for the loading process in \cite{BORNGRUND2022104013}. For blasted rock, the local rock fragmentation probably needs to be included in the model input. On the other hand, the level of accuracy needed ultimately depends on the specific application and the prediction horizon required.

It is an attractive option to train on actual field data (rather than simulated data) to eliminate any simulation bias. That would entail collecting data from several thousand loadings with variations in pile shape and action parameters. This would be demanding but not unrealistic given the number of vehicles (of the same model) that are in operation worldwide. If all were equipped with the same bucket-filling controller, it would ultimately be a question of instrumenting these machines and sites to scan the pile shape before and after loading and tracking the vehicle’s motion and force measurements.

Long-horizon planning of sequential loading requires long rollouts of repeated model inference. The error accumulation in the pile state predictor model (Fig.~\ref{fig:accumulated_errors}) might then become an obstacle. If that is the case, the pile state predictor can be replaced with mass removal and cellular automata, as described in Sec.~\ref{sec:avalanching_algorithm}, making the cellular automata the computational bottleneck. In our implementation, the cellular automata was about 200 times slower than model inference, about 1 s versus 5 ms per loading. An optimized implementation can be an order of magnitude faster at least. An alternative is to use the cellular automata as a post-processor, forcing the predicted pile state to be consistent with the soil's angle of repose. If the post-processing is only occasionally needed, instead of after each pile state prediction, the computational overhead may be marginal.

\subsection{Applications}
\label{sec:applications}
We envision that the prediction models can be used in several different ways.
They can be used to select the next loading action of an autonomous wheel loader
or to plan the movement of both wheel loaders and haul trucks in a way that is
optimal for coordinating multiple loading and hauling vehicles with the multi-objective 
goal of executing individual tasks efficiently while not obstructing the work of the 
other machines. If the problem of optimal sequential planning is computationally 
intractable, the prediction models can be useful in developing good planning policies,
for instance, using model-based reinforcement learning.

%
\section{Conclusion}
\label{sec:conclusion}
This paper shows the feasibility of learning wheel loader world models that predict the outcome of single loading cycles given the local shape of the pile and the choice of control parameters for automatic bucket-filling. The proposed models can be used for automatic planning and control to maximize the net performance of a sequence of loading cycles predicted through repeated model inference. Topics left to explore in future work include handling more complex materials and how the proposed methods can be used for optimal planning. The latter would also provide better insight into model accuracy and inference speed requirements.

\section*{Supplementary Material}
Supplementary videos to this article can be found online at \url{http://umit.cs.umu.se/wl-predictor/}.

\section*{Acknowledgement}
The research was supported in part by Komatsu Ltd, Algoryx Simulation AB, the Wallenberg AI, Autonomous Systems and Software Program (WASP) funded by the Knut and Alice Wallenberg Foundation, and Swedish National Infrastructure for Computing at High-Performance Computing Center North (HPC2N).

\bibliographystyle{unsrt} 
\bibliography{loader_predictor_model}
\newpage

\begin{appendices}

\section[\appendixname~\thesection]{Overview of the dataset}
\label{appendix:dataset_overview}
\begin{figure} [!htb]
    \centering
    \includegraphics[clip,trim=0 0 0 0, width=0.3\textwidth]{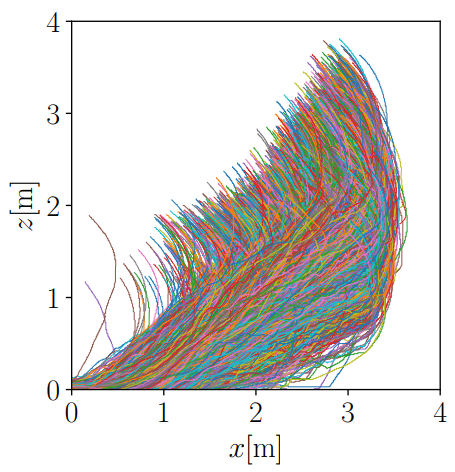}
    \caption{All bucket-tip trajectories in the collected dataset, randomly colored.
    }
    \label{fig:trajectories}
\end{figure}

\begin{figure} [!htb]
    \centering
    \includegraphics[clip,trim=0 0 0 0, width=1.0\textwidth]{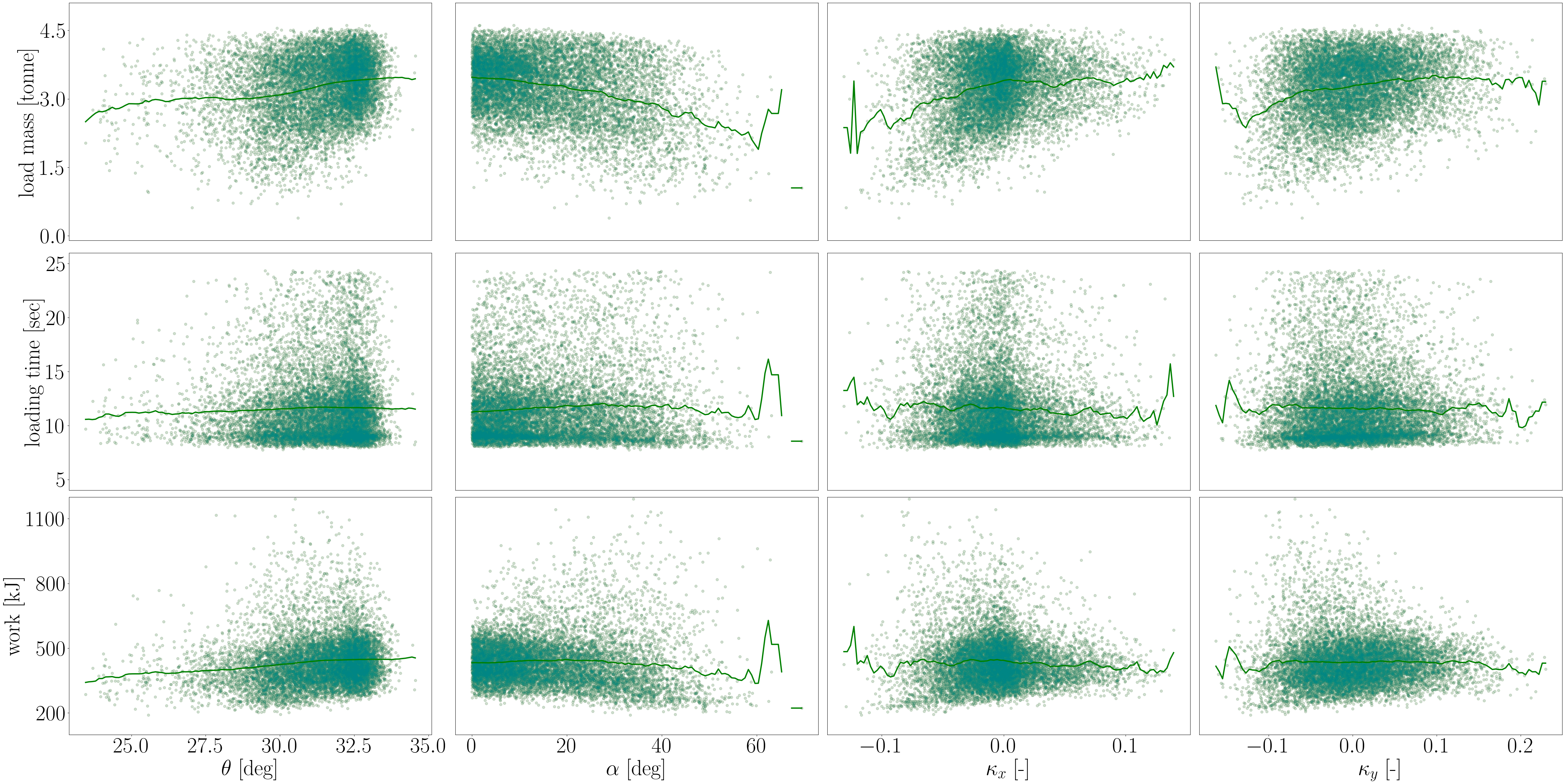}
    \caption{The distribution of the collected performance measurements and how they are correlated with the characteristic pile slope $\theta$, incidence angle $\alpha$, and curvature $\kappa_x$ and $\kappa_z$. The line is the performance moving average of each characteristics variable.}
    \label{fig:character_target}
\end{figure}

First, the general characteristics of the collected dataset were investigated.
As expected, we observed a widespread in bucket-tip trajectories because of variability in loading action parameters and pile states (Fig.~\ref{fig:trajectories}). 
The loading performance, shown in Fig.~\ref{fig:character_target}, is also well distributed in the intervals of 1\--4.5\,tonne, 7\--25\,s, and 200\--1100\,kJ.
As expected from a previous study~\cite{singh2006factors}, productivity (loaded mass per time unit) is positively correlated with the slope angle and negatively with the incidence angle.

\section[\appendixname~\thesection]{Model performance dependency on hyperparameters}
\label{appendix:hyper_parameters}
\begin{figure} [!htb]
    \centering
    \includegraphics[clip,trim=150 80 100 50, width=0.95\textwidth]{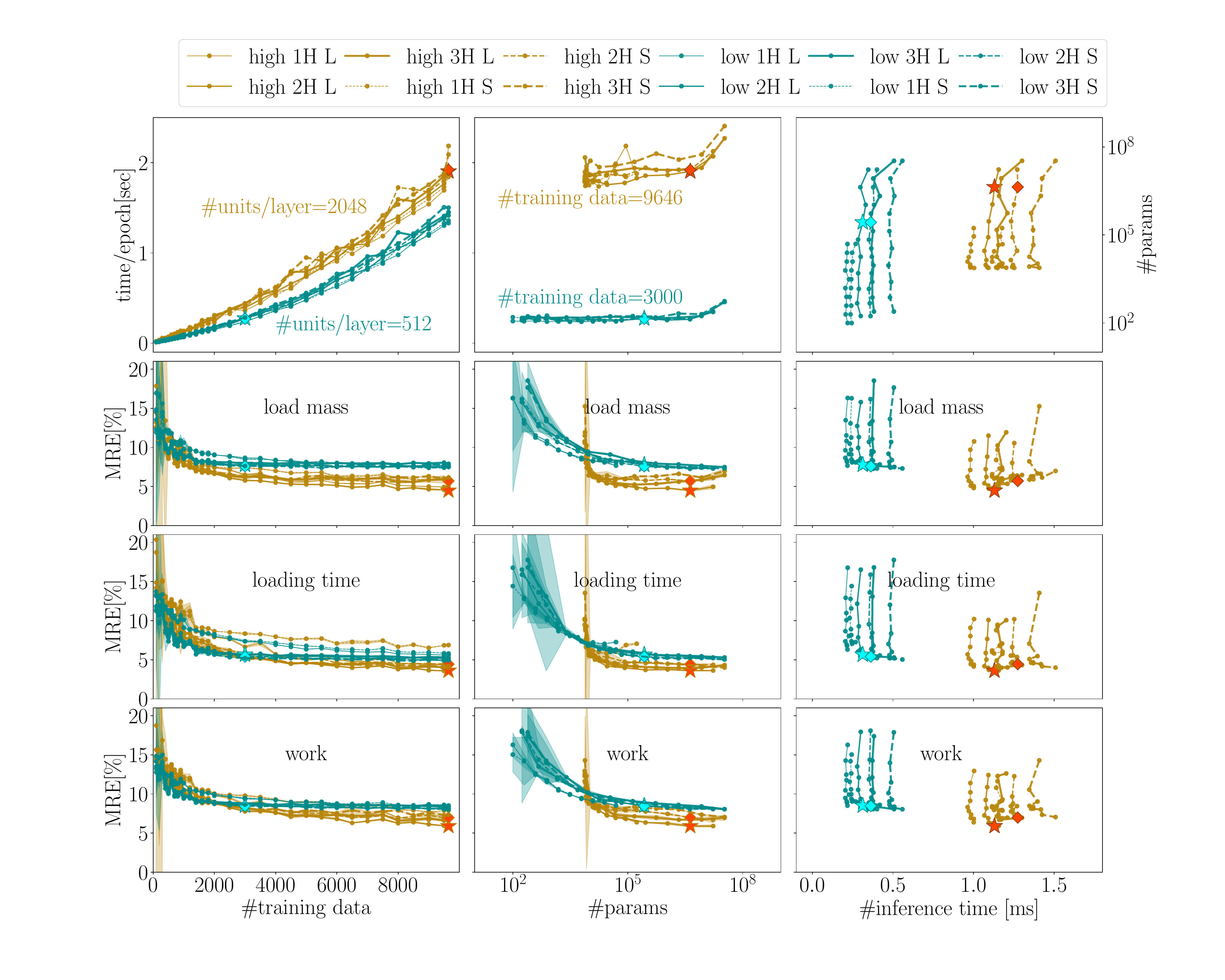}
    \caption{The trend of generalization error (MRE), inference time, and training time (time/epoch)
        with the number of training/validation data (\#training data) and parameters (\#params).
        The parameters are changed by the number of units/layer and the number of hidden layers (H).
        The activation function used is either Leaky ReLU (L) or Swish (S).
        Note that the number of units is fixed at 512 for $\bm{\Psi}^\text{low}$ and 2048 for $\bm{\Psi}^\text{high}$ in the figures of \#training data vs. MRE and time/epoch. The number of training/validation data is fixed at 3,000 for $\bm{\Psi}^\text{low}$ and 9,646 for $\bm{\Psi}^\text{high}$ in the other figures.
        The filled symbols $\diamond$ and $\star$ identify the models of Swish and Leaky ReLU as the extreme examples of higher inference speed $\bm{\Psi}^\text{low}$, and higher accuracy model $\bm{\Psi}^\text{high}$.
        Table~\ref{tab:extreme_models} shows these specific results.
        }
    \label{fig:generalization_errors}
\end{figure}

The effect on the model performance by changing the amount of training data and the number of model parameters in the MLP is
shown in Fig.~\ref{fig:generalization_errors}. 
As can be expected, the MRE decreases with increasing size of the dataset and the model, but the accuracy eventually levels out.
The best high-dimensional model was achieved when using the full training dataset (9,646 samples). 
The precise MLP architecture was less important.

\section[\appendixname~\thesection]{Model differentiability}
\label{appendix:differentiability}
\begin{figure} [!htb]
    \centering
    \includegraphics[clip,trim=0 0 0 0, width=0.95\textwidth]{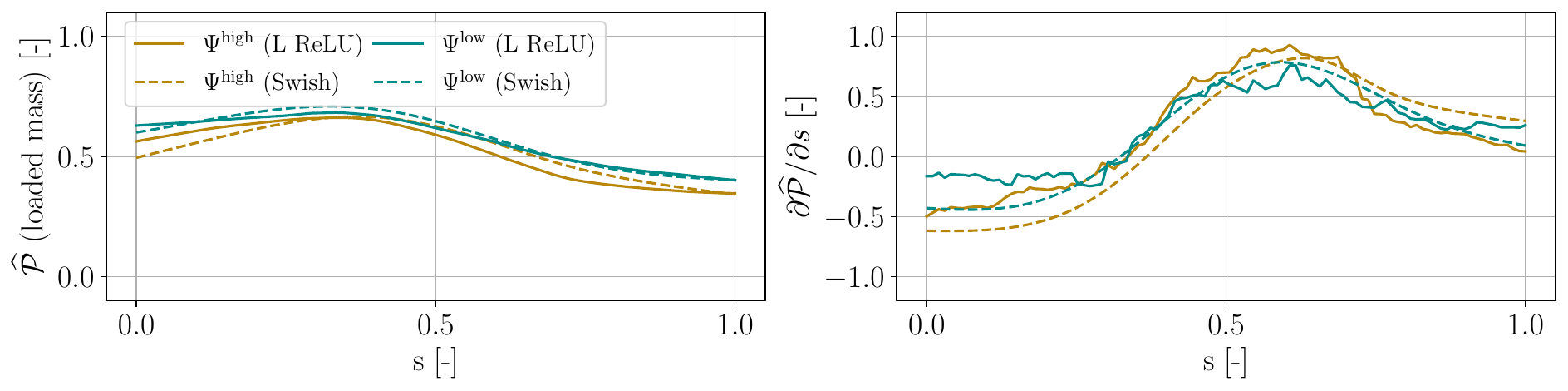}
    \caption{The function values (left) and the Autograd gradients (right) of Leaky ReLU and Swish models with respect to the action $\bm{a}$ along a direction in the action space. The prediction values are normalized.}
    \label{fig:derivative}
\end{figure}

The choice of activation function has a marginal effect on the model's accuracy,
but the quality of gradients also needs to be investigated. 
To this end, we parameterized a line through the action space using $\bm{a} = \bm{a}_0 + (\bm{a}_1-\bm{a}_0)s$, with $\bm{a}_0 = [0.7, 0.0, 0.0, 5.0]^\text{T}$, $\bm{a}_1 = [0.0, 5.0, 0.7, 0.0]^\text{T}$, and $s \in [0,1]$.
The gradients were calculated by algorithmic differentiation using Pytorch Autograd,  with respect to the actions along $\bm{a}(s)$ using the chain rule,
\begin{equation*}
    \cfrac{\partial \mathcal{P}}{\partial s} = \cfrac{\partial \mathcal{P}}{\partial \bm{a}} \cfrac{\partial \bm{a}}{\partial s}.
\end{equation*}
Fig.~\ref{fig:derivative} shows the result for the selected models. The Swish model produces smoother derivatives than the Leaky ReLU model, which can be expected since Leaky ReLU is not everywhere differentiable, although both appear useful for root-seeking. 
\end{appendices}
\end{document}